\documentclass[10pt,letterpaper,twocolumn]{article}
\usepackage[latin9]{inputenc}
\usepackage{array}
\usepackage{float}
\usepackage{multirow}
\usepackage{amsmath}
\usepackage{amssymb}
\usepackage{graphicx}
\usepackage[unicode=true,
 bookmarks=false,
 breaklinks=true,pdfborder={0 0 1},backref=section,colorlinks=false]
 {hyperref}
\hypersetup{
 pagebackref=true,letterpaper=true,colorlinks}

\makeatletter

\pdfpageheight\paperheight
\pdfpagewidth\paperwidth

\providecommand{\tabularnewline}{\\}
\newcommand{\lyxdot}{.}

\floatstyle{ruled}
\newfloat{algorithm}{tbp}{loa}
\providecommand{\algorithmname}{Algorithm}
\floatname{algorithm}{\protect\algorithmname}

\usepackage{cvpr}
\usepackage{times}
\usepackage{epsfig}
\usepackage{graphicx}

\usepackage{subfigure}
\usepackage{adjustbox}
\usepackage{algorithm,algpseudocode}

\cvprfinalcopy 

\ifcvprfinal\fi

\newcommand{\bx}{\mathbf{x}}
\newcommand{\bz}{\mathbf{z}}
\newcommand{\bc}{\mathbf{c}}
\newcommand{\noti}{{\backslash}i}
\newcommand{\btheta}{\boldsymbol{\theta}}
\newcommand{\bomega}{\boldsymbol{\omega}}

\makeatother

\begin{document}

\title{Infinite Variational Autoencoder for Semi-Supervised Learning}

\author{Ehsan Abbasnejad \and Anthony Dick \and Anton van den Hengel \\
 The University of Adelaide \\
 \texttt{\small \{ehsan.abbasnejad, anthony.dick, anton.vandenhengel\}@adelaide.edu.au}} 

\maketitle

\begin{abstract}
This paper presents an infinite variational autoencoder (VAE) 
whose capacity
adapts to suit the input data. This is achieved using
a mixture model where the mixing coefficients are modeled by a Dirichlet
process, allowing us to integrate over the coefficients when performing
inference. Critically, this then allows us to automatically vary the
number of autoencoders in the mixture based on the data. Experiments
show the flexibility of our method, particularly for semi-supervised
learning, where only a small number of training samples are available. 
\end{abstract}

\section{Introduction}

The Variational Autoencoder (VAE) \cite{KingmaWelling2014} is a newly
introduced tool for unsupervised learning of a distribution $p(\bx)$
from which a set of training samples $\bx$ is drawn. It learns the
parameters of a generative model, based on sampling from a latent
variable space $\bz$, and approximating the distribution $p(\bx|\bz)$.
By designing the latent space to be easy to sample from (e.g. Gaussian)
and choosing a flexible generative model (e.g. a deep belief network)
a VAE can provide a flexible and efficient means of generative modeling.

One limitation of this model is that the dimension of the latent space
and the number of parameters in the generative model are fixed in
advance. This means that while the model parameters can be optimized
for the training data, the capacity of the model must be chosen a
priori, assuming some foreknowledge of the training data characteristics.




In this paper we present an approach that utilizes Bayesian non-parametric
models \cite{e2013,GelmanCarlinStern2014,OrbanzTeh2011,HjortHolmesMuellerEtAl2010} to produce an \emph{infinite} mixture of autoencoders. This
infinite mixture is capable of growing with the complexity of the
data to best capture its intrinsic structure.

Our motivation for this work is the task of semi-supervised learning.
In this setting, we have a large volume of unlabelled data but only
a small number of labelled training examples. In our approach, we
train a generative model using unlabelled data, and then use this
model combined with whatever labelled data is available to train a
discriminative model for classification.

We demonstrate that our infinite VAE outperforms both the classical
VAE and standard classification methods, particularly when the number
of available labelled samples is small. This is because the infinite
VAE is able to more accurately capture the distribution of the unlabelled
data. It therefore provides a generative model that allows the discriminative
model, which is trained based on its output, to be more effectively
learnt using a small number of samples.

The main contribution of this paper is twofold: (1) we provide a Bayesian
non-parametric model for combining autoencoders, in particular variational
autoencoders. This bridges the gap between non-parametric Bayesian
methods and the deep neural networks; (2) we provide a semi-supervised
learning approach that utilizes the infinite mixture of autoencoders
learned by our model for prediction with from a small number of labeled examples. 

The rest of the paper is organized as follows. In Section \ref{sec:related} we  review relevant methods, while in Section \ref{sec:Variational-Auto-encoder} we briefly provide
background on the variational autoencoder.
In Section \ref{sec:Infinite-Mixture-of} our non-parametric Bayesian
approach to infinite mixture of VAEs is introduced. We provide the
mathematical formulation of the problem and how the combination of
Gibbs sampling and Variational inference can be used for efficient
learning of the underlying structure of the input.
Subsequently in Section \ref{sec:Semi-Supervised-Learning-using},
we combine the infinite mixture of VAEs as an unsupervised generative
approach with discriminative deep models to perform prediction in
a semi-supervised setting. 
In Section \ref{sec:Experiments} we provide empirical evaluation
of our approach on various datasets including natural images and 3D
shapes. We use various discriminative models including Residual Network
\cite{HeZhangRenEtAl2015} in combination with our model and show
our approach is capable of outperforming our baselines. 

\section{Related Work}
\label{sec:related}

Most of the successful learning algorithms, specially with deep learning,
require large volume of labeled instance for training. Semi-supervised
learning seeks to utilize the unlabeled data to achieve strong generalization
by exploiting small labeled examples. For instance unlabeled data
from the web is used with label propagation in \cite{EbertFritzSchiele2013}
for classification. Similarly, semi supervised learning for object
detection in videos \cite{MisraShrivastavaHebert2015} or images \cite{WangHebert2015,FuSigal2016}. 

Most of these approaches are developed by either (a) performing a
projection of the unlabeled and labeled instances to an embedding
space and using nearest neighbors to utilize the distances to infer
the labeled similar to label propagation in shallow \cite{KangJinSukthankar2006,wang2009multi,InKimTompkinPfisterEtAl2015}
or deep networks \cite{WestonRatleMobahiEtAl2012}; or (b), formulating
some variation of a joint generative-discriminative model that uses
the latent structure of the unlabeled data to better learn the decision
function with labeled instances. For example ensemble methods \cite{ChenWang2008,MallapragadaJinJainEtAl2009,LeistnerSaffariSantnerEtAl2009,Zhou2011,DaiGool2013}
assigns pseudo-class labels based on the constructed ensemble learner
and in turn uses them to find a new proper learner to be added to
the ensemble.

In recent years, deep generative models have gained attention with
success in Restricted Boltzman machines (and its infinite variation
\cite{CoteLarochelle2015}) and autoencoders (e.g. \cite{KingmaMohamedRezendeEtAl2014,LarochelleMandelPascanuEtAl2012})
with their stacked variation \cite{VincentLarochelleLajoieEtAl2010}.
The representations learned from these unsupervised approaches are
used for supervised learning. 

Other related approaches to ours are adversarial networks \cite{GoodfellowPouget-AbadieMirzaEtAl2014,MiyatoMaedaKoyamaEtAl2016,MakhzaniShlensJaitlyEtAl2015}
in which the generative and discriminative model are trained jointly.
This model penalizes the generative model for as long as the samples
drawn from it does not perform well in the discriminative model in
a min-max optimization. Although theoretically well justified, training
such models proved to be difficult.

Our formulation for semi-supervised learning is also related to the
Highway \cite{SrivastavaGreffSchmidhuber2015} and Memory
\cite{WestonChopraBordes2014} networks that seek to combine multiple
channels of information that capture various aspects of the data for
better prediction, even though their approaches mainly focus on depth.

\section{Variational autoencoder\label{sec:Variational-Auto-encoder}}

While typically autoencoders assume a deterministic latent space,
in a variational autoencoder the latent variable is stochastic. The
input $\bx$ is generated from a variable in that latent space $\bz$.
Since the joint distribution of the input when all the latent variables
are integrated out is intractable, we resort to a variational inference
(hence the name). The model is defined as: 
\begin{eqnarray*}
p_{\btheta}(\bz) & = & \mathcal{N}(\bz;0,\mathbf{I}),\\
p_{\btheta}(\bx|\bz) & = & \mathcal{N}(\bx;\mu(\bz),\sigma(\bz)\mathbf{I}),\\
q_{\phi}(\bz|\bx) & = & \mathcal{N}(\bx;\mu(\bx),\sigma(\bx)\mathbf{I}),
\end{eqnarray*}
where $\btheta$ and $\boldsymbol{\phi}$ are the parameters of the
model to be found. The objective is then to minimize the following
loss,
\begin{eqnarray}
 & -\underbrace{\mathbb{E}_{\bz\sim q(\bz|\bx)}\left[\log p(\bx|\bz)\right]}_{\text{reconstruction error}}+\underbrace{\text{KL}\left(q_{\phi}(\bz|\bx)||p(\bz)\right)}_{\text{regularization}}.\label{eq:vae_loss}
\end{eqnarray}

\begin{figure}
\centering{}\includegraphics[scale=0.5]{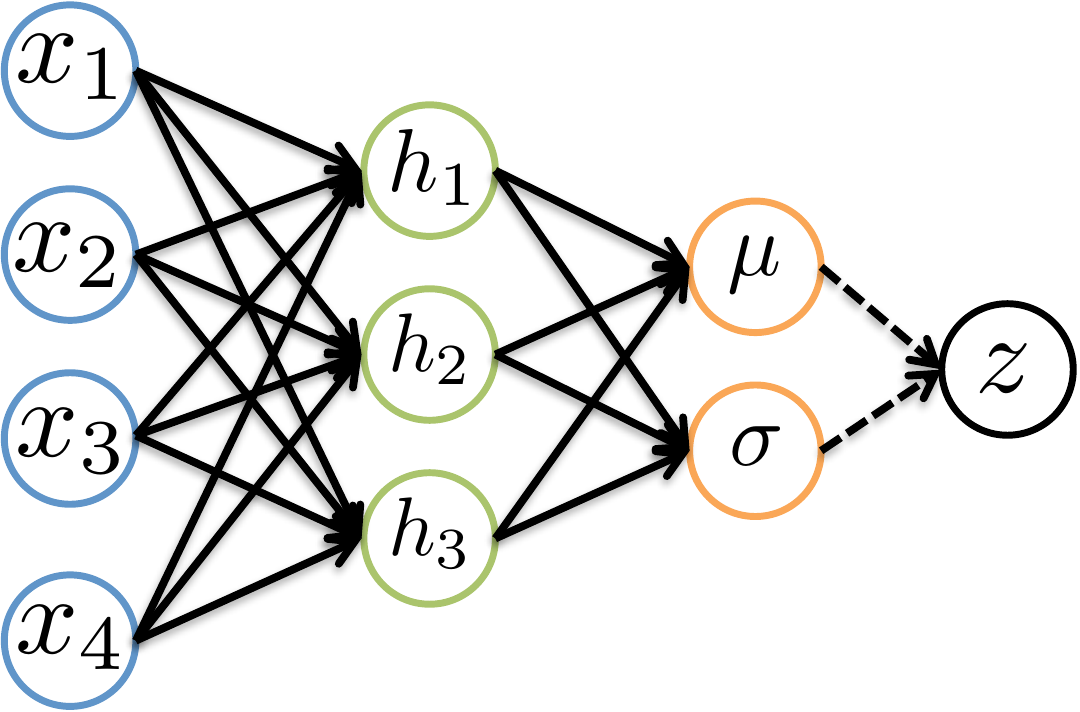}\caption{Variational encoder: the solid lines are direct connection and dotted
lines are sampled. The input layer represented by $\bx$ and the hidden
layer $\mathbf{h}$ determine moments of the variational distribution. From
the variational distribution the latent variable $\bz$ is sampled.}
\label{fig:var_encoder} 
\end{figure}

The first term in this loss is the reconstruction error, or expected
negative log-likelihood of the datapoint. The expectation is taken
with respect to the encoder's distribution over the representations
by taking a few samples. This term encourages the decoder to learn
to reconstruct the data when using samples from the latent distribution.
A large error indicates the decoder is unable to reconstruct the
data. A schematic network of the encoder is shown in Figure \ref{fig:var_encoder}.
As shown, deep network learns the mean and variance of a Gaussian
from which subsequent samples of $\bz$ are generated. 

The second term is the Kullback-Leibler divergence between the encoder's distribution
$q_{\theta}(\bz|\bx)$ and $p(\bz)$. This divergence measures how
much information is lost when using $q$ to represent a prior over
$\bz$ and encourages its values to be Gaussian. To perform
inference efficiently a reparameterization trick is employed~\cite{KingmaWelling2014} that
in combination with the deep neural networks allow for the model to
be trained with the backpropagation.

\section{Infinite Mixture of Variational autoencoder\label{sec:Infinite-Mixture-of}}

\begin{figure}
\begin{centering}
\includegraphics[scale=0.5]{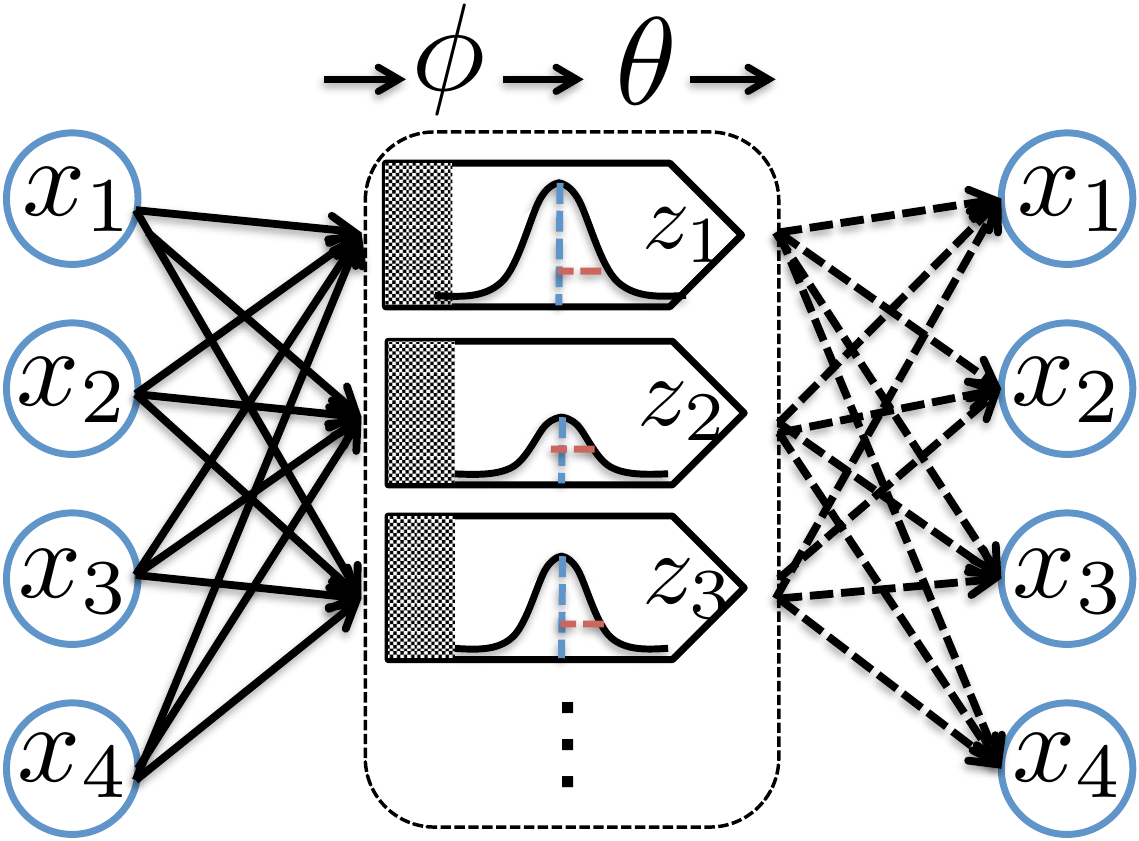} 
\par\end{centering}
\caption{Infinite mixture of variational inference is shown as a block within
which VAE components operate. Each latent variable $z_{i}$ (one dimensional
in this illustration) in each VAE is drawn from a Gaussian distribution.
Solid lines indicate nonlinear encoding and the dashed lines are decoders.
In this diagram, $\boldsymbol{\phi}$ and $\btheta$ are the parameters
of the encoder and decoder respectively.}
\label{fig:Infinite-mixture-of-VAE}
\end{figure}

An auto encoder in its classical form seeks to find an embedding
of the input such that its reproduction has the least discrepancy.
A variational autoencoder modifies this notion by introducing a \emph{non-parametric
Bayesian} view where the conditional distribution of the latent variables,
given the input, is similar to the distribution of the input given
the latent variable, while ensuring the distribution of the latent variable
is close to a Gaussian with zero mean and variance one.

A single variational encoder has a fixed capacity and thus might not be able to capture the complexity
of the input well. However by using a collection of VAEs,  we can ensure that we are able to model the data, by adapting the number of VAEs in the collection to fit the data. In our \emph{infinite mixture}, we seek to find a mixture of
these variational autoencoders such that its capacity can theoretically
grow to infinity. Each autoencoder then is able to capture a
particular aspect of the data. For instance, one might be better at
representing round structures, and another better at straight lines.
This mixture intuitively represents the various underlying aspects
of the data. Moreover, since each VAE models the \emph{uncertainty}
of its representations through the density of the latent variable,
we know how confident each autoencoder is in reconstructing the input.

One advantage of our non-parametric mixture model
is that we are taking a Bayesian approach in which the distribution
of the parameters are taken into account. As such, we capture the
uncertainty of the model parameters. The autoencoders that are
less confident about their reconstruction, have less effect on the output. As
shown in Figure \ref{fig:Infinite-mixture-of-VAE}, each encoder finds
a distribution for the embedding variable with some probability through
a nonlinear transform (convolution or fully connected layers in neural
net). Each autoencoder in the mixture block produces a probability
measure for its ability to reconstruct the input. 
This behavior has parallels
to the brain's ability to develop specialized regions responsible
for particular visual tasks and processing particular types of image pattern.

Mixture models are traditionally built using a pre-determined number
of weighted components. Each weight coefficient determines how likely it is for
a predictor to be successful in producing an accurate output. These
coefficients are drawn from a multinomial distribution where the number
of these coefficients are fixed. On the other hand, to learn an infinite
mixture of the variational autoencoders in a non-parametric Bayesian
manner we employ \emph{Dirichlet process}. In Dirichlet process, unlike
traditional mixture models, we assume the probability of each component
is drawn from a multinomial with a Dirichlet prior. The
advantage of taking this approach is that we can integrate over all
possible mixing coefficients. This allows for the number of components
to be determined based on the data.

\begin{algorithm}
\begin{algorithmic}
	\State Initalize VAE assignments $\bc$
	\State $A_c = \{\}$ $\quad \forall c = 1,\ldots,C$
	\While {not converged}
		\For {$\bx_i \in X$}
			\Comment{VAE assignments}
			\State Assign $\bc_i^\text{new}$ to new VAE according to Eq.~\ref{eq:new_label_prob}
			\State Otherwise, sample $\bc_i^\text{new}$ according to Eq.~\ref{eq:current_label_prob}
			\If {$\bc_i^\text{new}\neq \bc_i$}
				\State $A_{\bc_i} = A_{\bc_i} \cup \{i\}$ \Comment{Given VAE has to forget}
			\EndIf
		\EndFor
		\State Update $C$ for new VAEs
		\For {$c=1,\ldots,C$}
			\Comment{Update VAEs}
			\State Forget $A_c$ in $c$th VAE
			\State Learn $c$th VAE $\quad \forall i $ where $\bc_i^\text{new}=c$
		\EndFor
	\EndWhile
		\State Return \emph{Infinite Mixture of VAEs}
\end{algorithmic}\caption{Learning Infinite mixture of Variational autoencoders}
\label{alg:alg1} 
\end{algorithm}

Formally, let $\bc$ be the assignment matrix for each instance to
a VAE component (that is, which VAE is able to best reconstruct instance
$i$) and $\boldsymbol{\pi}$ be the mixing coefficient
prior for $\bc$. For $n$ unlabeled instances we model the infinite
mixture of VAEs as,
{\small
\begin{eqnarray*}
p(\bc, \boldsymbol{\pi},\btheta,\bx_{1,\ldots,n},\alpha)\!\!\!\! & = &\!\!\!\! p(\bc|\boldsymbol{\pi})p(\boldsymbol{\pi}|\alpha) \int p_{\btheta}(\bx_{1,\ldots,n}|\bc,\bz)p(\bz)d\bz
\end{eqnarray*}
}

\begin{figure}[t]
\begin{centering}
\subfigure[$\alpha=0.99$]{\includegraphics[scale=0.14]{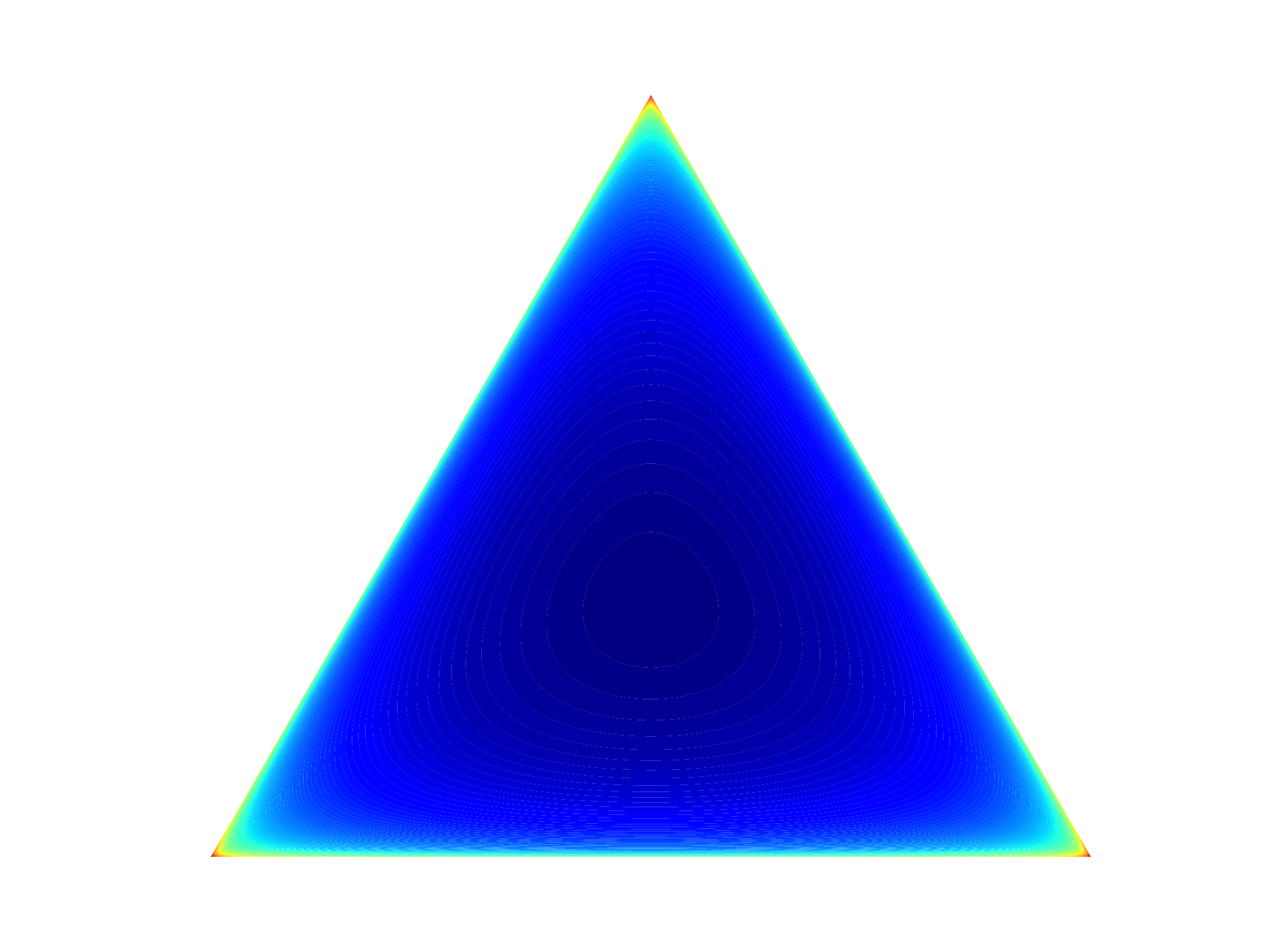}}\subfigure[$\alpha=2$]{\includegraphics[scale=0.14]{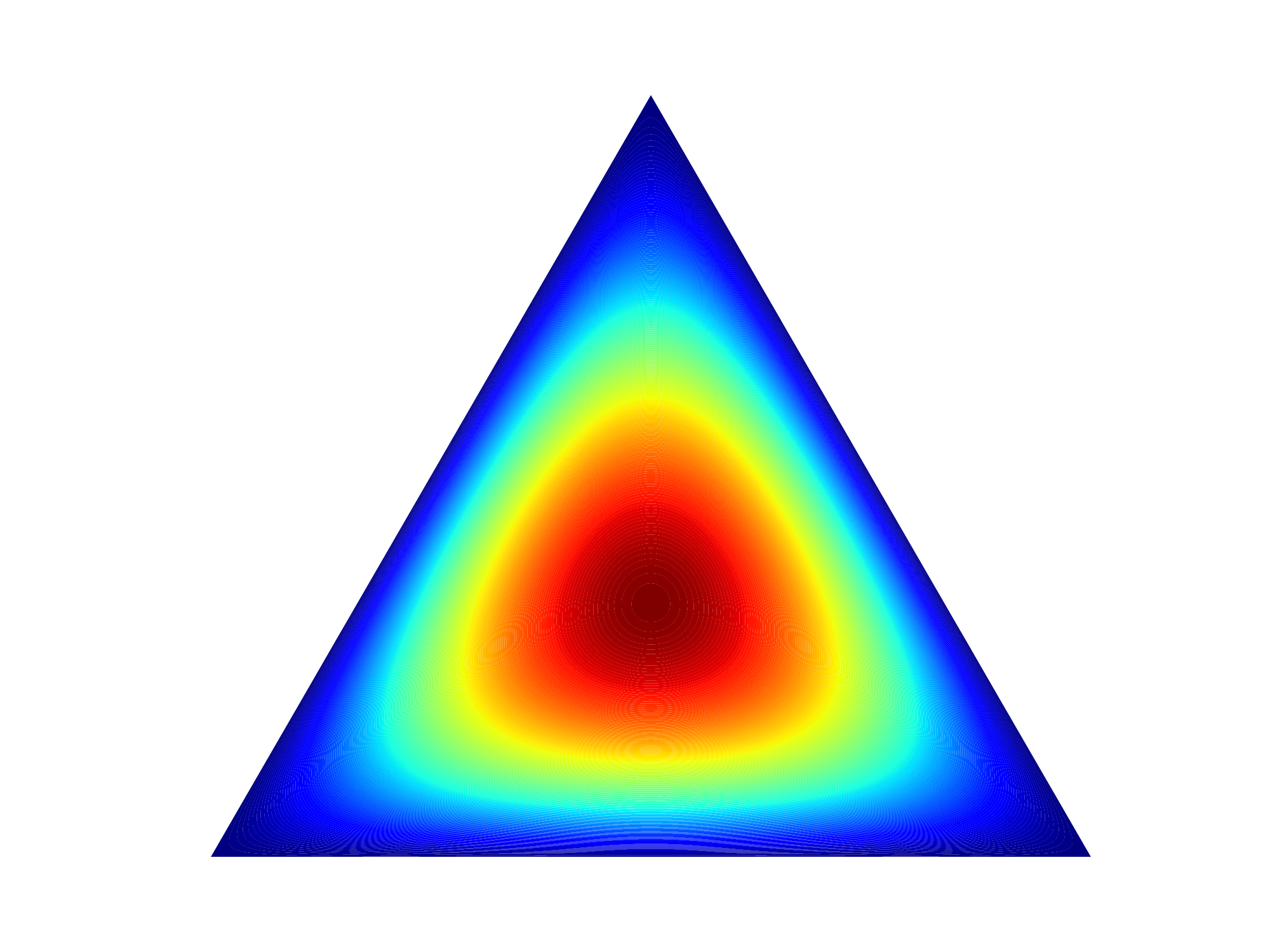}}\subfigure[$\alpha=50$]{\includegraphics[scale=0.14]{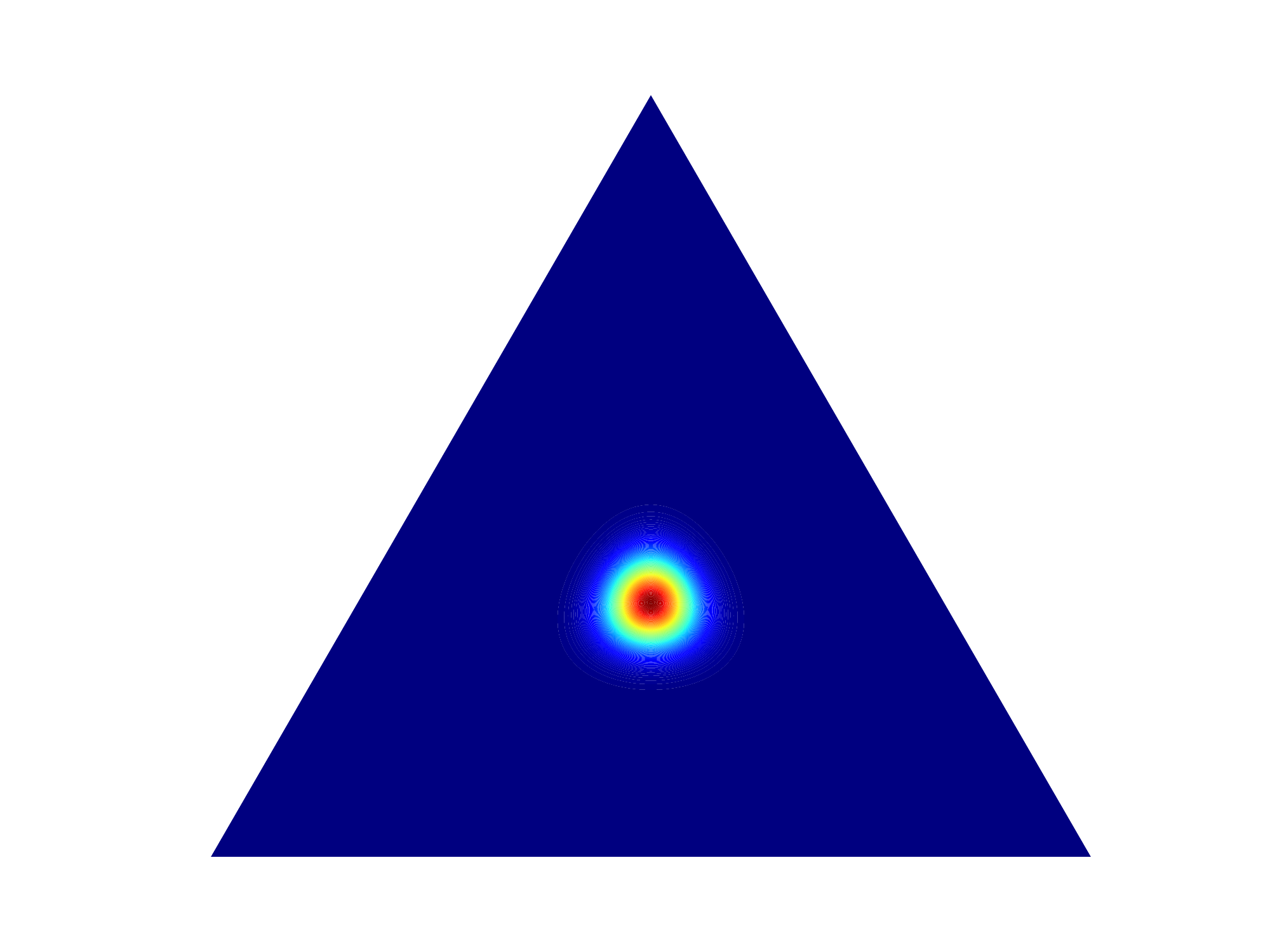}} 
\par\end{centering}
\caption{Dirichlet distribution with various values of $\alpha$. Smaller values
of $\alpha$ tend to concentrate the mass in the corners (in this
simplex example and in general as the dimensions increase). These
smaller values reduce the chance of generating new autoencoder components. }
\label{fig:dir_alpha} 
\end{figure}

We assume the mixing coefficients are drawn from a Dirichlet distribution
with parameter $\alpha$ (see Figure \ref{fig:dir_alpha} for examples),
\begin{eqnarray*}
p(\pi_{1},\ldots,\pi_{C}|\alpha) & \sim & \text{Dir}(\alpha/C),
\end{eqnarray*}
To determine the membership of each instance in one of the components
of the mixture model, i.e. the likelihood that each variational autoencoder
is able to encode the input and reconstruct it with minimum loss,
we compute the conditional probability of membership. This conditional
probability of each instance belonging to an autoencoder component
is computed by integrating over all mixing components $\boldsymbol{\pi}$,
that is \cite{Rasmussen2000,RasmussenGhahramani2002}, 
{\small
\begin{eqnarray*}
p(\bc,\btheta,\bx_{1,\ldots,n},\alpha) \!\!\!\!\!\!& = & \!\!\!\!\!\!
\int\!\!\int\prod_{i}^{n}p{}_{\btheta_{\bc_{i}}}(\bx_{i}|\bz_{\bc_{i}})p(\bz_{\bc_{i}})p(\bc|\boldsymbol{\pi})p(\boldsymbol{\pi}|\alpha)d\boldsymbol{\pi}d\bz_{\bc_{i}}
\end{eqnarray*}
}
This integration accounts for \emph{all possible} membership coefficients
for all the assignments of the instances to VAEs. The distribution
of $\bc$ is multinomial, for which the Dirichlet distribution is its conjugate
prior, and as such this integration is tractable. To perform inference
for the parameters $\btheta$ and $\bc$ we perform block Gibbs sampling,
iterating between optimizing for $\btheta$ 
for each VAE and updating the assignments in $\bc$. Optimization uses the variational autoencoder's trick by minimizing the loss in Equation \ref{eq:vae_loss}. To update
$\bc$, we perform the following Gibbs sampling:
\begin{itemize}
\item The conditional probability that an instance $i$ belongs to VAE $c$:
\begin{eqnarray}
p(\bc_{i}=c|\bc_{\noti},\bx_{i},\alpha) & = & \frac{\eta_{c}(\bx_{i})}{n-1+\alpha}\label{eq:current_label_prob}
\end{eqnarray}
where $\eta_{c}(\bx_{i})$ is the \emph{occupation number} of cluster
$c$, excluding instance $i$ for $n$ instances. We define, 
\begin{eqnarray*}
\eta_{c}(\bx_{i}) & = & (n-1)p_{\btheta_{c}}(\bc_{i}=c|\bx_{i}),
\end{eqnarray*}
and {\footnotesize
\begin{eqnarray*}
p_{\btheta_{c}}(\bc_{i}=c|\bx_{i})\!\!\!\!\! & =\!\!\!\!\! & \frac{\exp\left(\mathbb{E}_{\bz_{c}\sim q_{\phi_{c}(\bz|\bx)}}\left[\log p_{\btheta_{c}}(\bx_{i}|\bz_{c})\right]\right)}{\sum_{j}\exp\left(\mathbb{E}_{\bz_{j}\sim q_{\phi_{j}(\bz|\bx)}}\left[\log p_{\btheta_{j}}(\bx_{i}|\bz_{j})\right]\right)}\label{eq:label_assign}
\end{eqnarray*}}
which in evaluates how likely an instance $\bx_{i}$
is to be assigned to the $c$th VAE using latent
samples $\bz_{c}$. 
\item The probability that instance $i$ is not well represented by any
of the existing autoencoders and a new encoder has to be generated: 
\end{itemize}
\begin{eqnarray}
p(\bc_{i}=c|\bc_{\noti},\bx_{i},\alpha) & = & \frac{\alpha}{n-1+\alpha}.\label{eq:new_label_prob}
\end{eqnarray}

Note that in principle, $\eta_{c}(\bx_{i})$ is the a measure calculated
by excluding the $i$th instance in the observations so that its membership
is calculated with respect to its \char`\"{}similarity\char`\"{} to
other members of the cluster. However, here we use $c$th VAE as an
estimate of this occupation number for performance reasons. This is justified
so long as the influence of a single observation on the
latent representation of an encoder is negligible. In Equation \ref{eq:current_label_prob}
when a sample for the new assignment is drawn from this multinomial
distribution there is a chance for completely different VAE to fit
this new instance. If the new VAE is not successful in fitting, the
instance will be assigned to its original VAE with high probability
in the subsequent iteration.

The entire learning process is summarised in Algorithm \ref{alg:alg1}. To improve performance, at each iteration of our approach, we keep track of the $c$th VAE assignment changes in a set $A_{c}$. This allows us to efficiently update each
VAE using a backpropagation operation for the new assignments. We
perform two operations after VAE assignments are done: (1) \emph{forget},
and (2) \emph{learn}. In forgetting stage, we tend to unlearn the
instances that were assigned to the given VAE. It is done by performing
a gradient update with negative learning-rate, i.e. \emph{reverse
backpropagation}. In the learning stage on the other hand, we update
the parameters of the given VAE with positive learning-rate, as is
commonly done using backpropagation. This alternation allows for structurally
similar instances that can share latent variables to be learned with
a single VAE, while forgetting those that are not well suited.

To reconstruct an input $\bx$ with an infinite mixture, the expected
reconstruction is defined as: 
\begin{equation}
\mathbb{E}[\bx]=\sum_{c}p_{\btheta_{c}}(\bc_{i}=c|\bx_{i})\mathbb{E}_{q_{\phi}(\bz_{c}|\bx)}\left[\bx|\bz_c\right].\label{eq:expected_x}
\end{equation}
That is, we use each VAE to reconstruct the input and weight it with
the probability of that VAE (this probability is inversely proportionate
to the variance of each VAE).

\section{Semi-Supervised Learning using Infinite autoencoders\label{sec:Semi-Supervised-Learning-using}}

Many of deep neural networks' greatest successes have been in supervised learning, which depends on the availability of large labeled
datasets. However, in many problems such datasets are unavailable
and alternative approaches, such as combination of generative and discriminative
models, have to be employed. In semi-supervised learning, where the
number of labeled instances is small, we employ our infinite mixture
of VAEs to assist supervised learning. Inspired by the \emph{mixture
of experts} \cite[Chapter 11]{Murphy2012} we formulate the problem of predicting
output $y^{*}$ for the test example $\bx^{*}$ as,
\begin{eqnarray*}
p(y^{*}|\bx^{*}) & = & \sum_{c}^{C}\underbrace{p(y^{*}|\bx^{*},\bomega_{c})}_{\text{deep discriminative}}\times\,\,\underbrace{p_{\btheta_{c}}(\bc_{i}=c|\bx_{i})}_{\text{deep generative}}.
\end{eqnarray*}
This formulation for prediction combines the discriminative power
of a deep learner 
with parameter set $\bomega_{c}$, and a flexible generative model. For
a given test instance $\bx^{*}$, each discriminative expert produces
a tentative output that is then weighted by the generative model.
As such, each discriminative expert learns to perform better with
instances that are more structurally similar from the generative model's
perspective.

\begin{figure}
\includegraphics[scale=0.5]{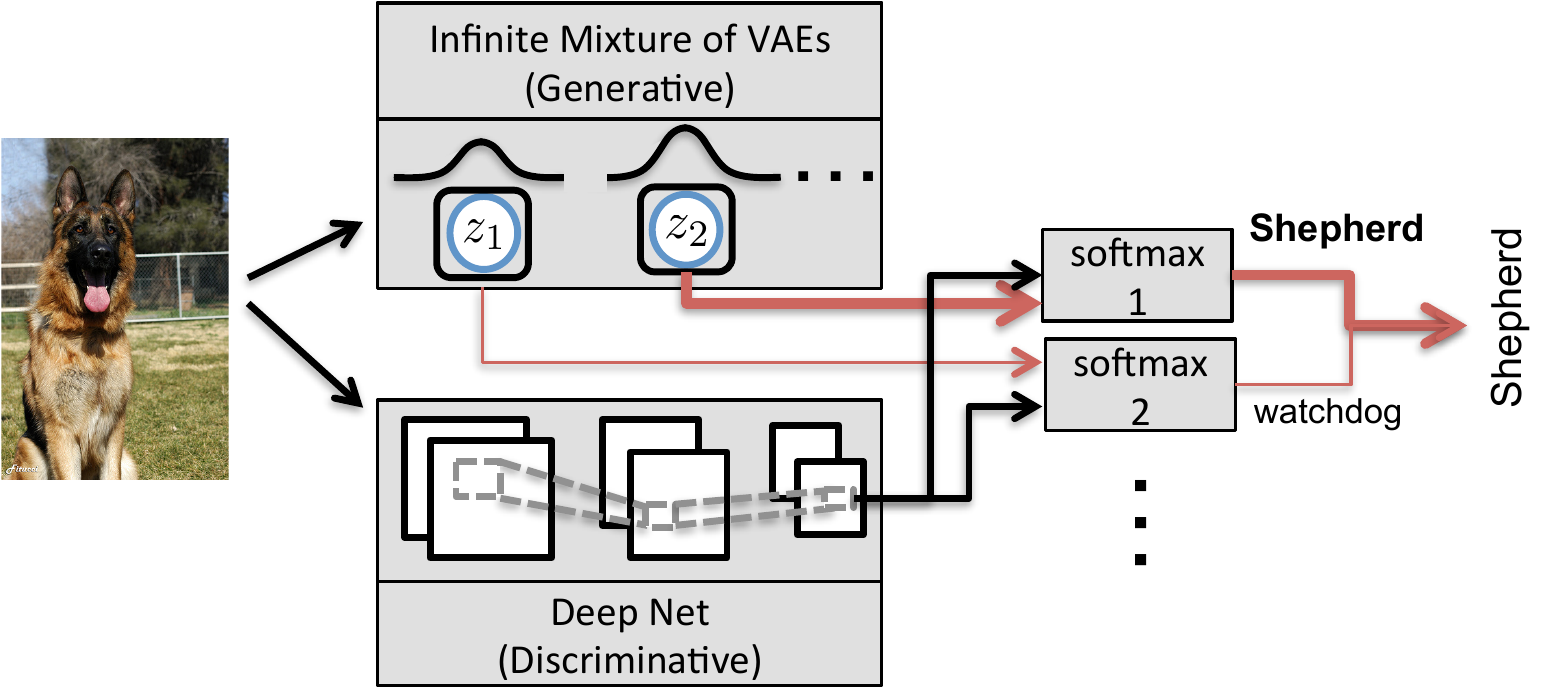}

\caption{Our framework for infinite mixture of VAEs and semi-supervised learning.
We share the parameters of the discriminative model at the lower levels
for more efficient training and prediction. For each VAE in the mixture
we have an expert (e.g. softmax) before the output. Thicker arrows
indicate more probable connection. }
\label{fig:complete_model} 
\end{figure}

During training we minimize the negative log of the discriminative
term (log loss) weighted by the generative weight. Each instance's
weight\textendash as calculated by the infinite autoencoder\textendash acts
as an additional coefficient for the gradient in the backpropagation.
It leads to similar instances getting stronger weights in the neural
net during training. Moreover, it should be noted that the generative
and discriminative models can share deep parameters $\bomega_{c}$
and $\btheta_{c}$ at some level. In particular in our implementation,
we only consider parameters of the last layer to be distinct for each
discriminative and generative component. We summarize our framework
in Figure \ref{fig:complete_model}.

While combining an unsupervised generative model and a supervised
discriminative models is not itself novel, in our problem the
generative model can grow to capture the complexity of the data. In
addition, since we share the parameters of the discriminative and
generative models, each unsupervised learner does not need to learn
all the aspects of the input. In fact, in many classification problems
with images, each pixel value hardly matters in the final decision.
As such, by sharing parameters unsupervised model incurs a heavier
loss when the distribution of the latent variables does not encourage the
correct final decision. This sharing is done by reusing the parameters
that are initialized with labels. 

\section{Experiments \label{sec:Experiments}}

In this section, we examine the performance of our approach for semi-supervised classification on
various datasets. We investigate how the combination
of the generative and discriminative networks is able to perform semi-supervised
learning effectively. Since convergence of Gibbs sampling can be very slow we first pre-train the base VAE with all the unlabeled
examples. Each autoencoder is trained with a two dimensional latent
variable $\bz$ and initialized randomly. Hence each new VAE is already
capable of reconstructing the input to a certain extent. During the
sampling steps, this VAE becomes more specialized in a particular
structure of the inputs. To further facilitate sampling, we set the
number of clusters equal to the number of classes and use $100$ random
labeled examples to fine-tune VAE assignments. At each iteration,
if there is no instance assigned to a VAE, it will be removed. As
such, the mixture grows and shrinks with each iteration as instances
are assigned to VAEs. We report the results over 3 trials.

For comparing the autoencoder's ability to internally capture the
structure of the input, we compared latent representation obtained
by a single VAE and the expected latent representation from our approach
in Equation \ref{eq:expected_x} and subsequently trained a support
vector machine (SVM) with it. For computing expectations, we used
$20$ samples from the latent variable space.

Once the generative model is learned with all the unlabelled instances
using the infinite mixture model in Section \ref{sec:Infinite-Mixture-of},
we randomly select a subset of labeled instances for training the
discriminative model. Throughout the experiments, we share the parameters
in the discriminative architecture from the input to the last layer
so that each expert is represented by a softmax. 

We report classification results in various problems including handwritten
binary images, natural images and 3D shapes. Although the performance
of our semi-supervised learning approach depends on the choice of
the discriminative model, we observe our approach outperforms baselines
particularly with smaller labeled instances. For all trainings\textendash either
discriminative or generative\textendash we set the maximum number
of iterations to $1000$ with batch size $500$ for the stochastic
gradient descent with constant learning rate $0.001$. For VAEs we
use the Adam \cite{KingmaBa2014} updates with $\beta_{1}=0.9$, $\beta_{2}$$=0.999$.
However, we set a threshold on the changes in the loss to detect convergence
and stop the training. Except for the binary images where we use a
binary decoder ($p_{\btheta}(\bx|\bz)$ is binomial), our decoder
is continuous (($p_{\btheta}(\bx|\bz)$ is Gaussian) in which samples
from the latent space is used to regenerate the input to compute the
loss.

In problems when the input is too complex for the autoencoder to perform
well, we share the output of the last layer of the discriminative
model with the VAEs.

\subsection{MNIST Dataset}

\begin{figure*}
\begin{tabular}{l}
\vspace{-2mm}
\subfigure[Original Images]{\label{fig:mnist_reconst_1}\includegraphics[scale=0.1]{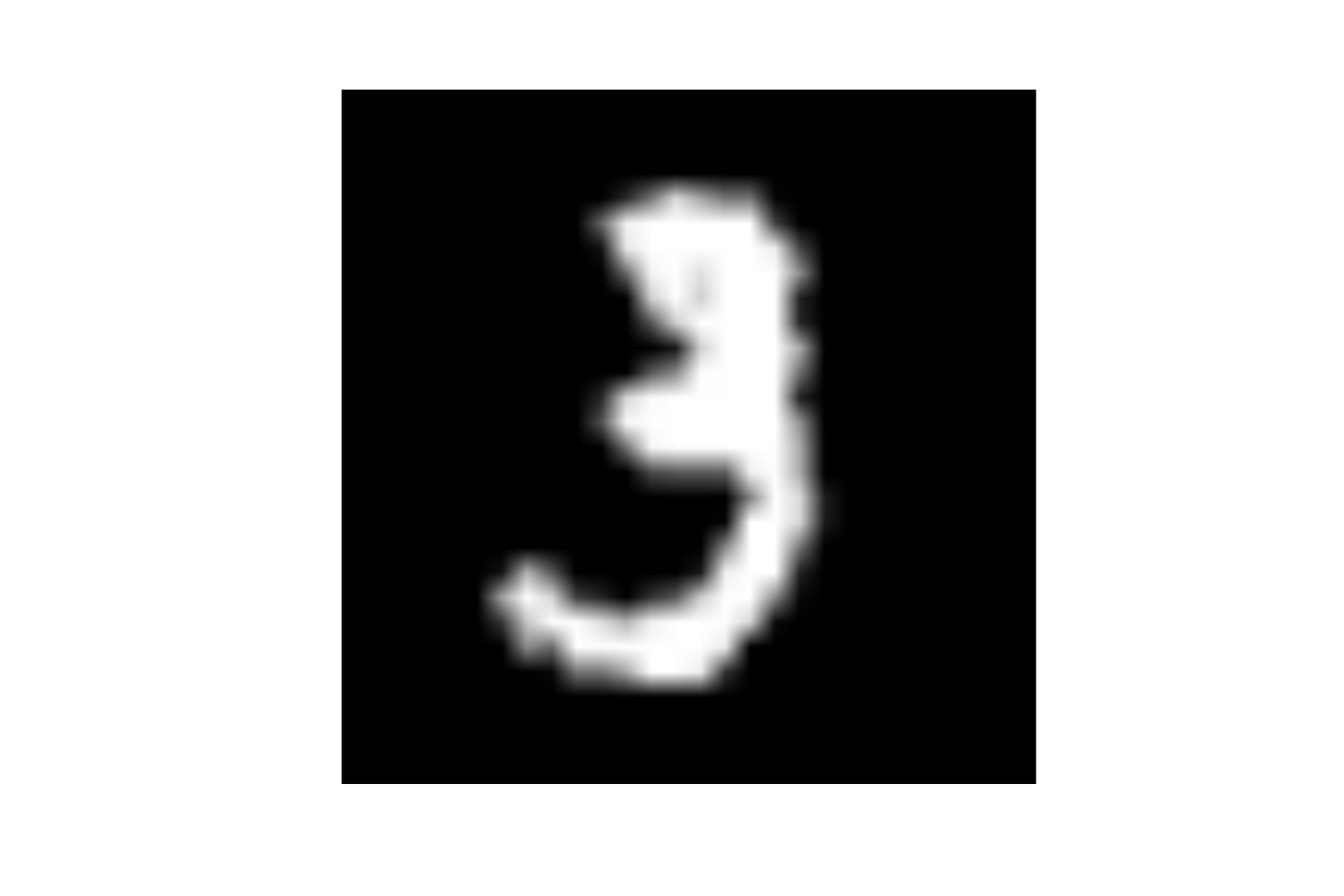}
 \hspace{-5mm}\includegraphics[scale=0.1]{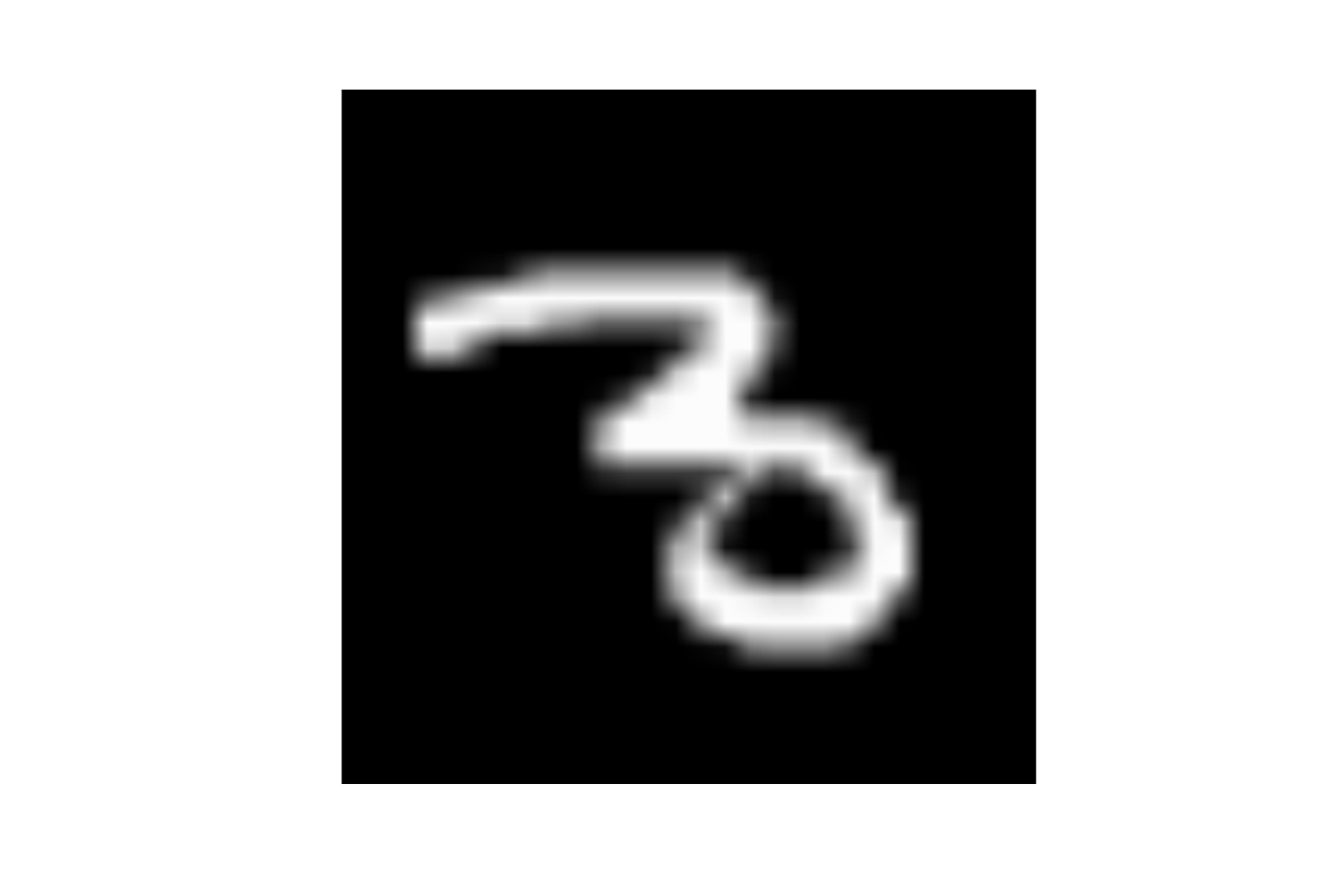}
 \hspace{-5mm}\includegraphics[scale=0.1]{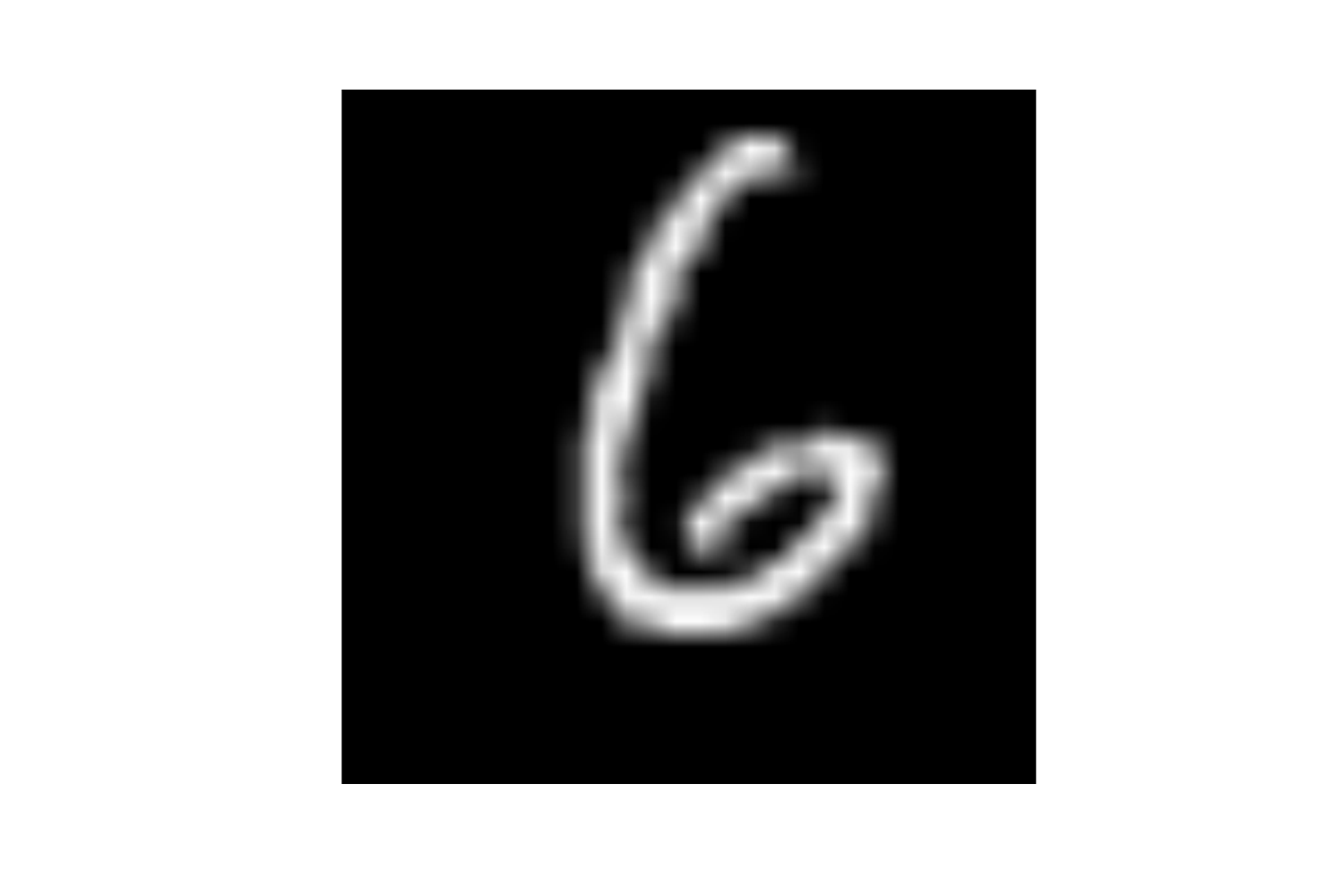}
 \hspace{-5mm}\includegraphics[scale=0.1]{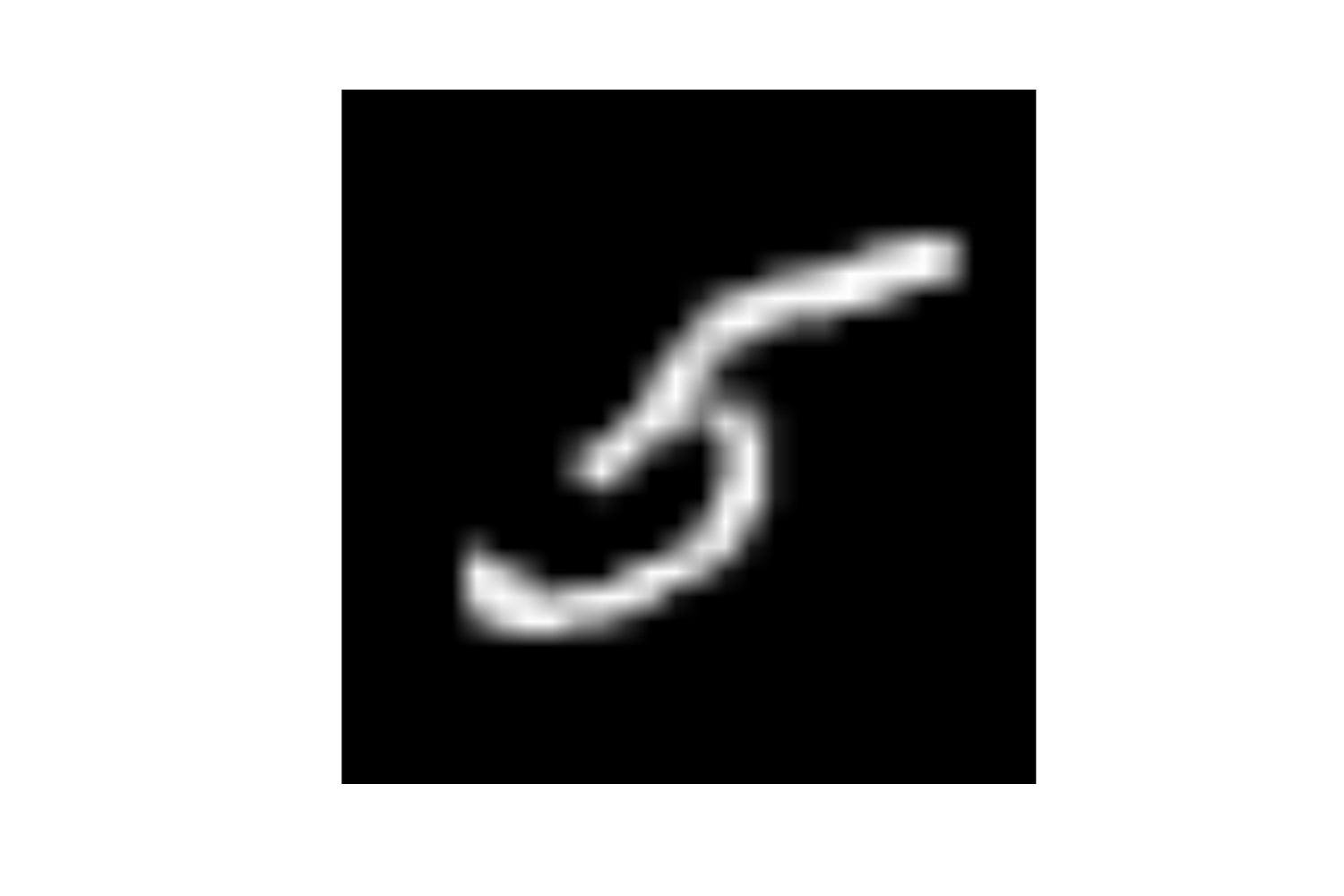}
 \hspace{-5mm}\includegraphics[scale=0.1]{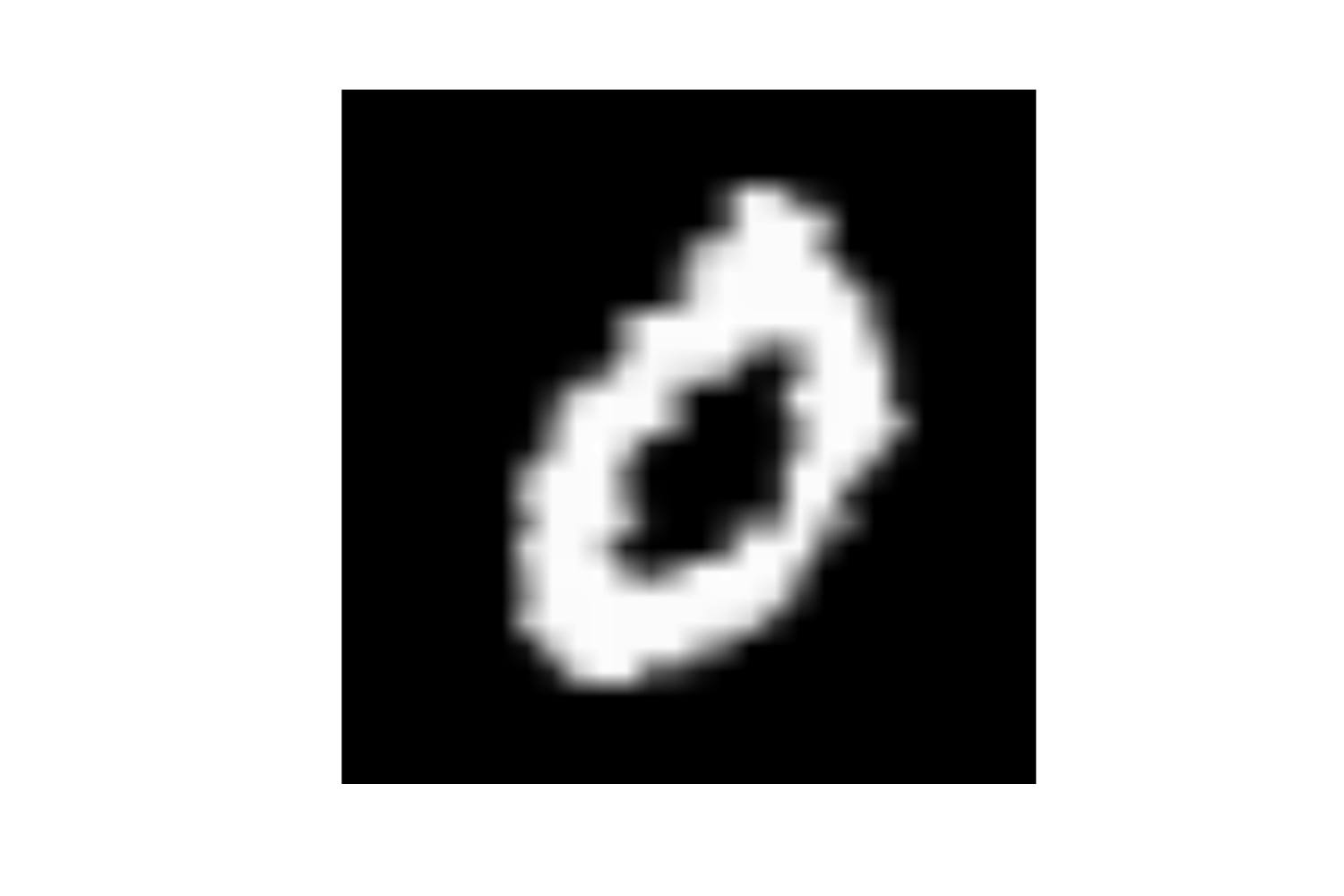}
 \hspace{-5mm}\includegraphics[scale=0.1]{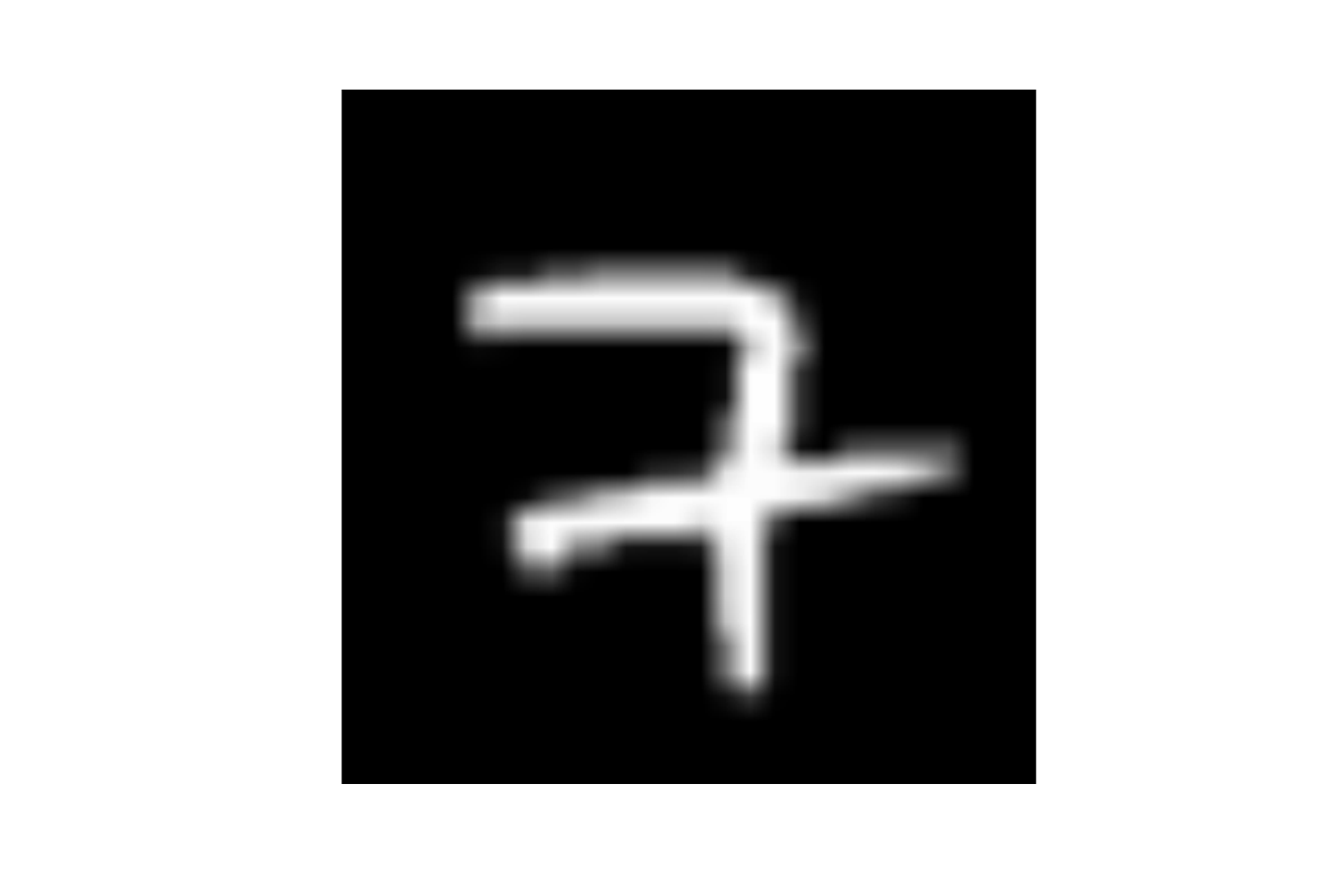}
 \hspace{-5mm}\includegraphics[scale=0.1]{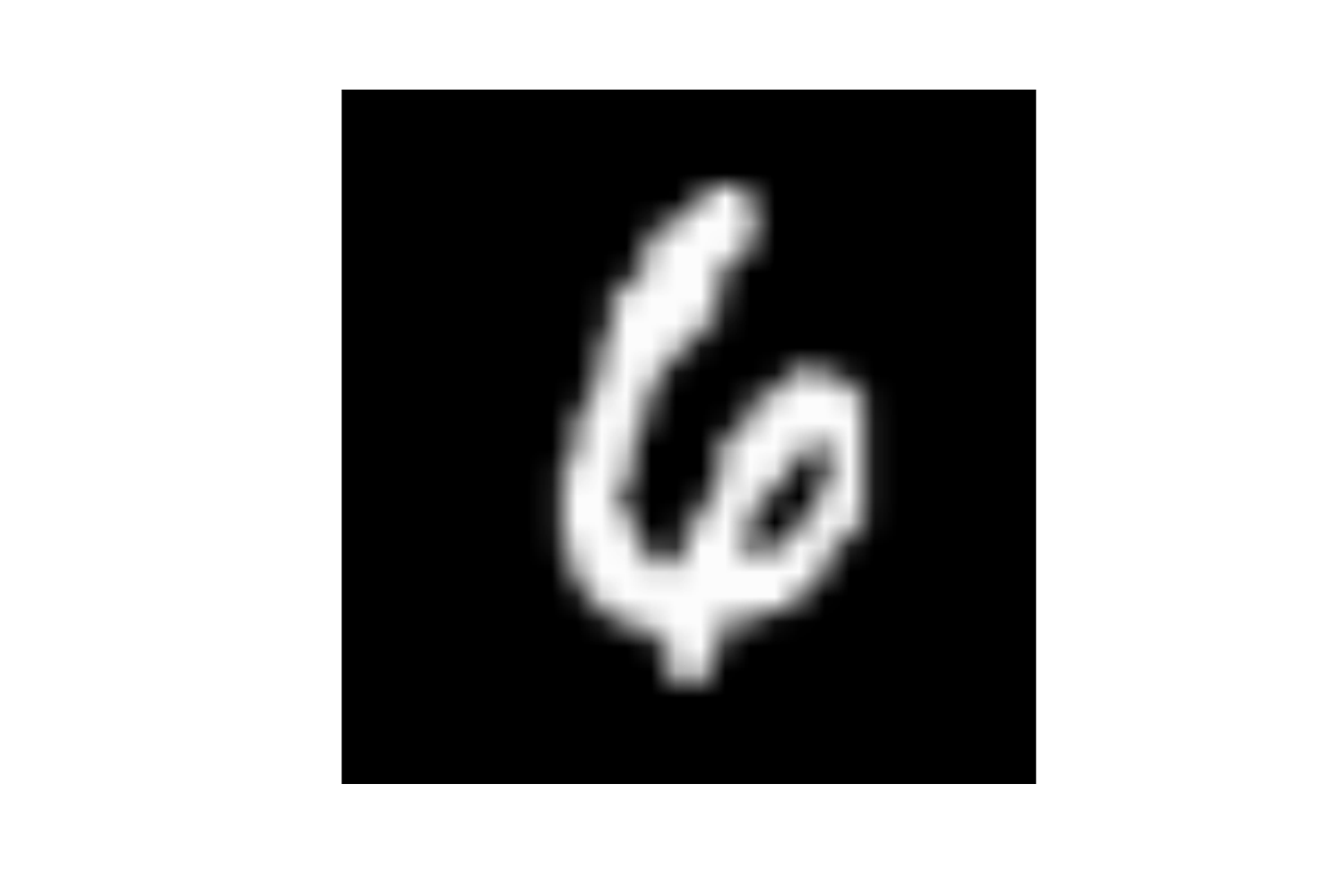}
 \hspace{-5mm}\includegraphics[scale=0.1]{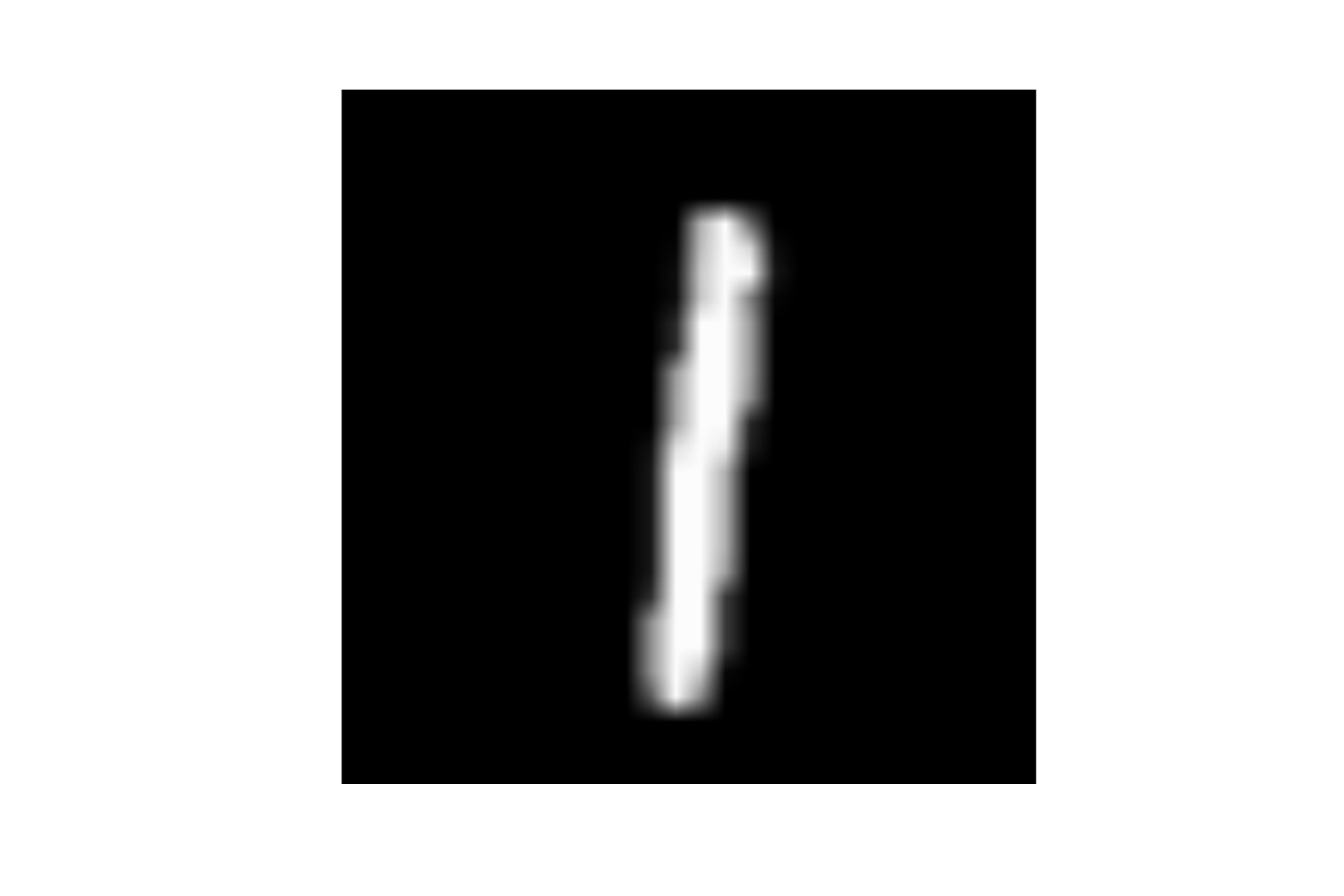}
 \hspace{-5mm}\includegraphics[scale=0.1]{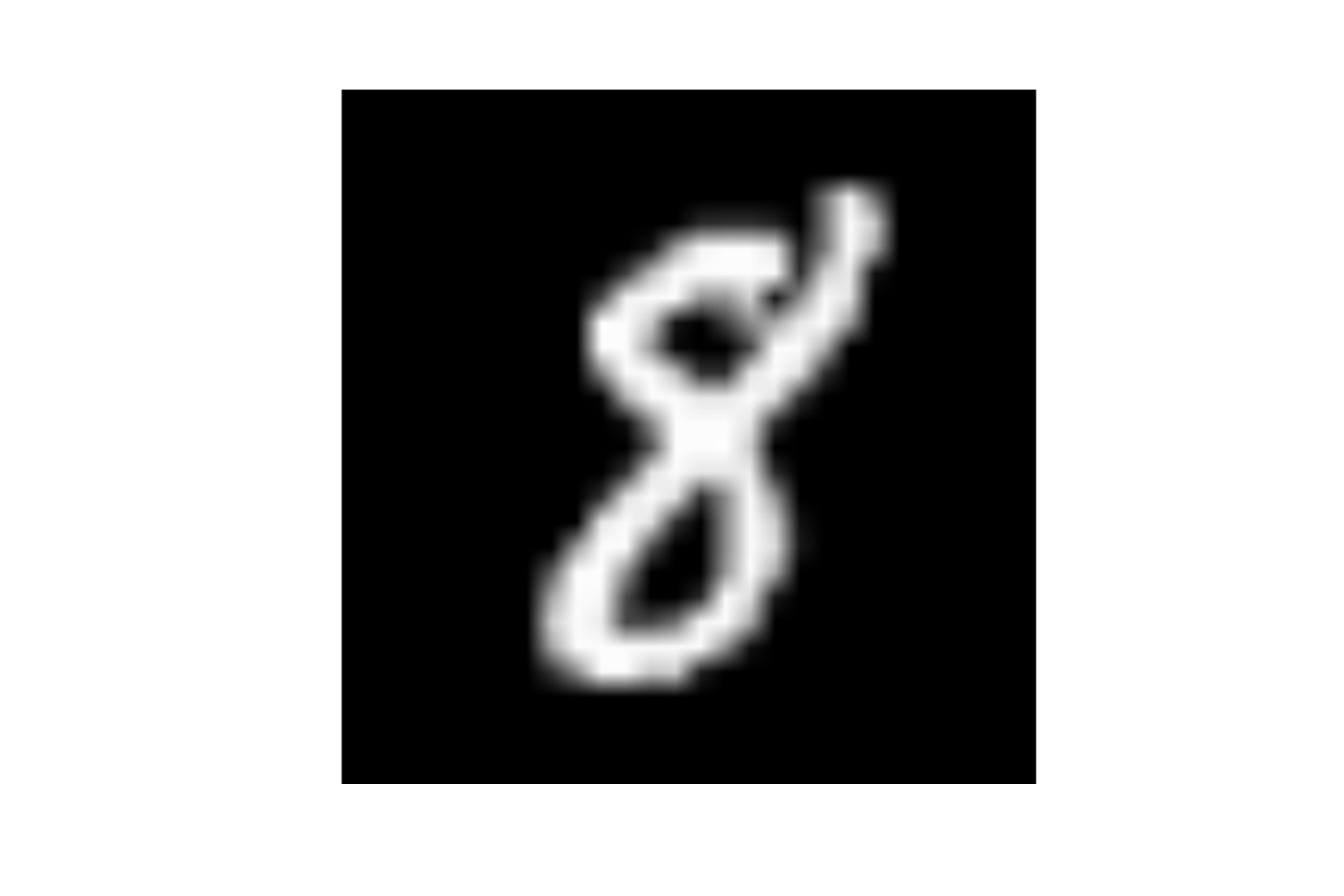}
 \hspace{-5mm}\includegraphics[scale=0.1]{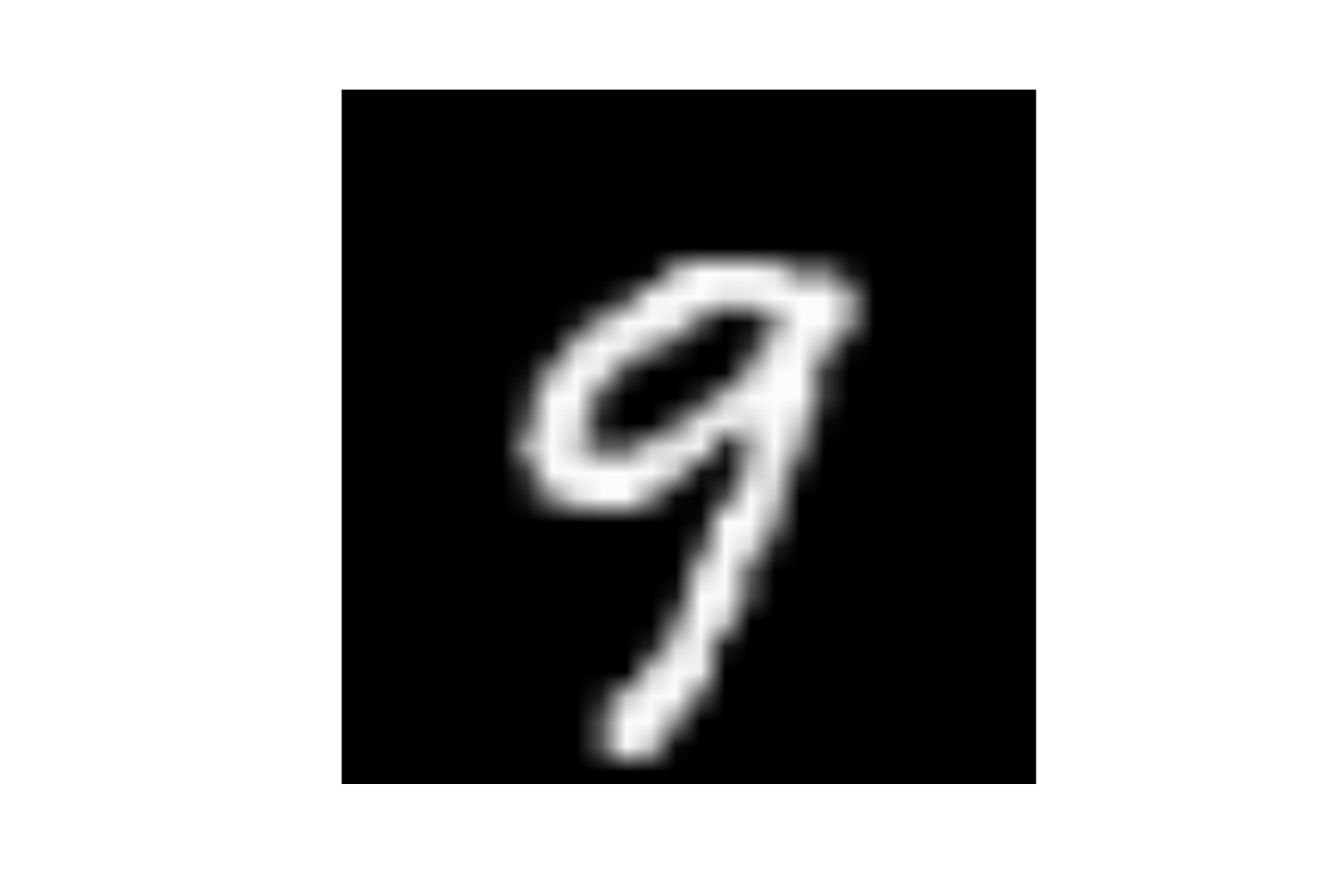}
 \hspace{-5mm}\includegraphics[scale=0.1]{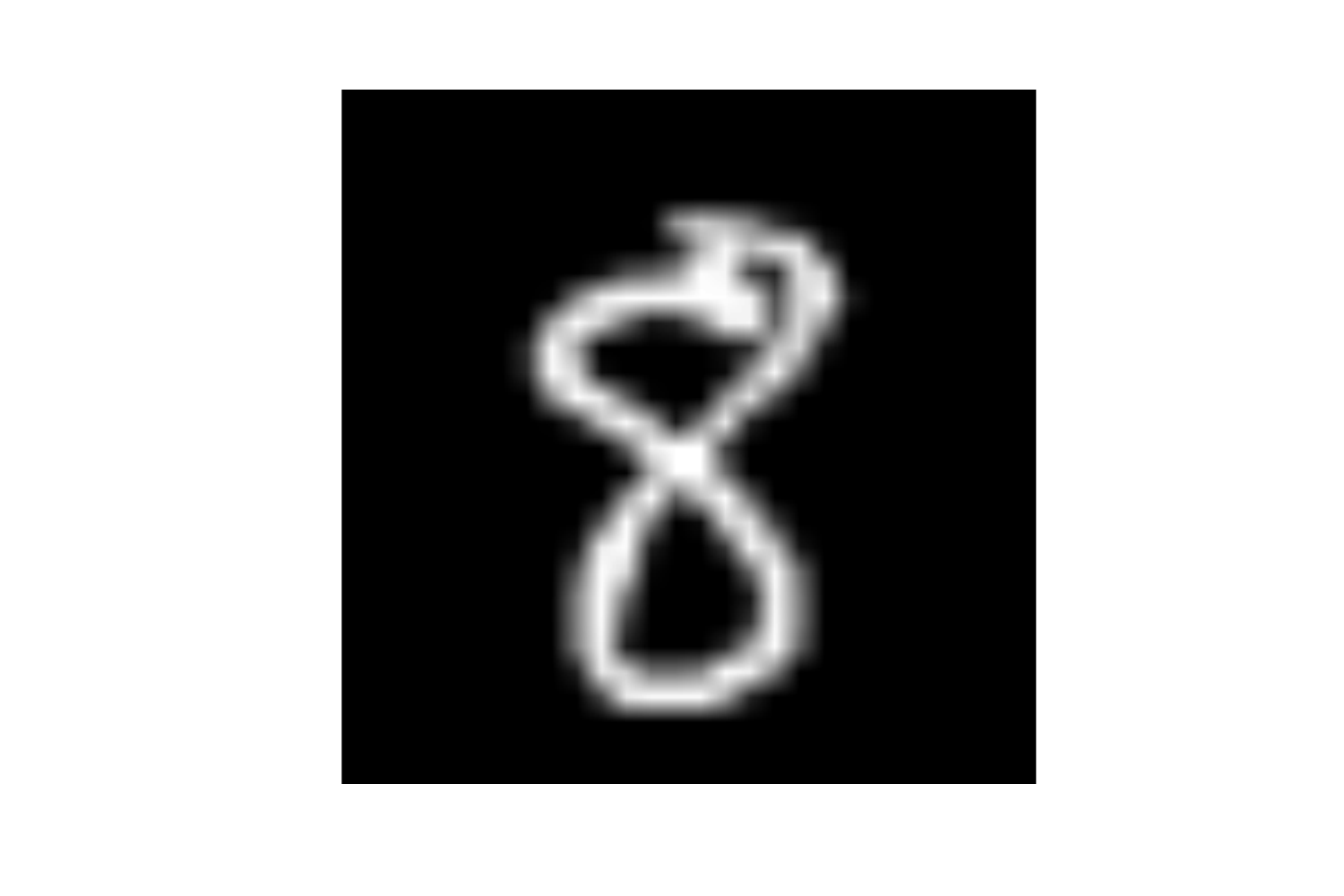}
 \hspace{-5mm}\includegraphics[scale=0.1]{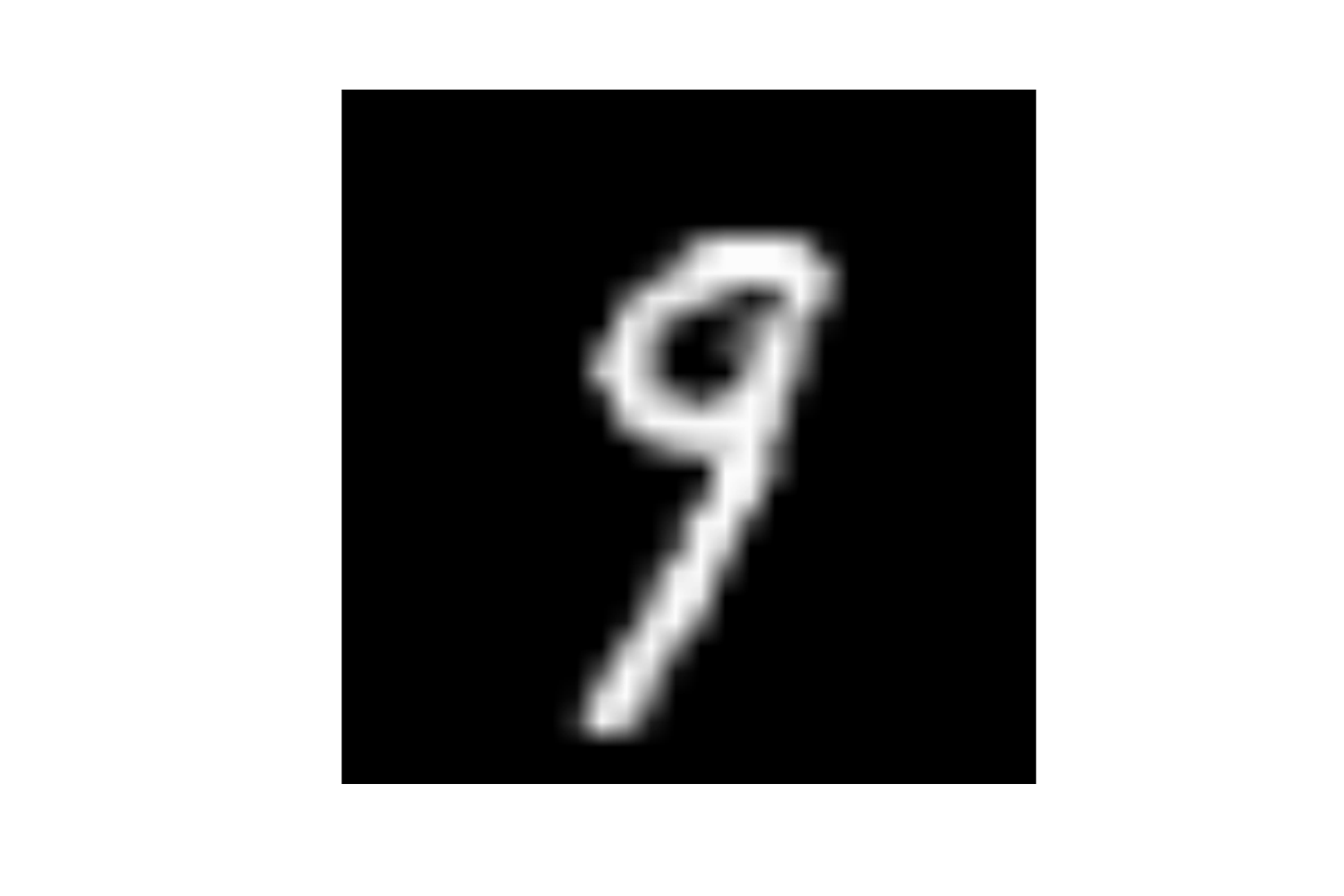}
 \hspace{-5mm}\includegraphics[scale=0.1]{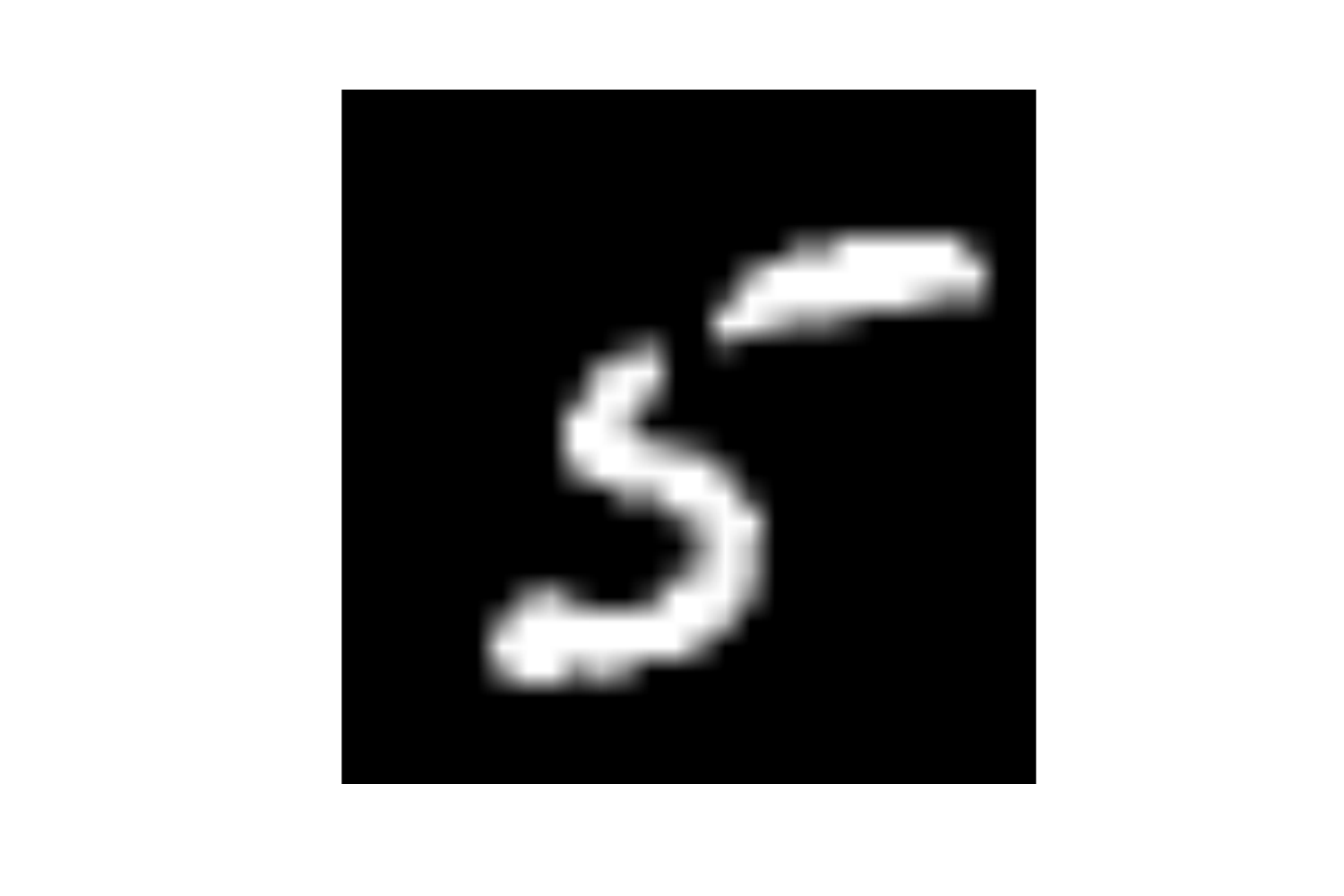}
 \hspace{-5mm}\includegraphics[scale=0.1]{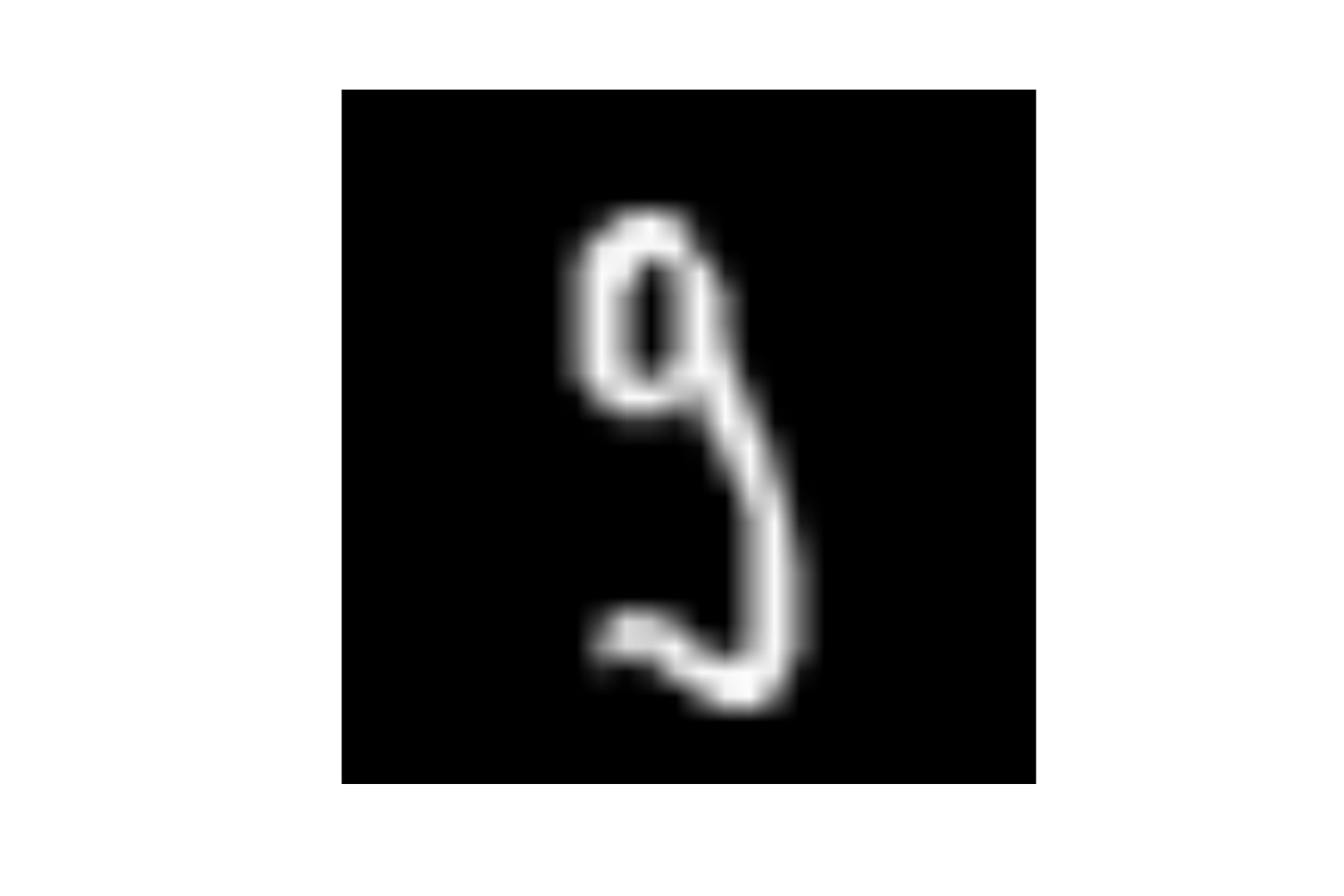}
 \hspace{-5mm}\includegraphics[scale=0.1]{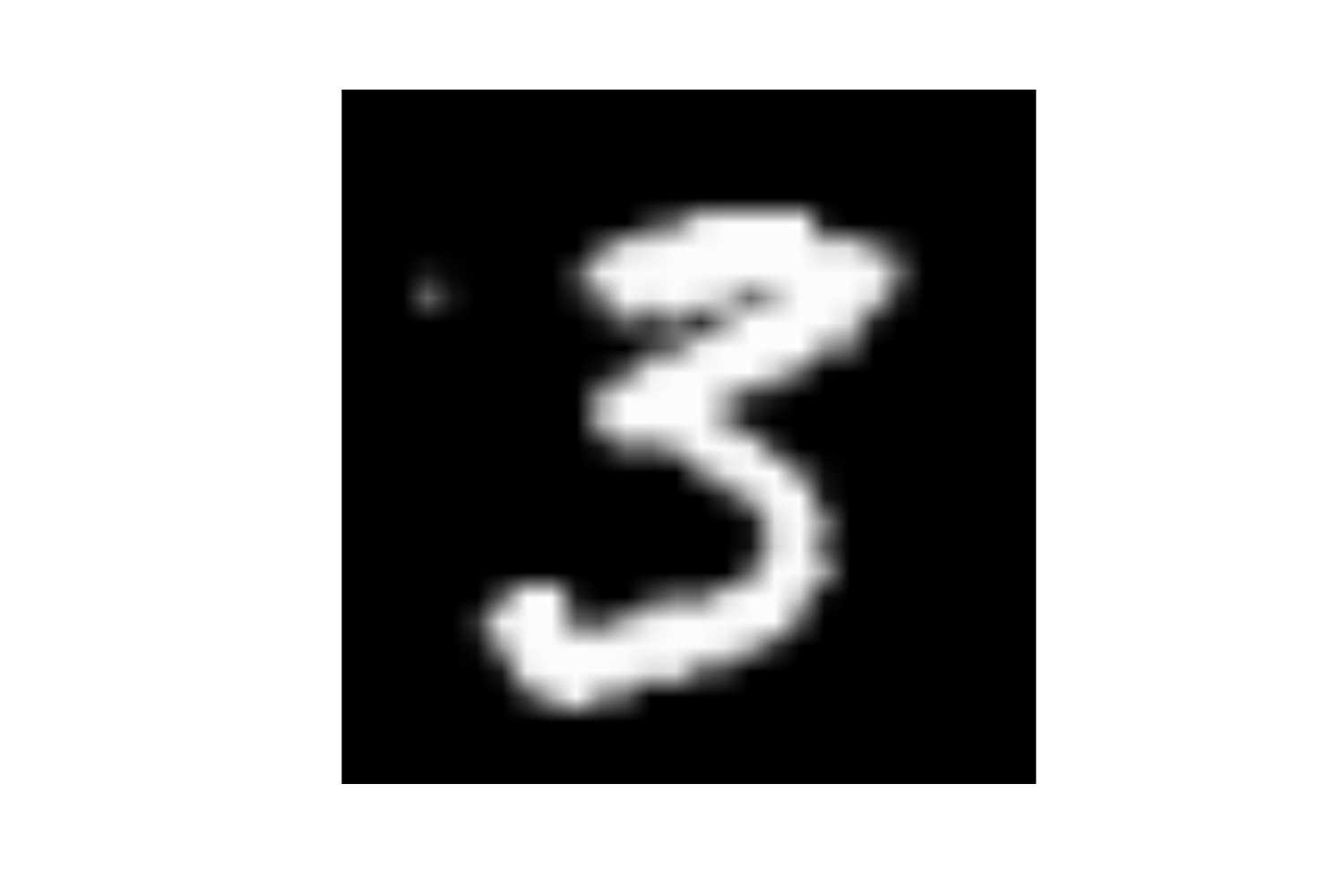}
 }
 \vspace{-1mm}
\\
\vspace{-2mm}
\subfigure[VAE reconstruction (number of hidden variables $50$)]{\label{fig:mnist_reconst_2}\includegraphics[scale=0.1]{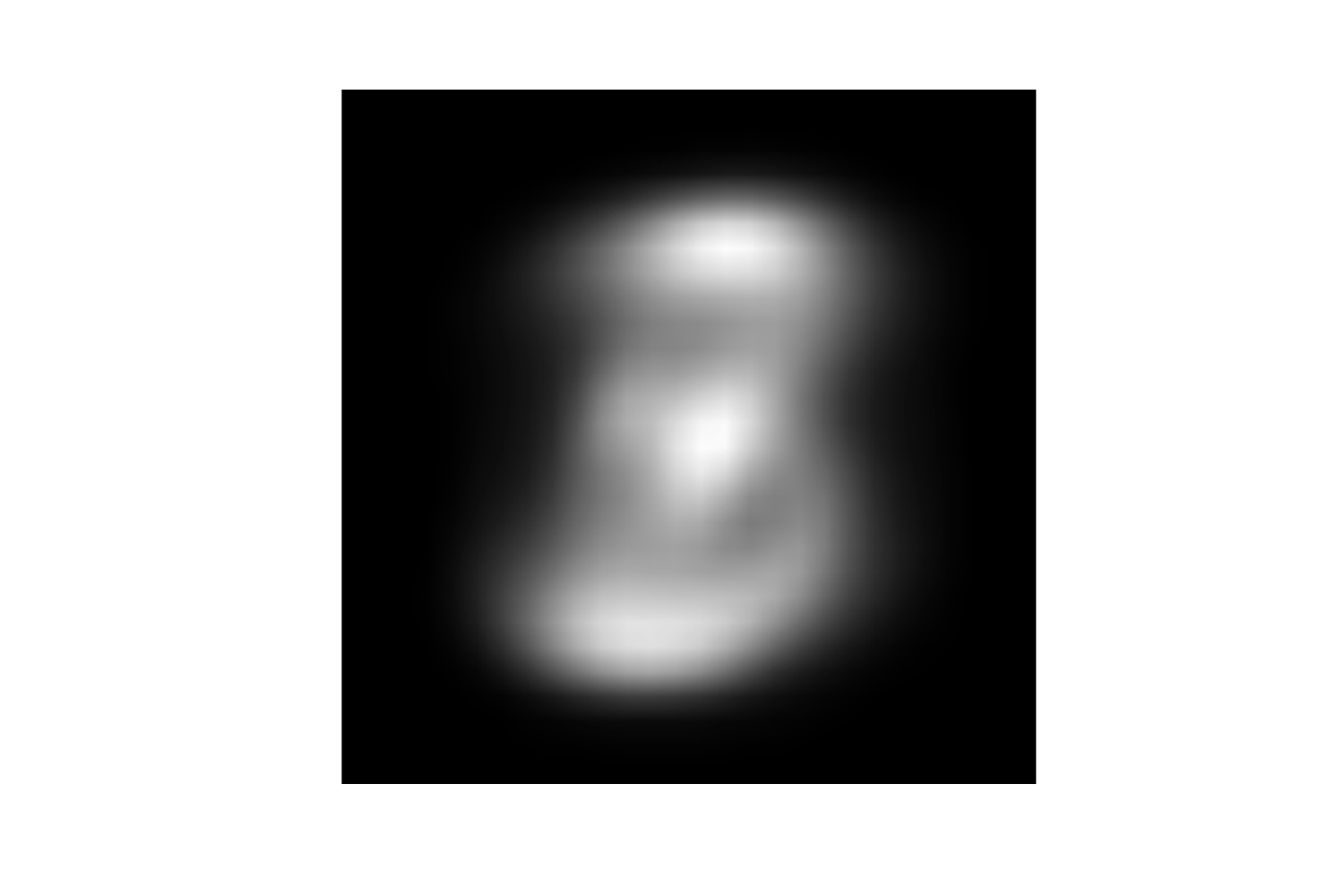}
 \hspace{-5mm}\includegraphics[scale=0.1]{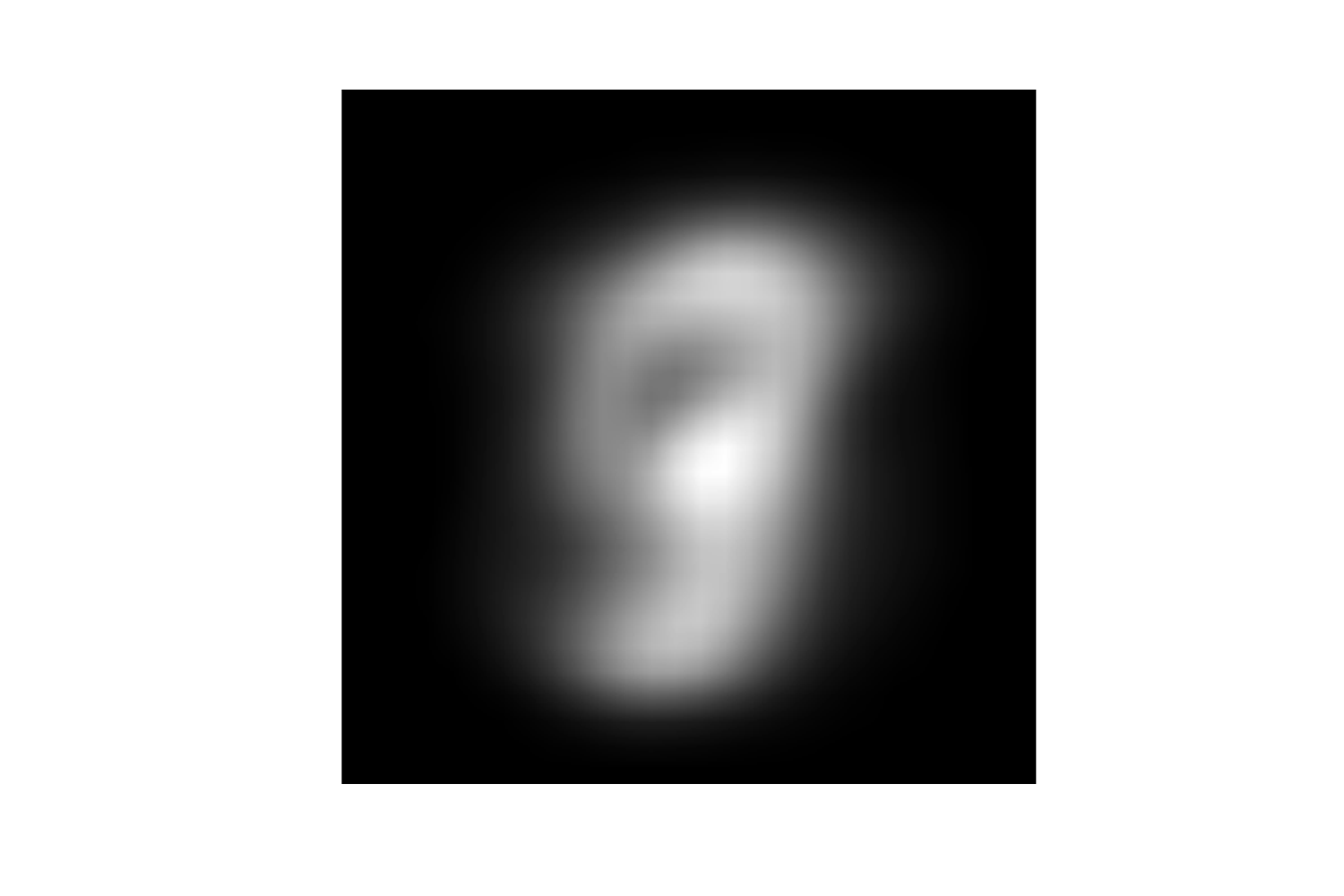}
 \hspace{-5mm}\includegraphics[scale=0.1]{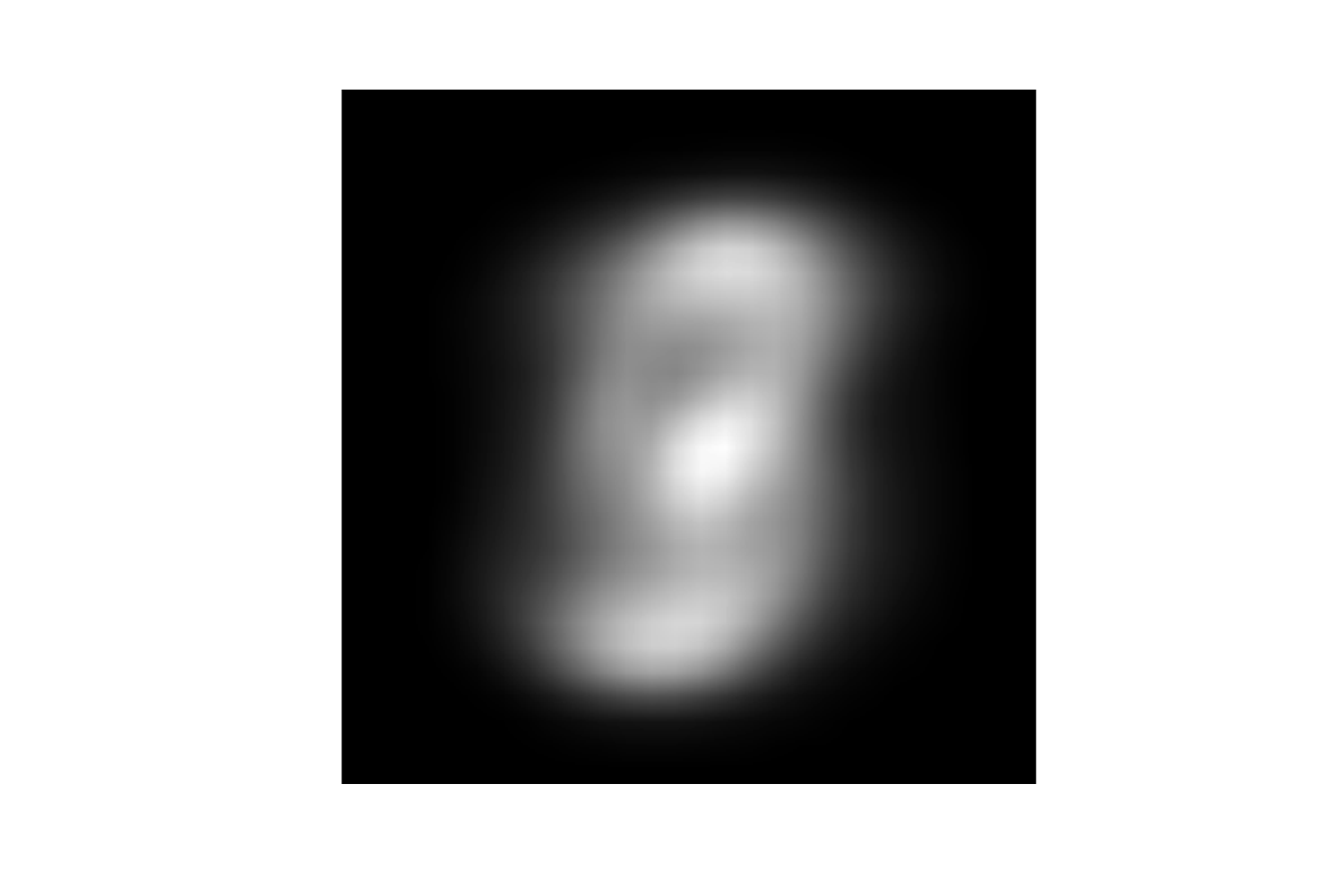}
 \hspace{-5mm}\includegraphics[scale=0.1]{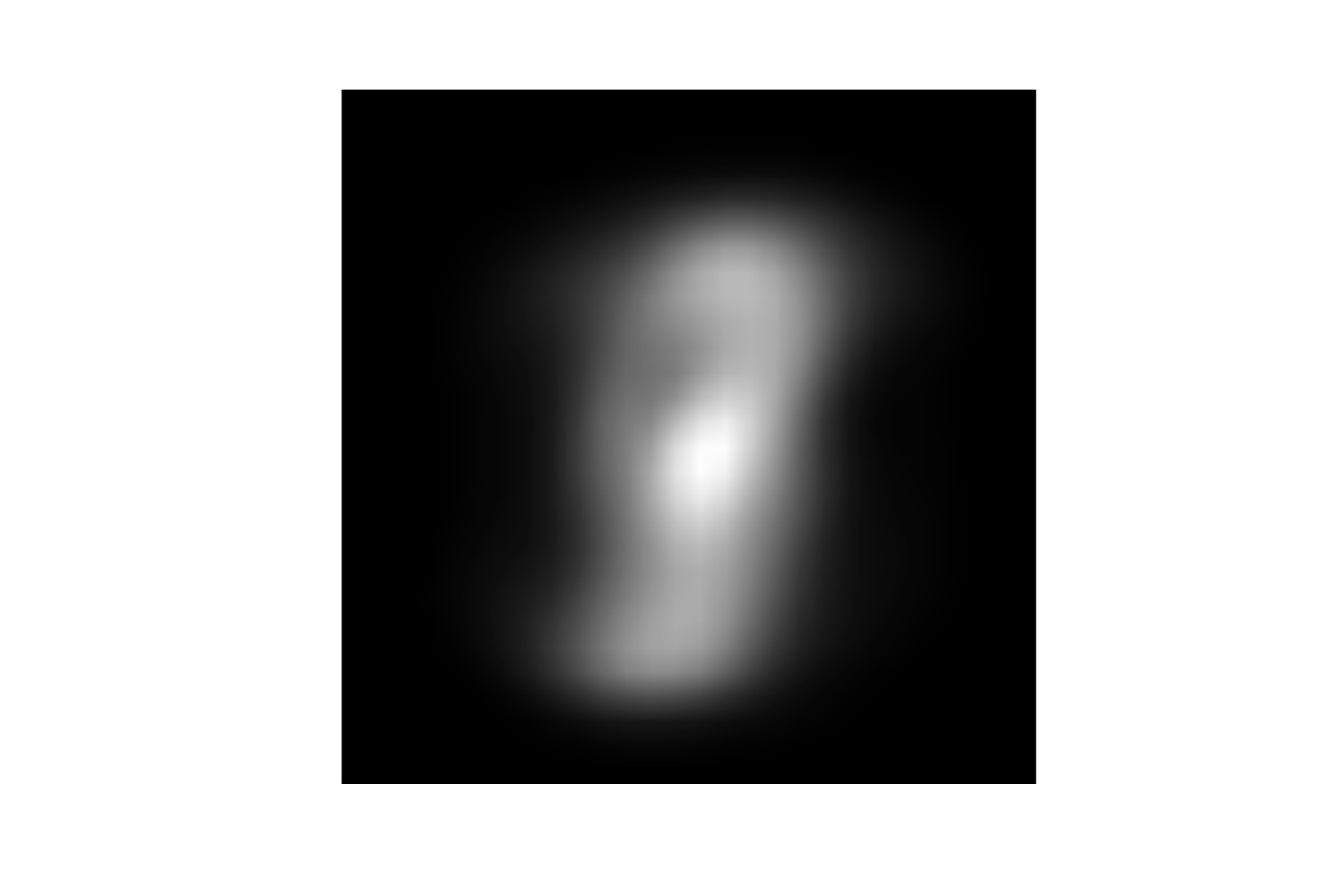}
 \hspace{-5mm}\includegraphics[scale=0.1]{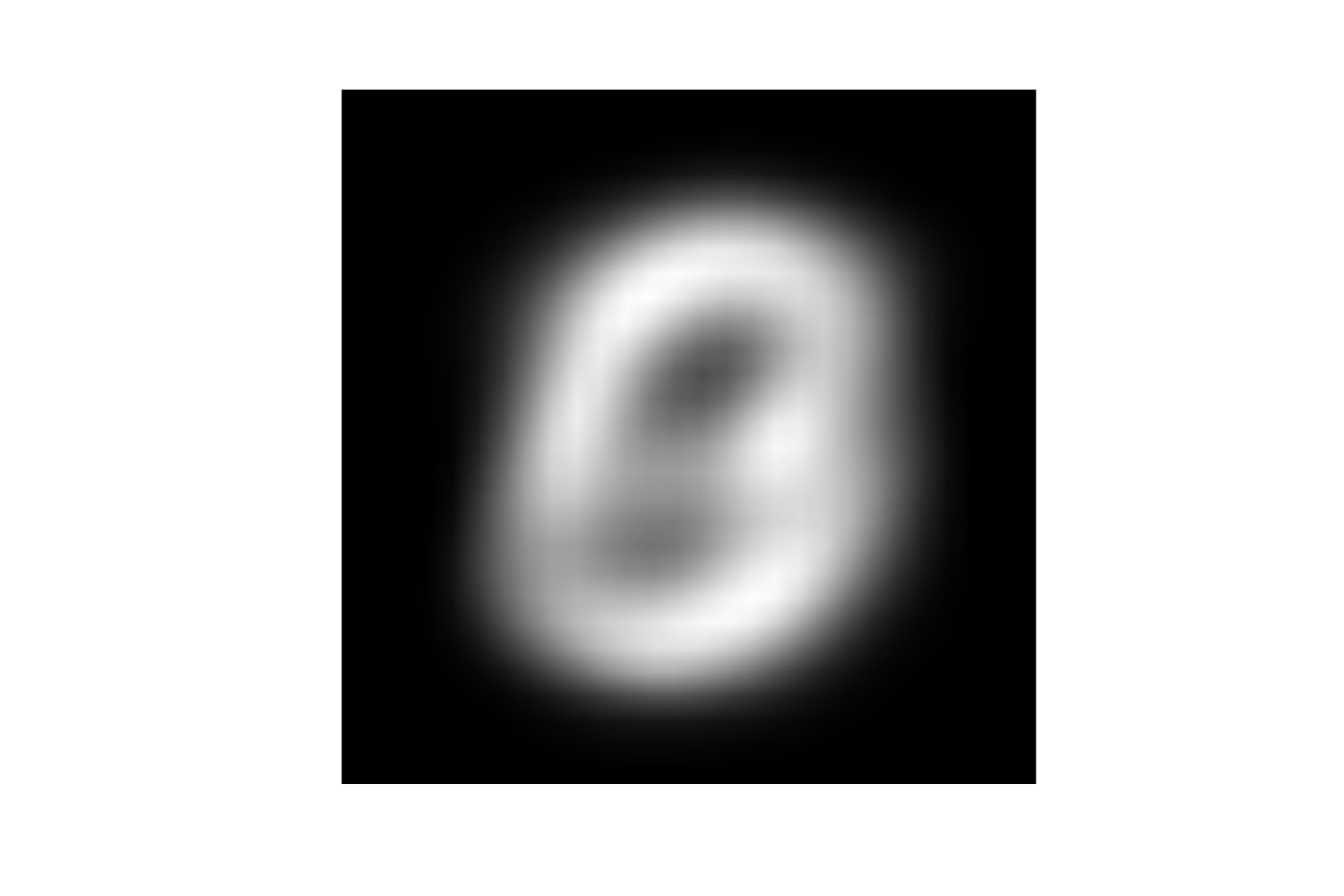}
 \hspace{-5mm}\includegraphics[scale=0.1]{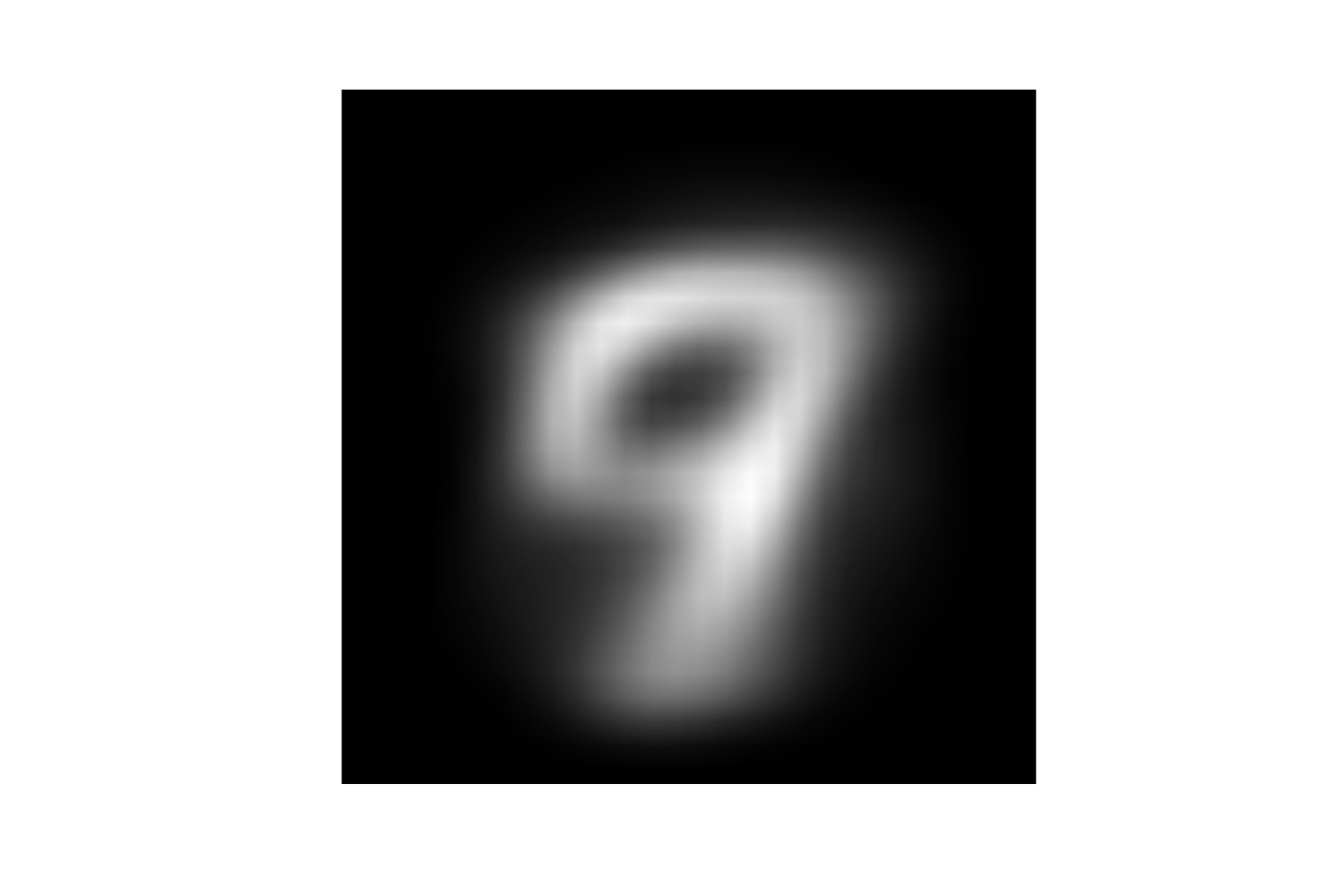}
 \hspace{-5mm}\includegraphics[scale=0.1]{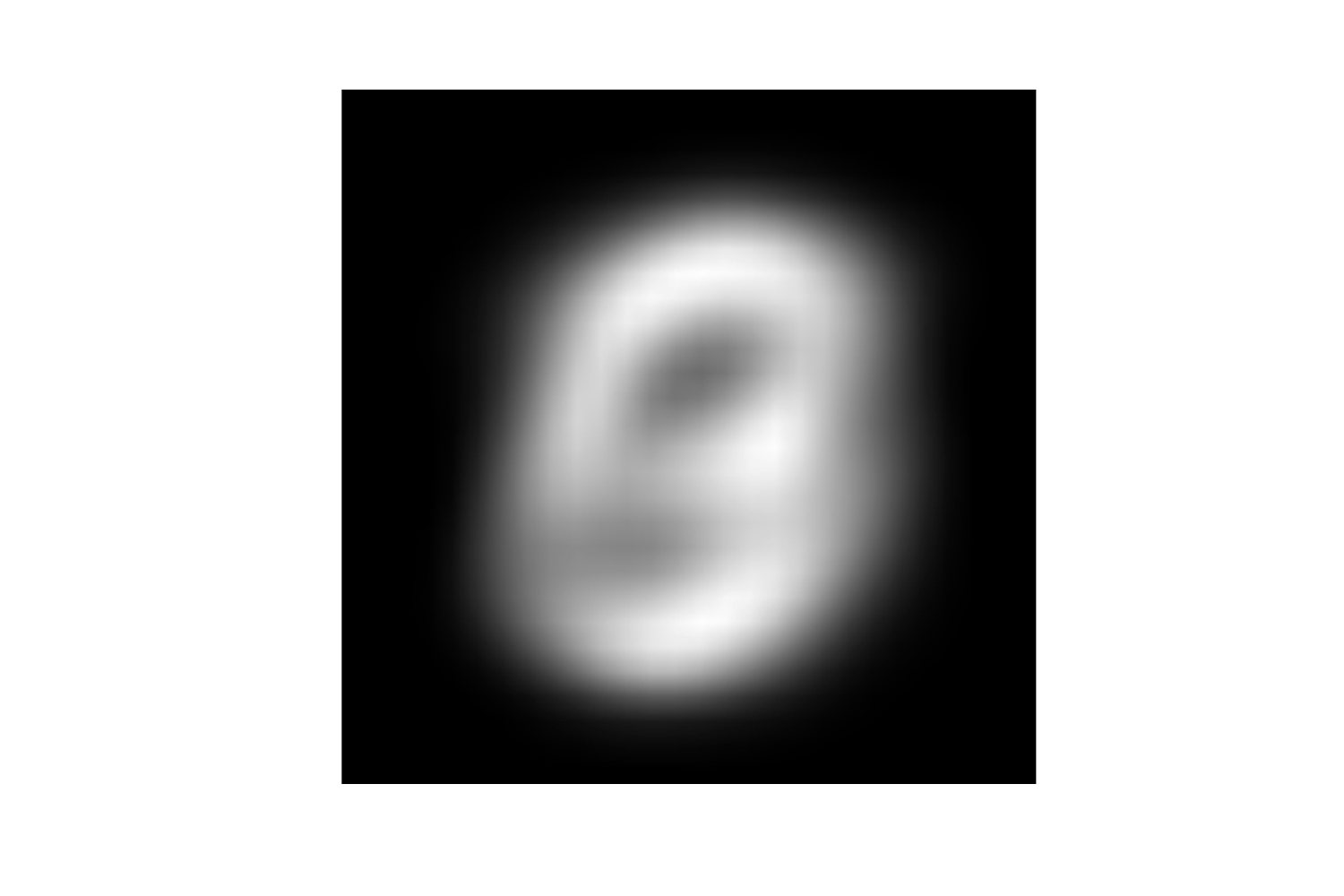}
 \hspace{-5mm}\includegraphics[scale=0.1]{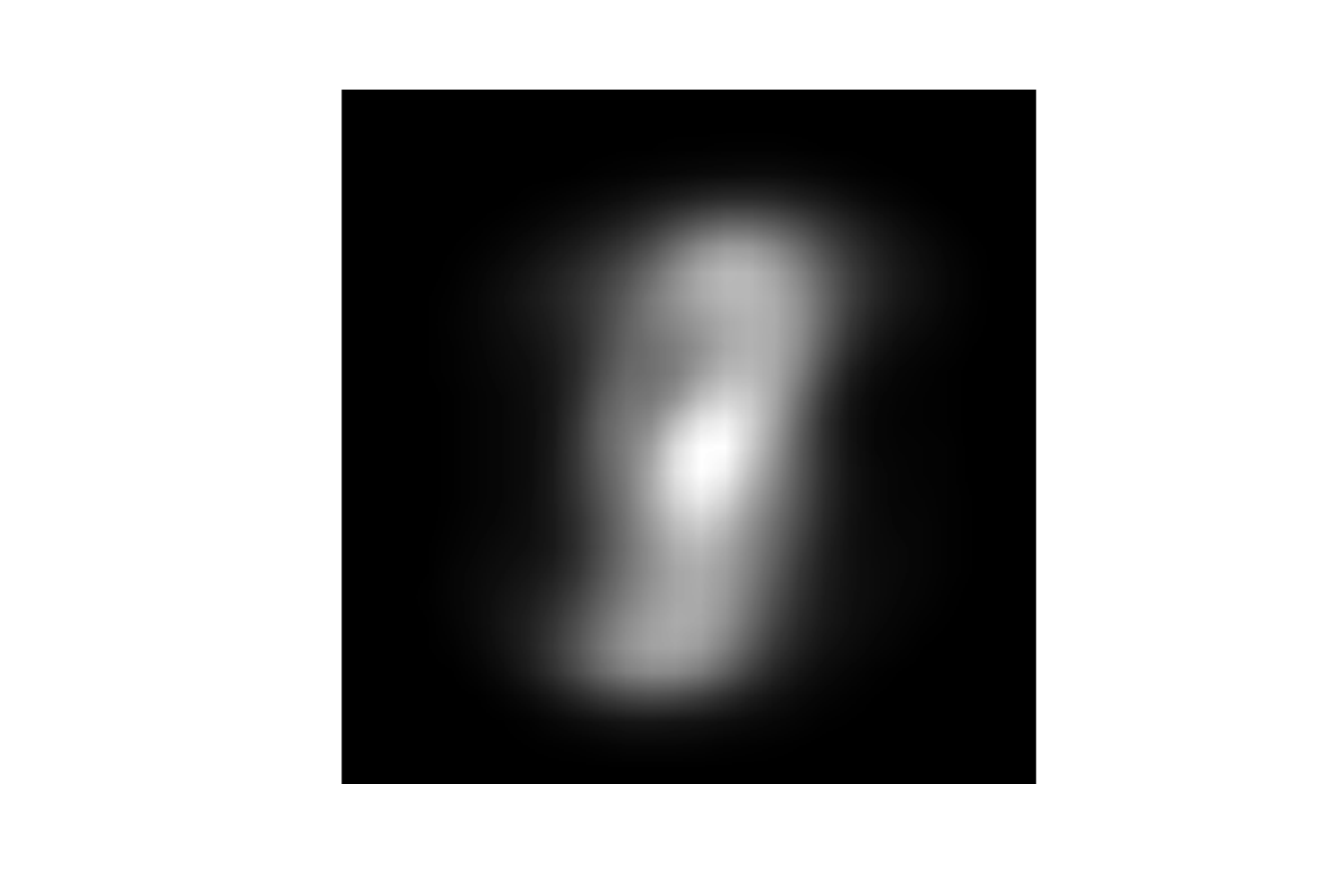}
 \hspace{-5mm}\includegraphics[scale=0.1]{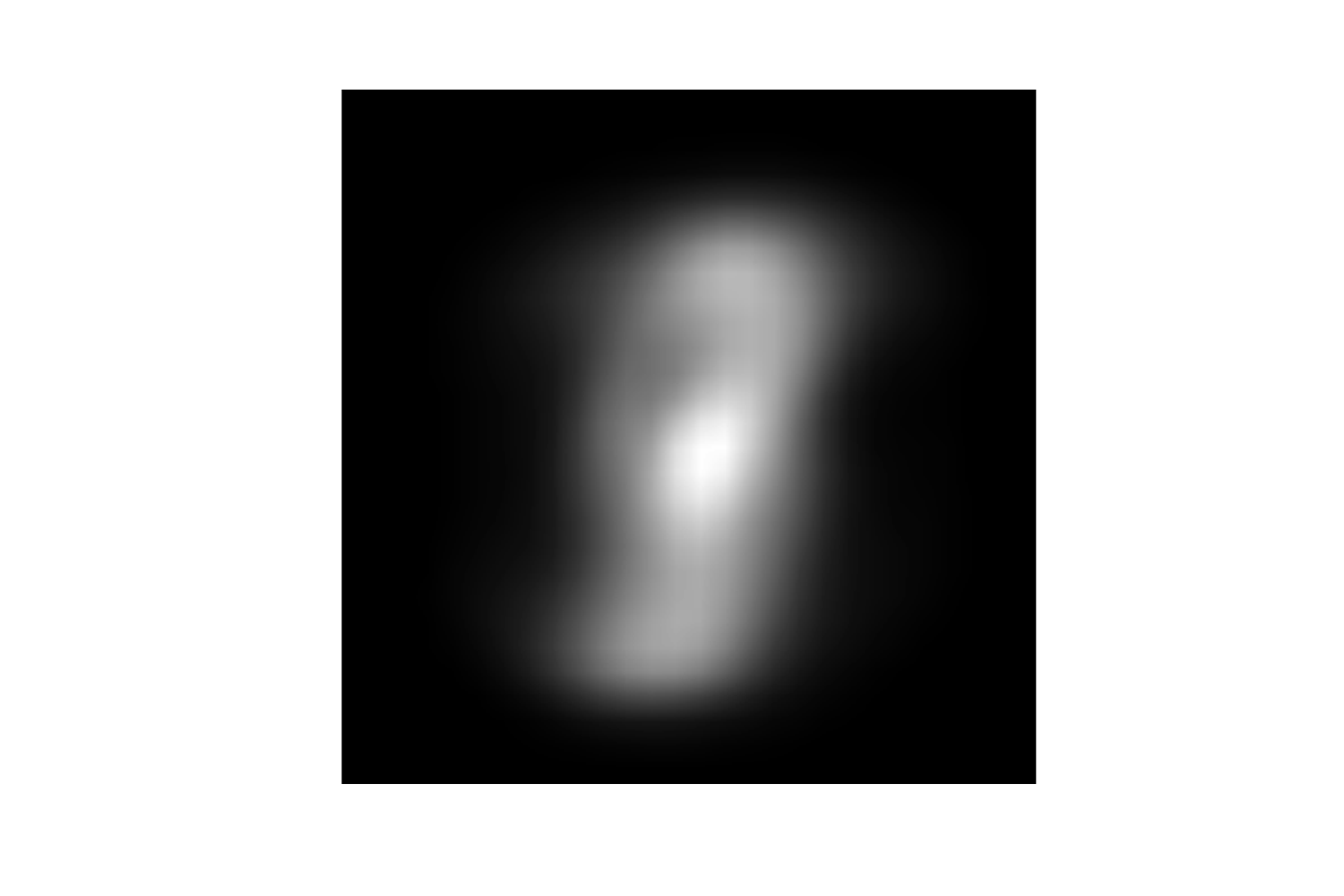}
 \hspace{-5mm}\includegraphics[scale=0.1]{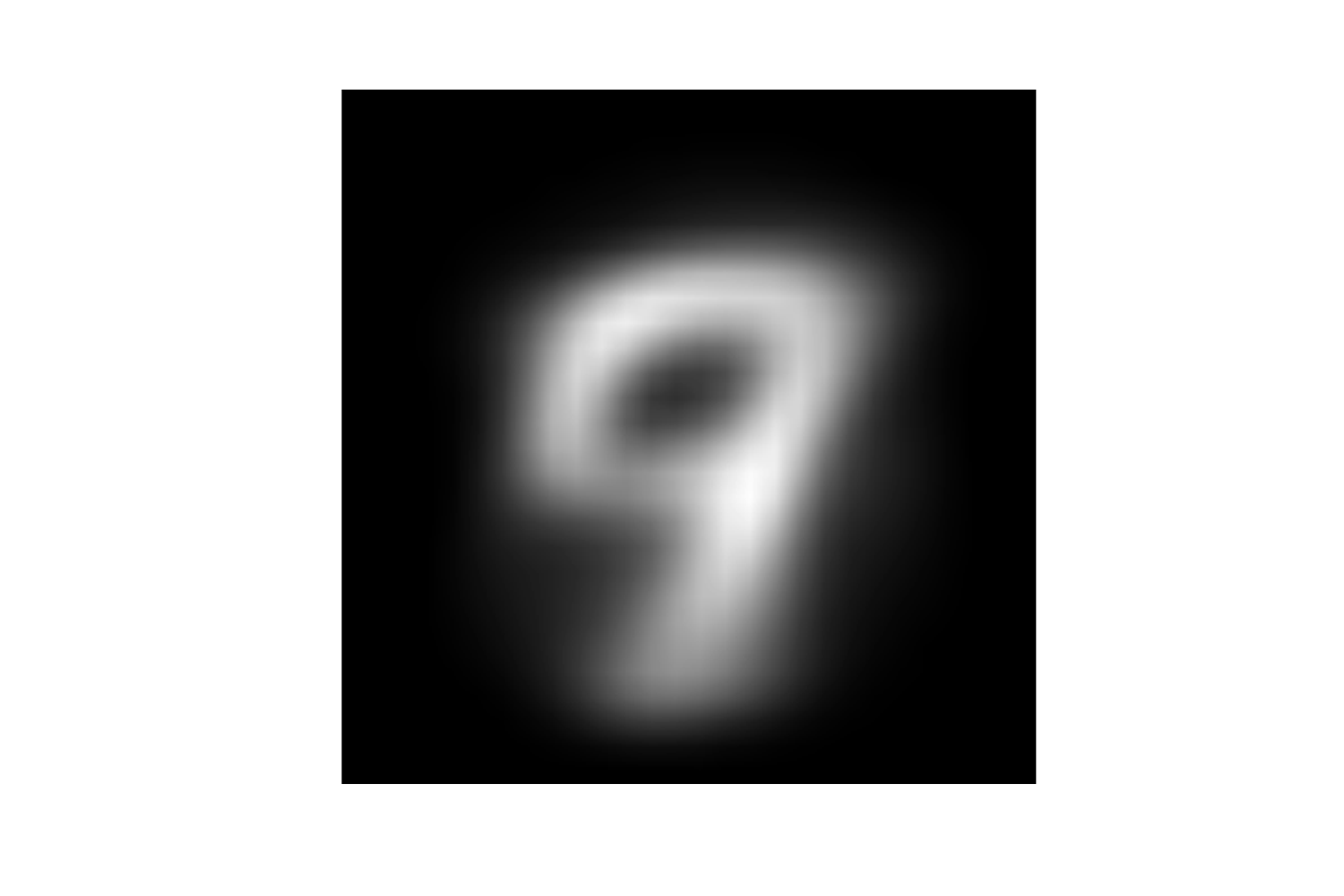}
 \hspace{-5mm}\includegraphics[scale=0.1]{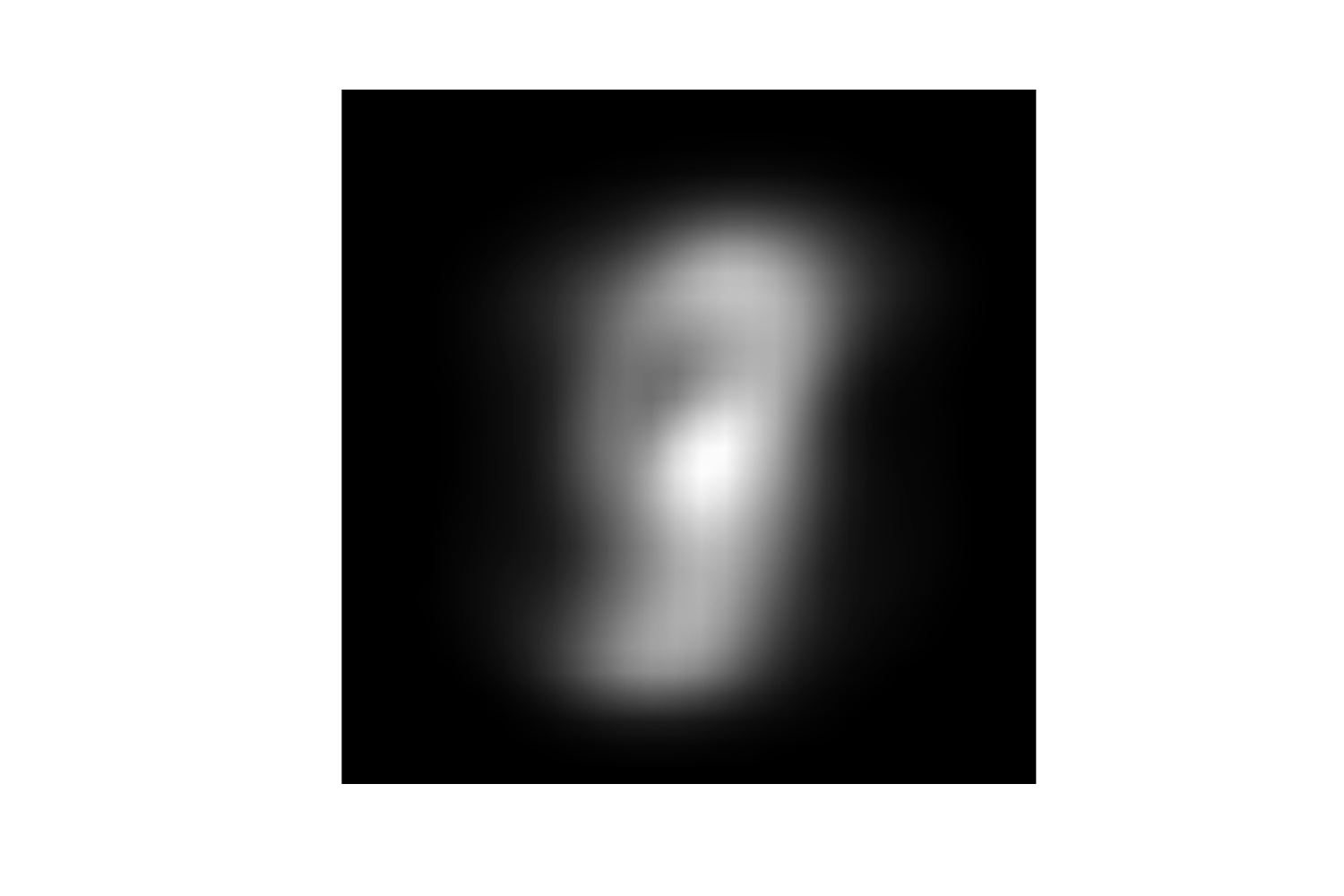}
 \hspace{-5mm}\includegraphics[scale=0.1]{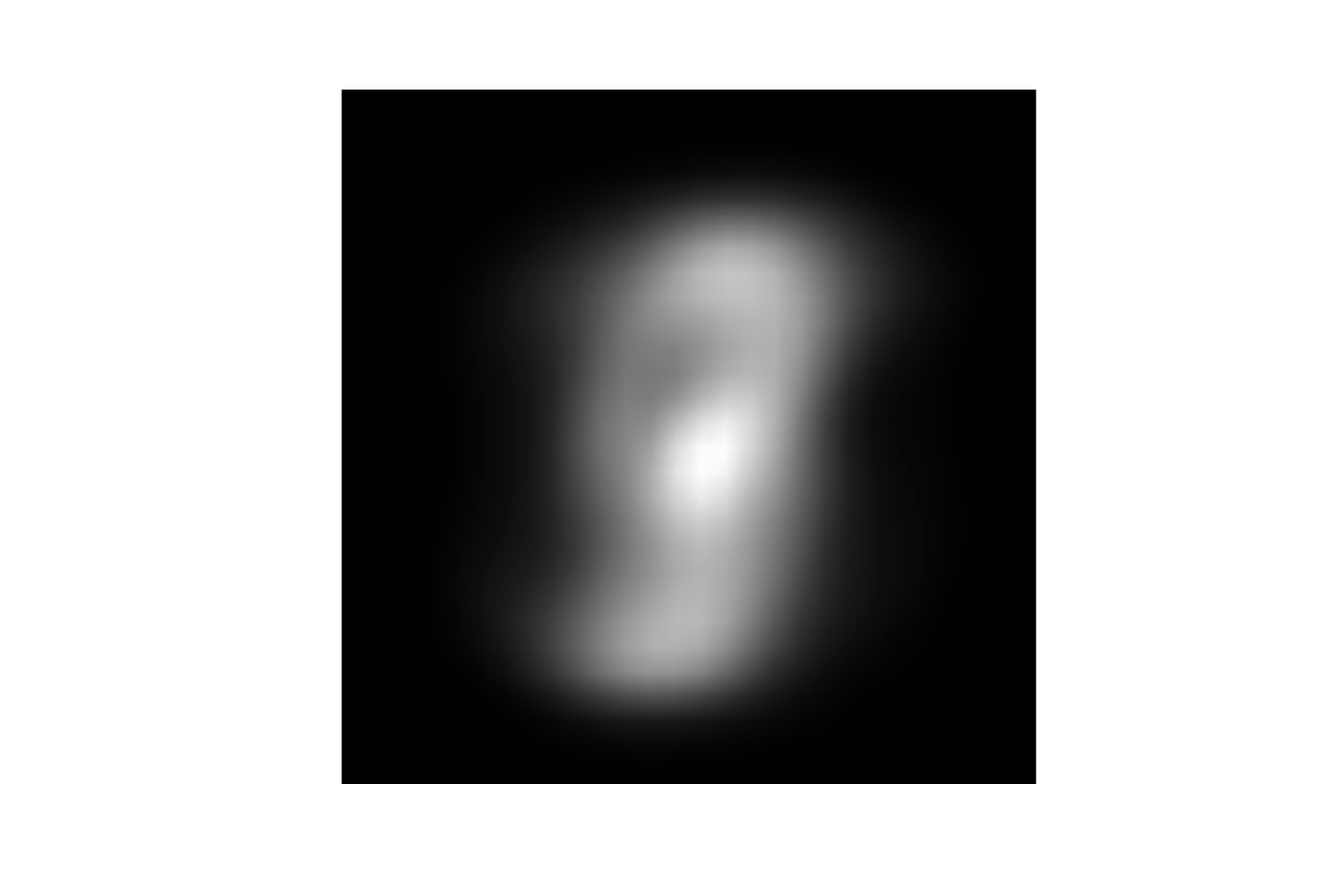}
 \hspace{-5mm}\includegraphics[scale=0.1]{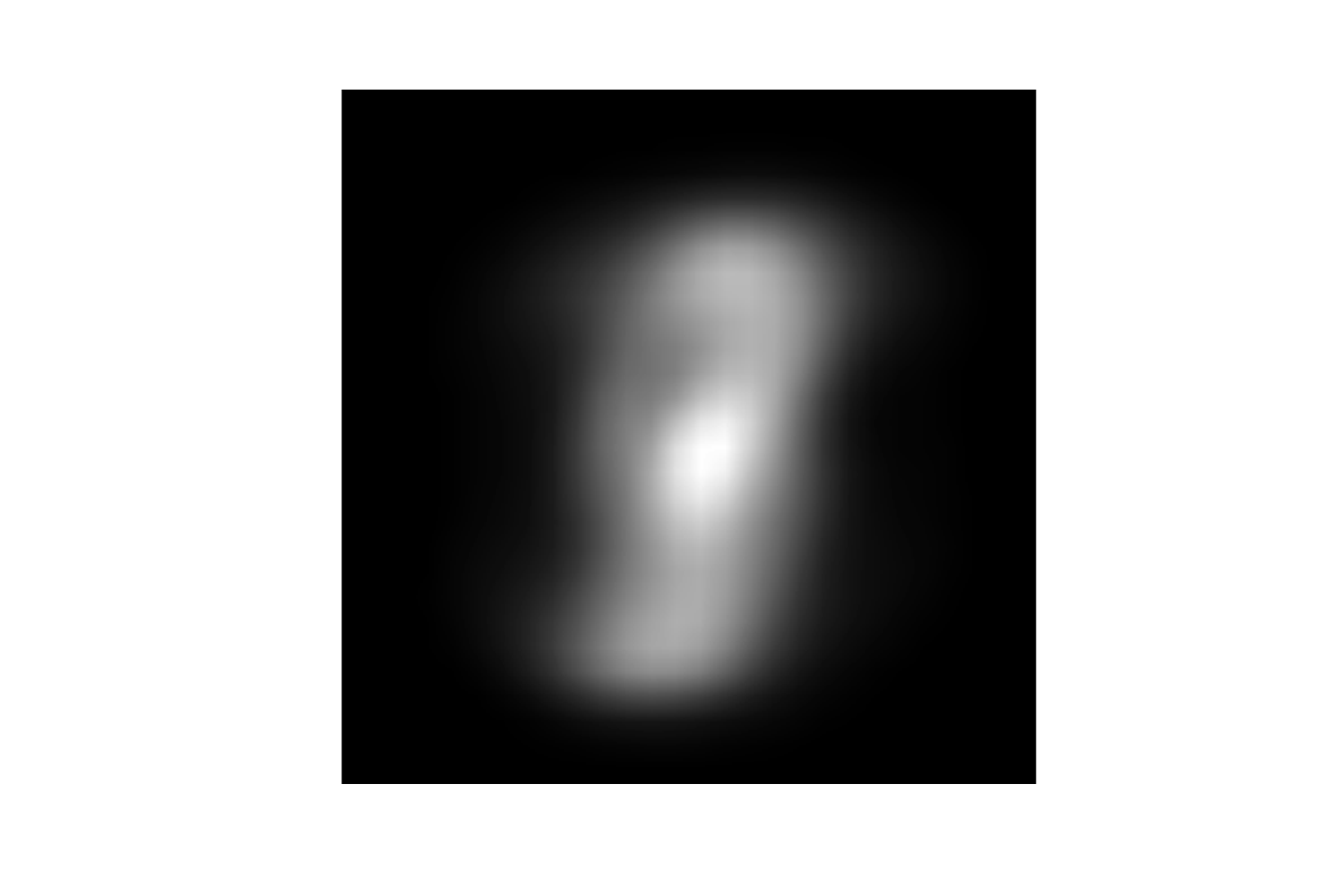}
 \hspace{-5mm}\includegraphics[scale=0.1]{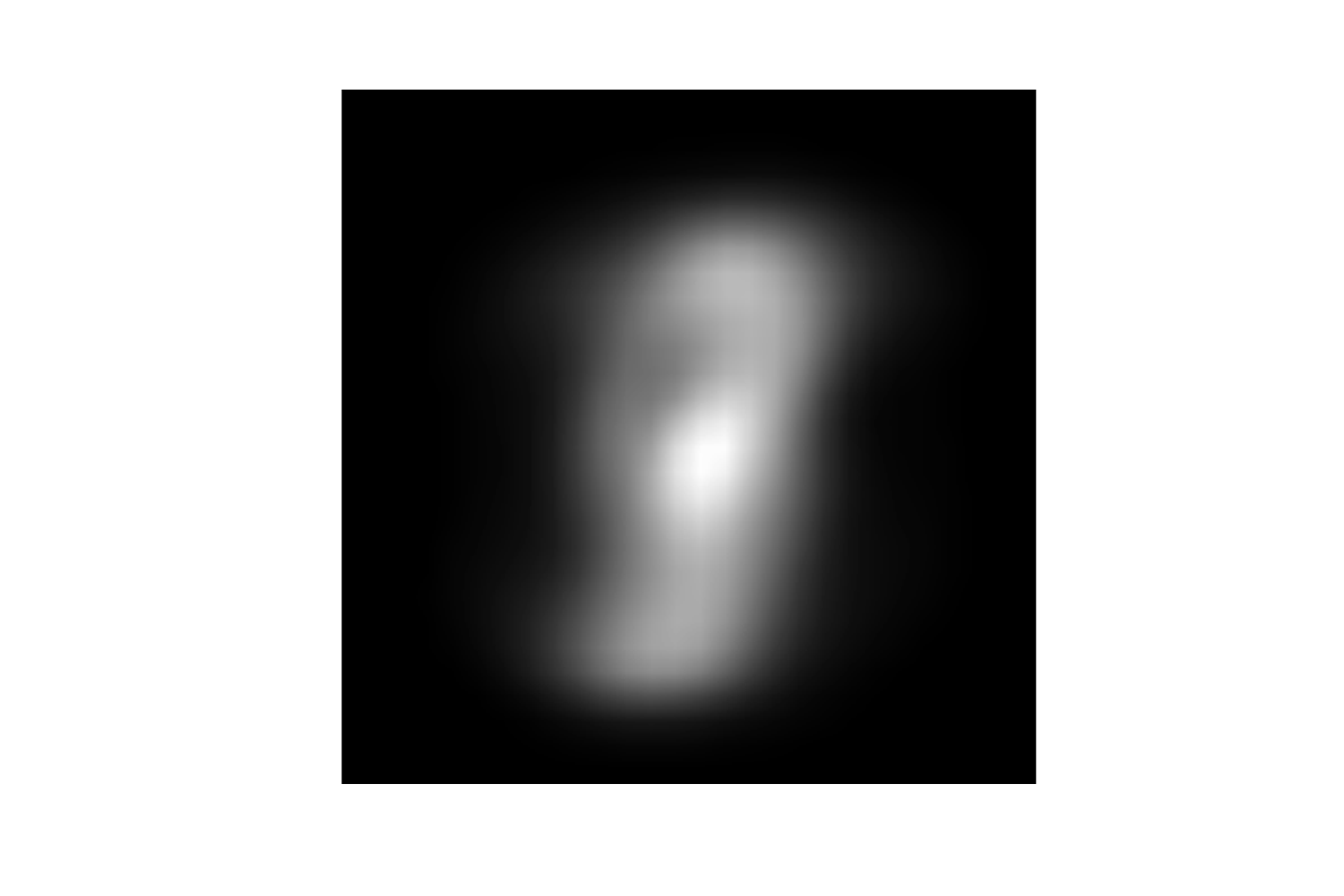}
 \hspace{-5mm}\includegraphics[scale=0.1]{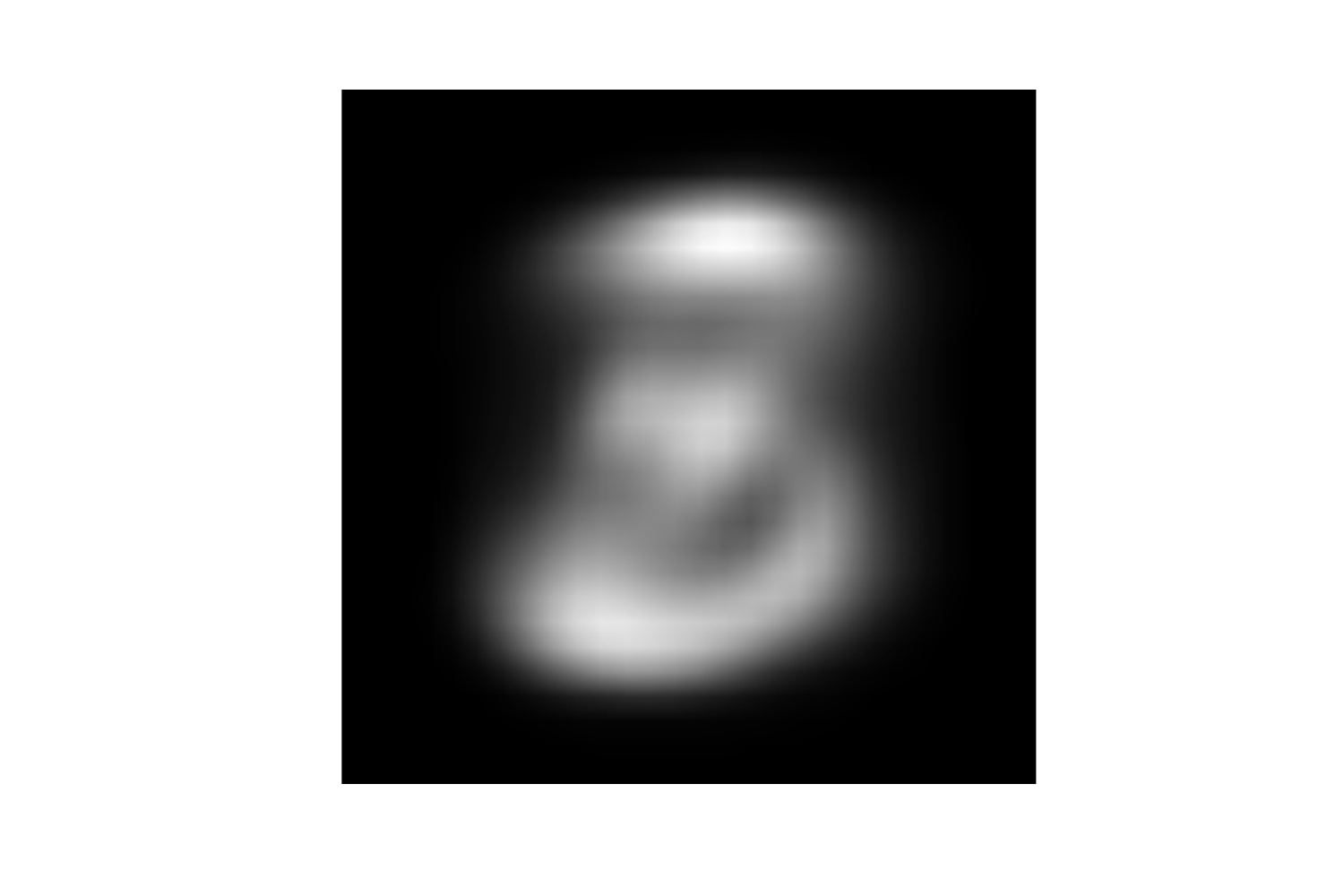}
 }
 \vspace{-1mm}
\\
\vspace{-2mm}
\subfigure[VAE reconstruction (number of hidden variables $1024$)]{\label{fig:mnist_reconst_3}\includegraphics[scale=0.1]{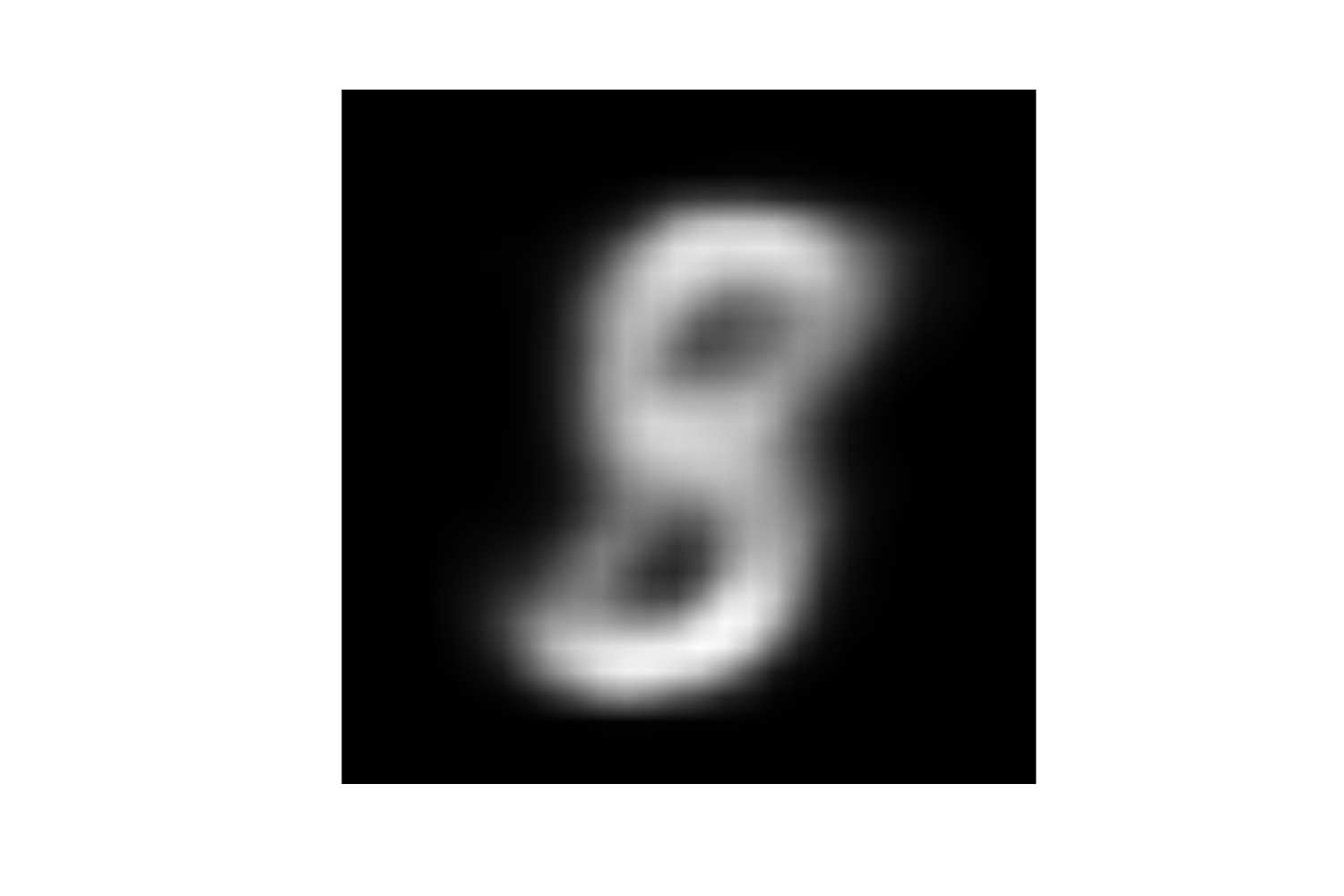}
 \hspace{-5mm}\includegraphics[scale=0.1]{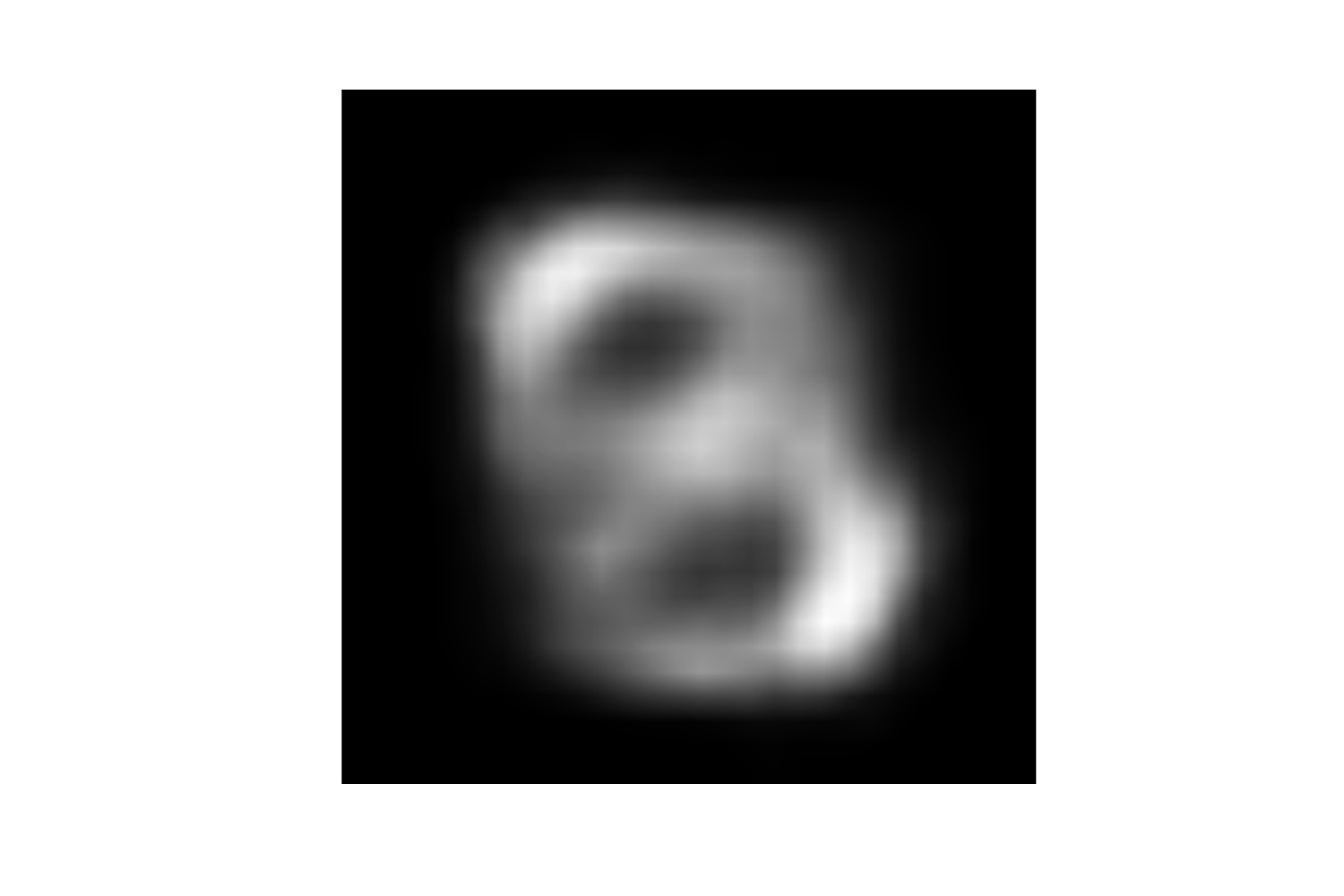}
 \hspace{-5mm}\includegraphics[scale=0.1]{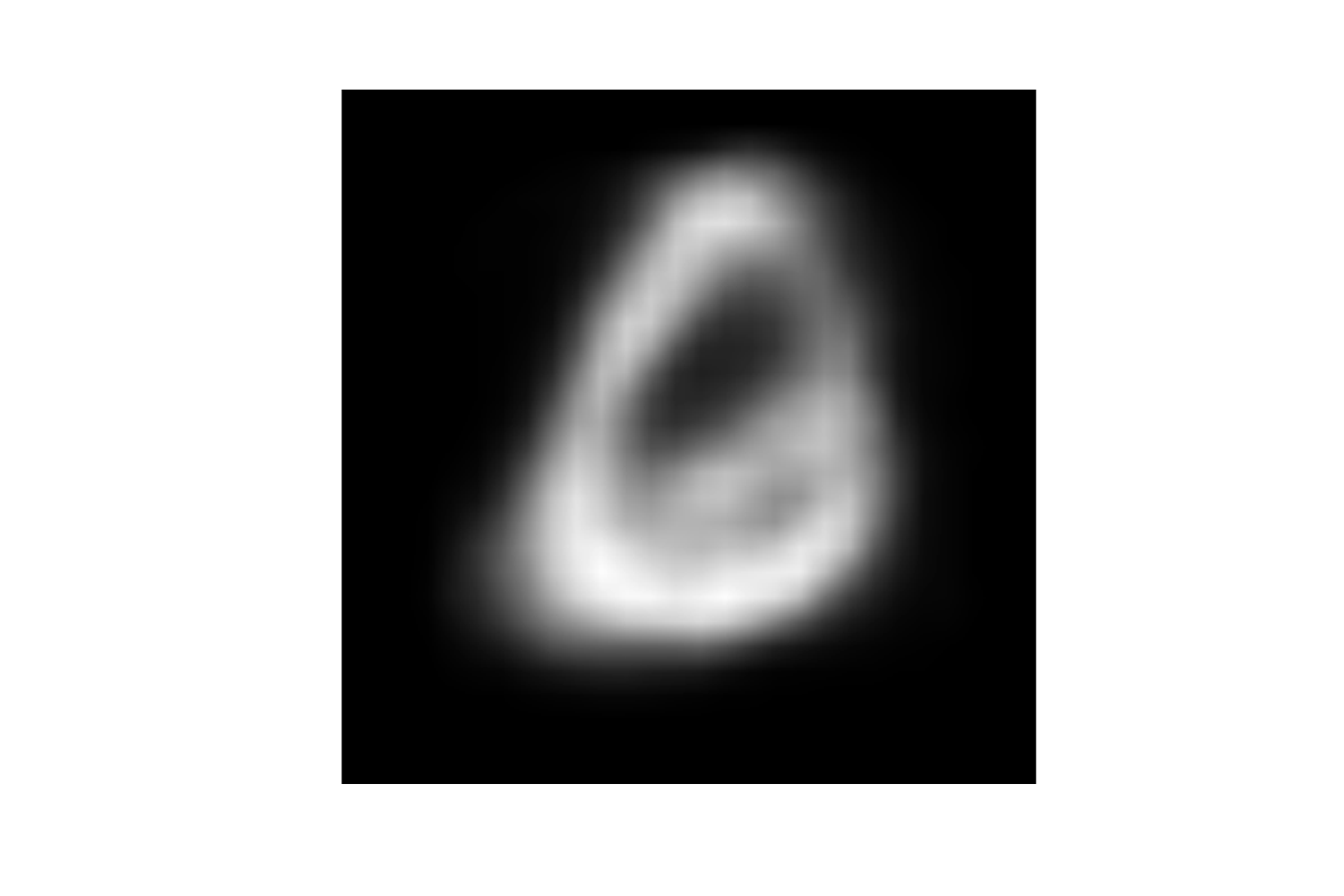}
  \hspace{-5mm}\includegraphics[scale=0.1]{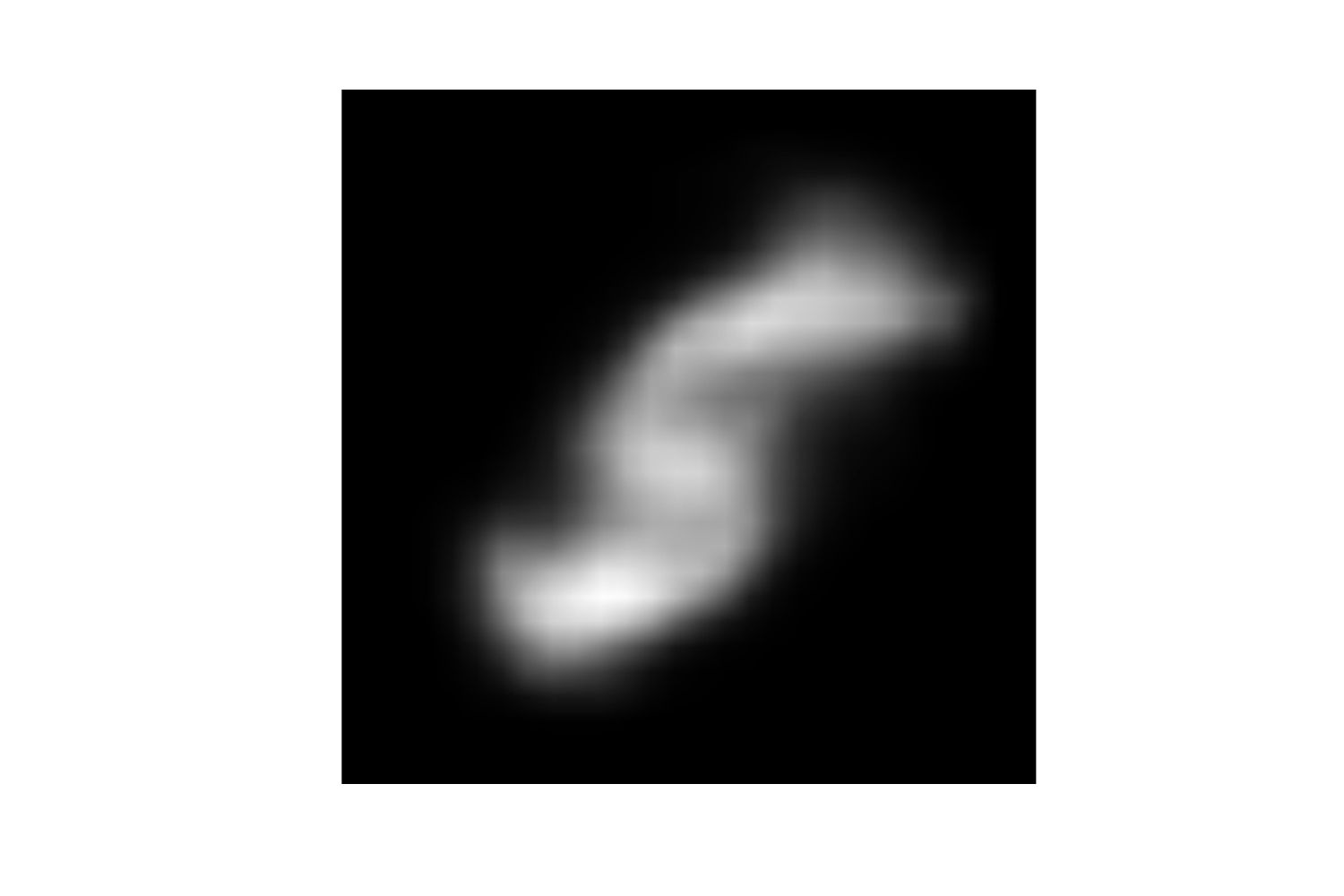}
 \hspace{-5mm}\includegraphics[scale=0.1]{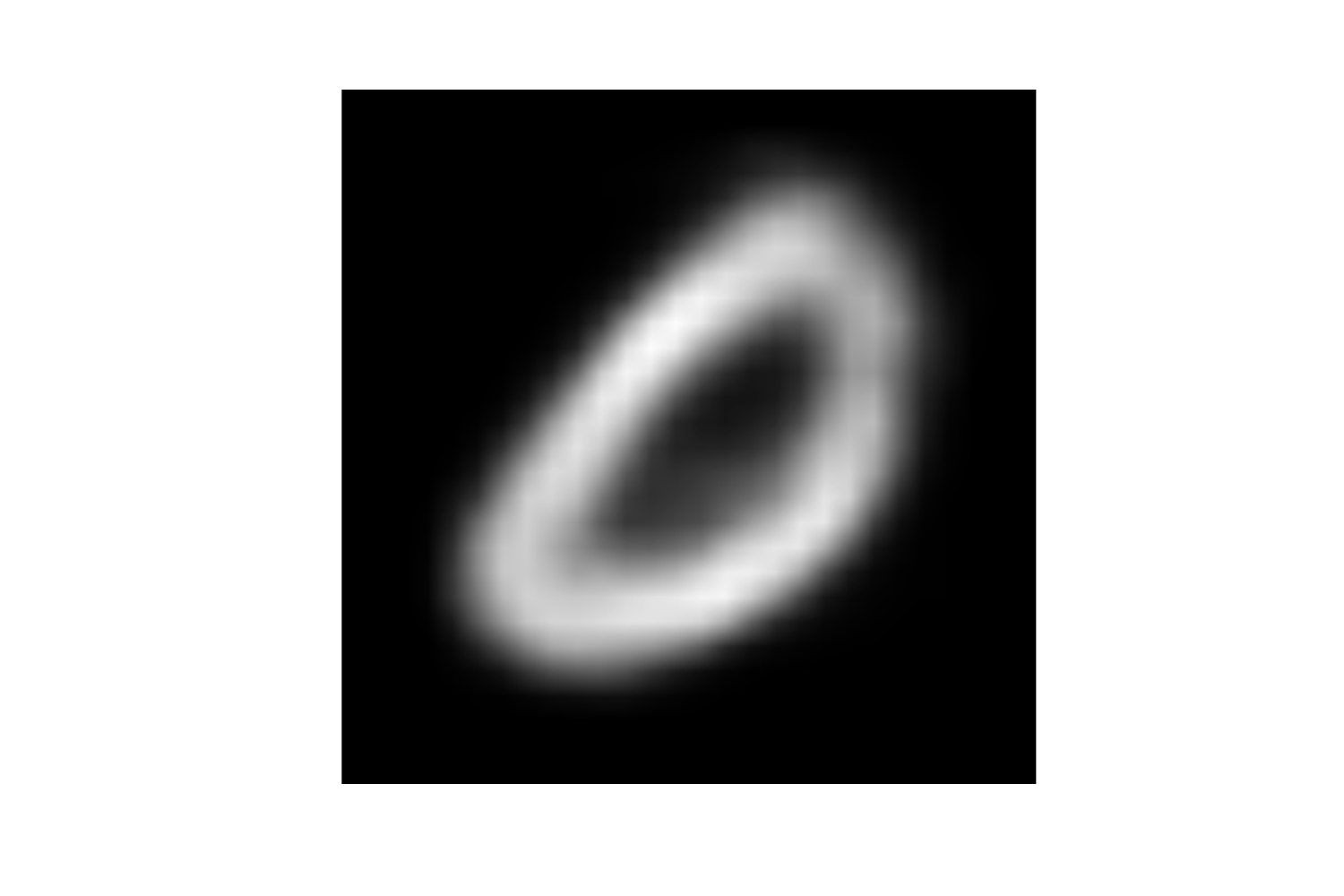}
 \hspace{-5mm}\includegraphics[scale=0.1]{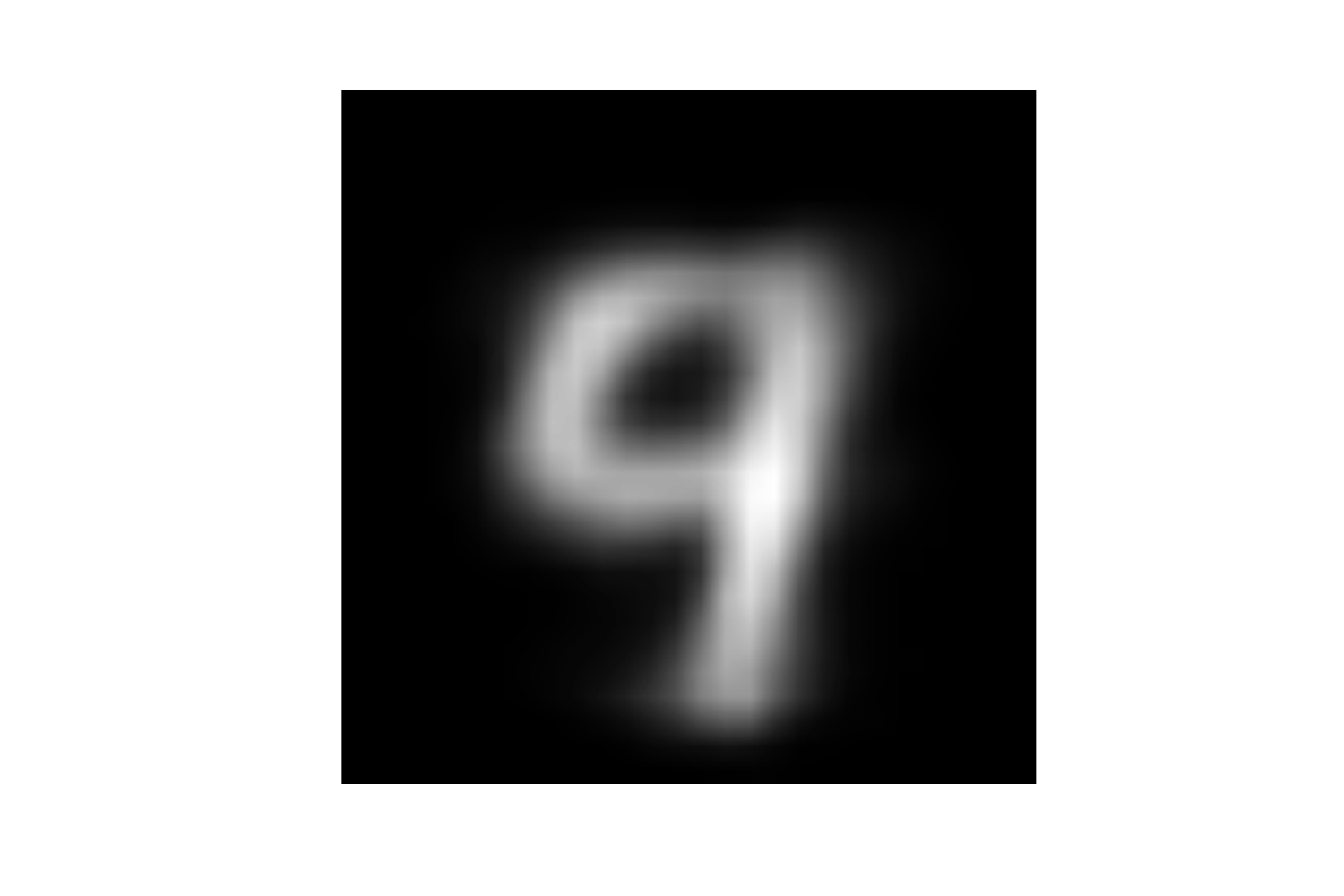}
 \hspace{-5mm}\includegraphics[scale=0.1]{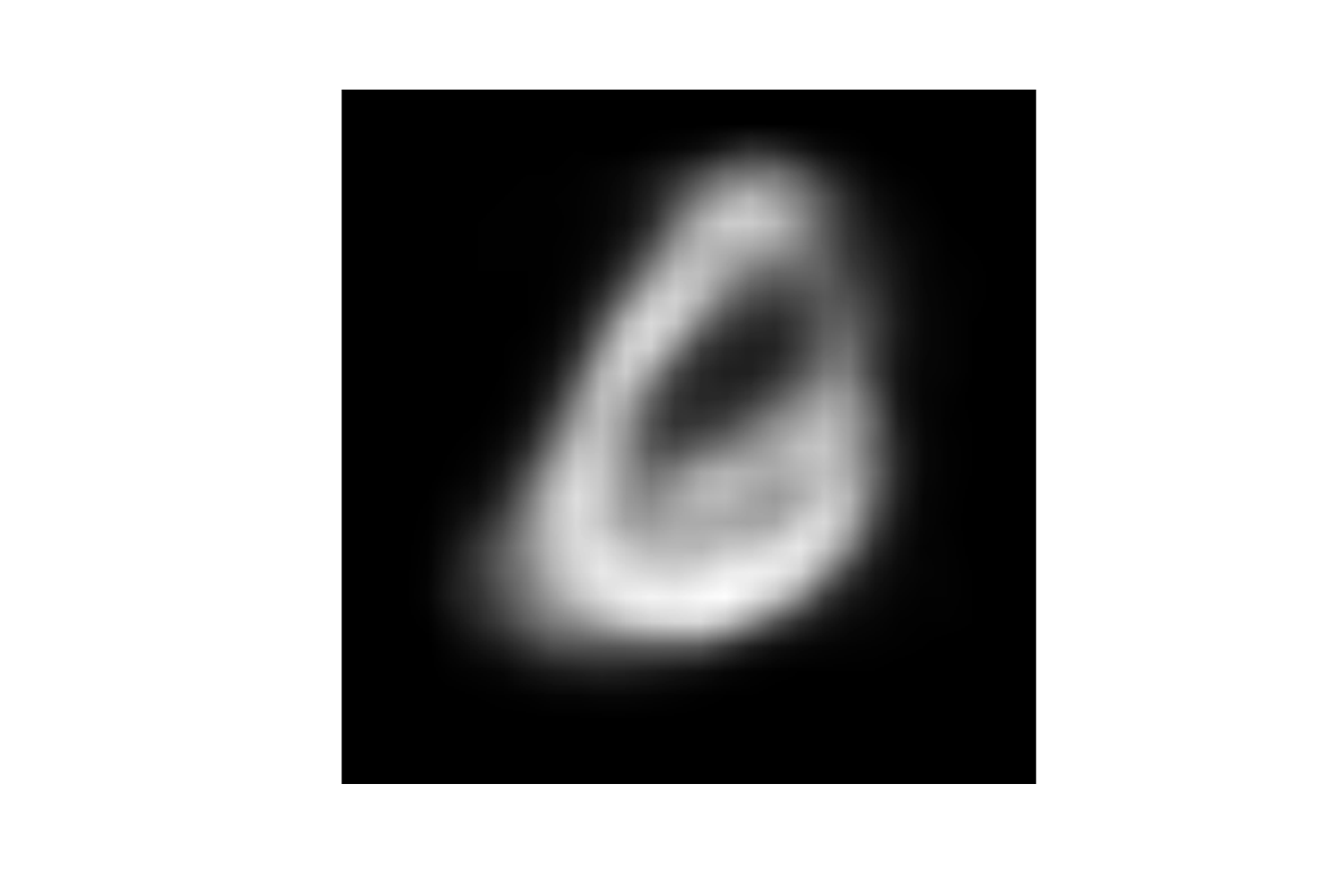}
 \hspace{-5mm}\includegraphics[scale=0.1]{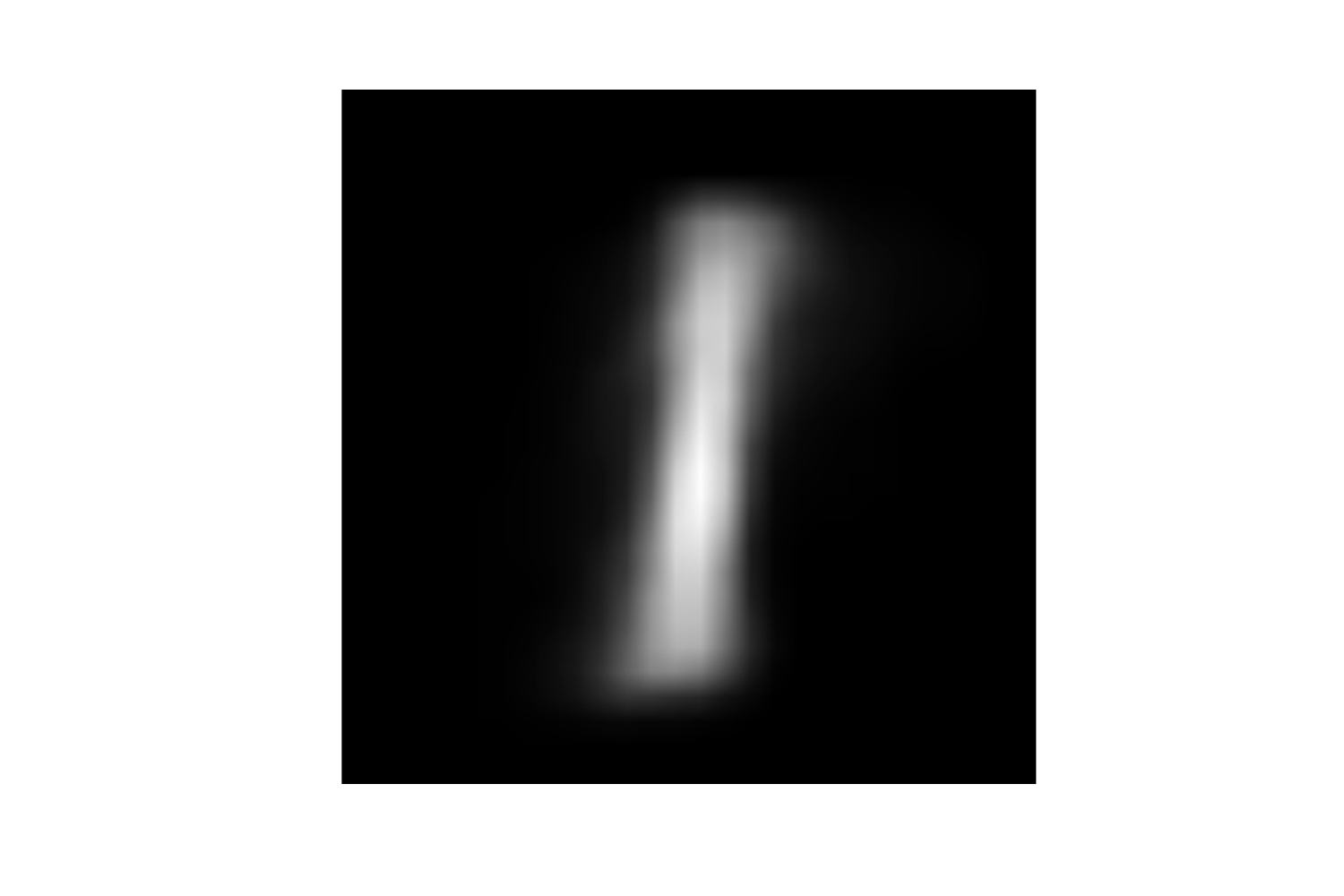}
 \hspace{-5mm}\includegraphics[scale=0.1]{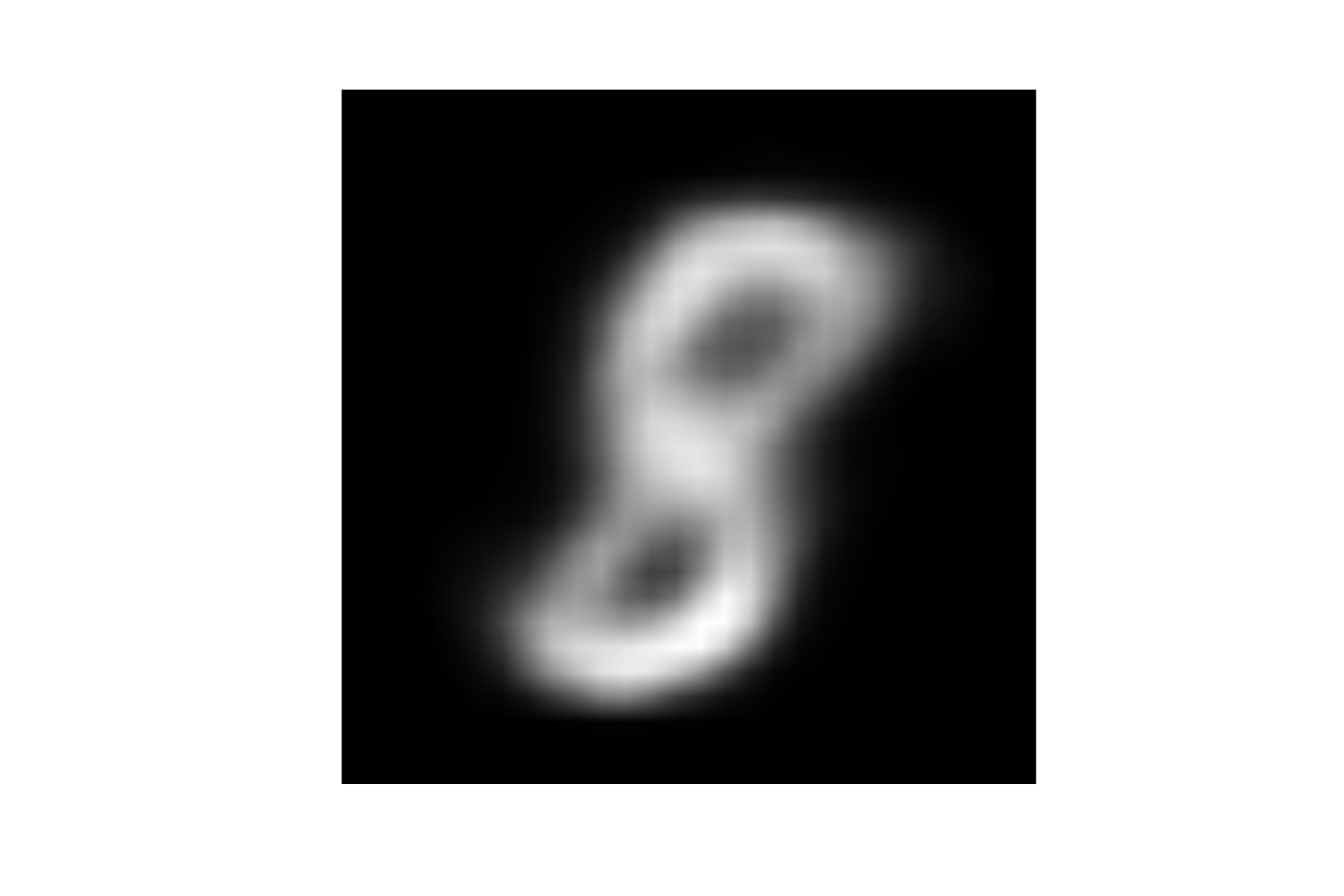}
 \hspace{-5mm}\includegraphics[scale=0.1]{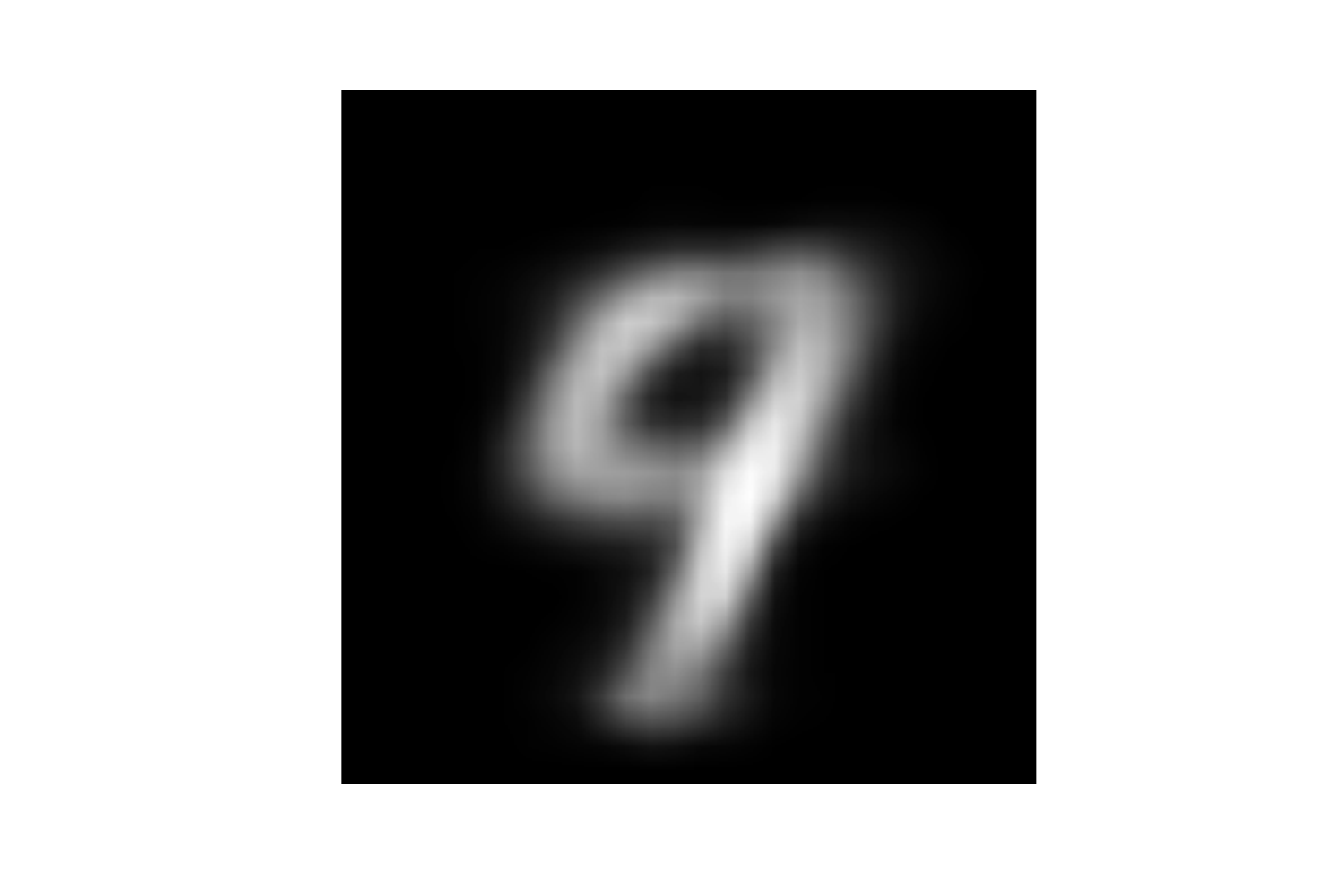}
 \hspace{-5mm}\includegraphics[scale=0.1]{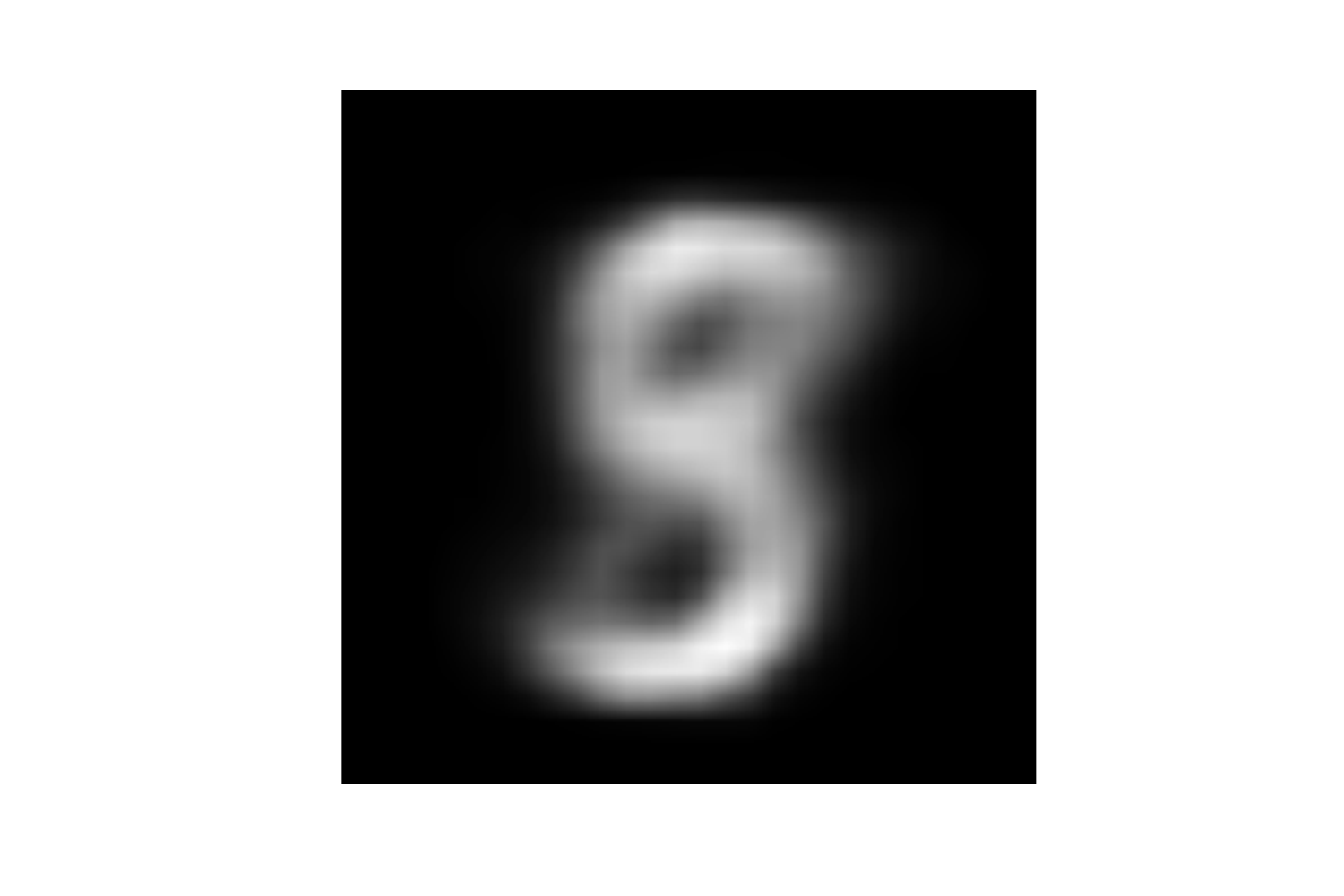}
 \hspace{-5mm}\includegraphics[scale=0.1]{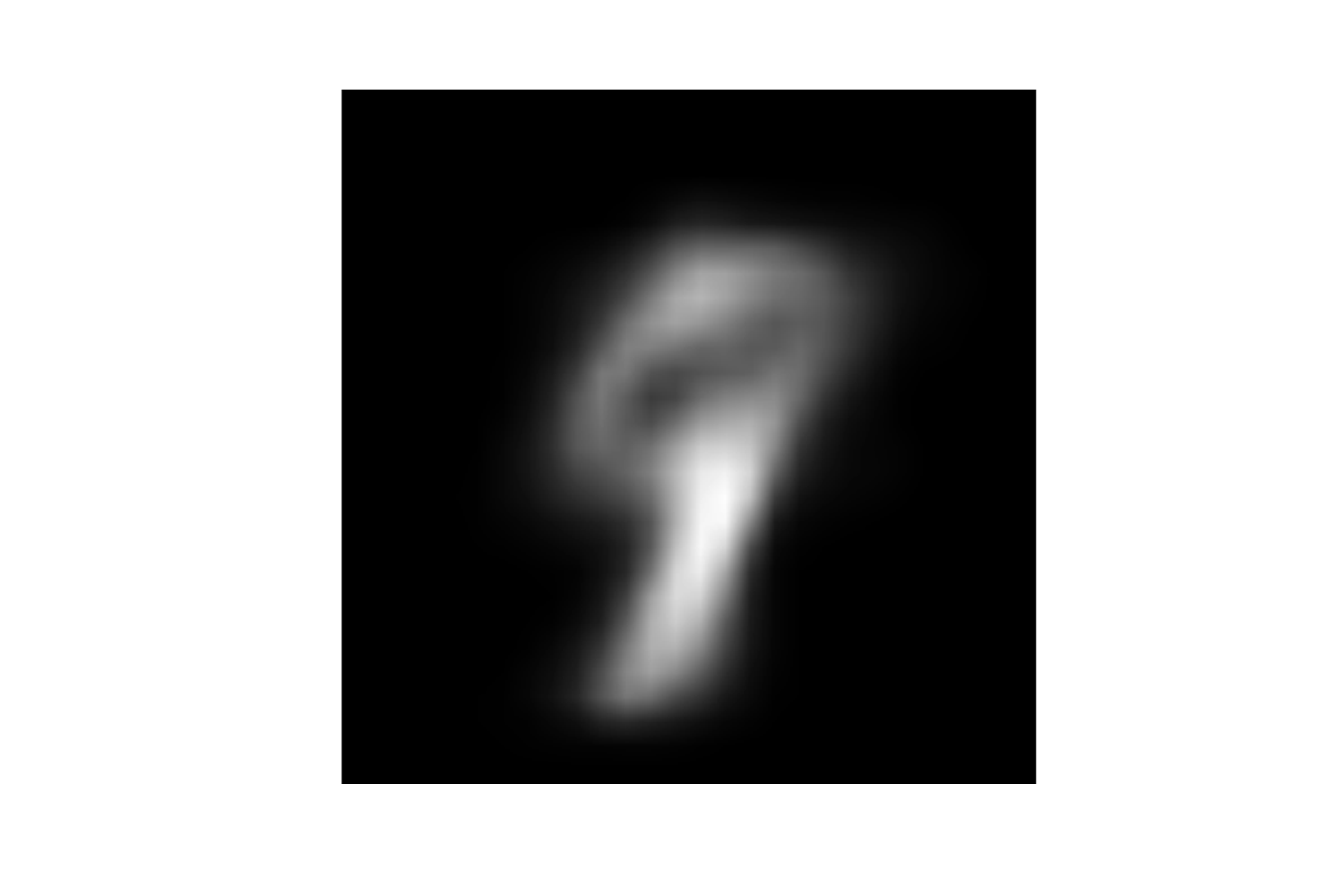}
 \hspace{-5mm}\includegraphics[scale=0.1]{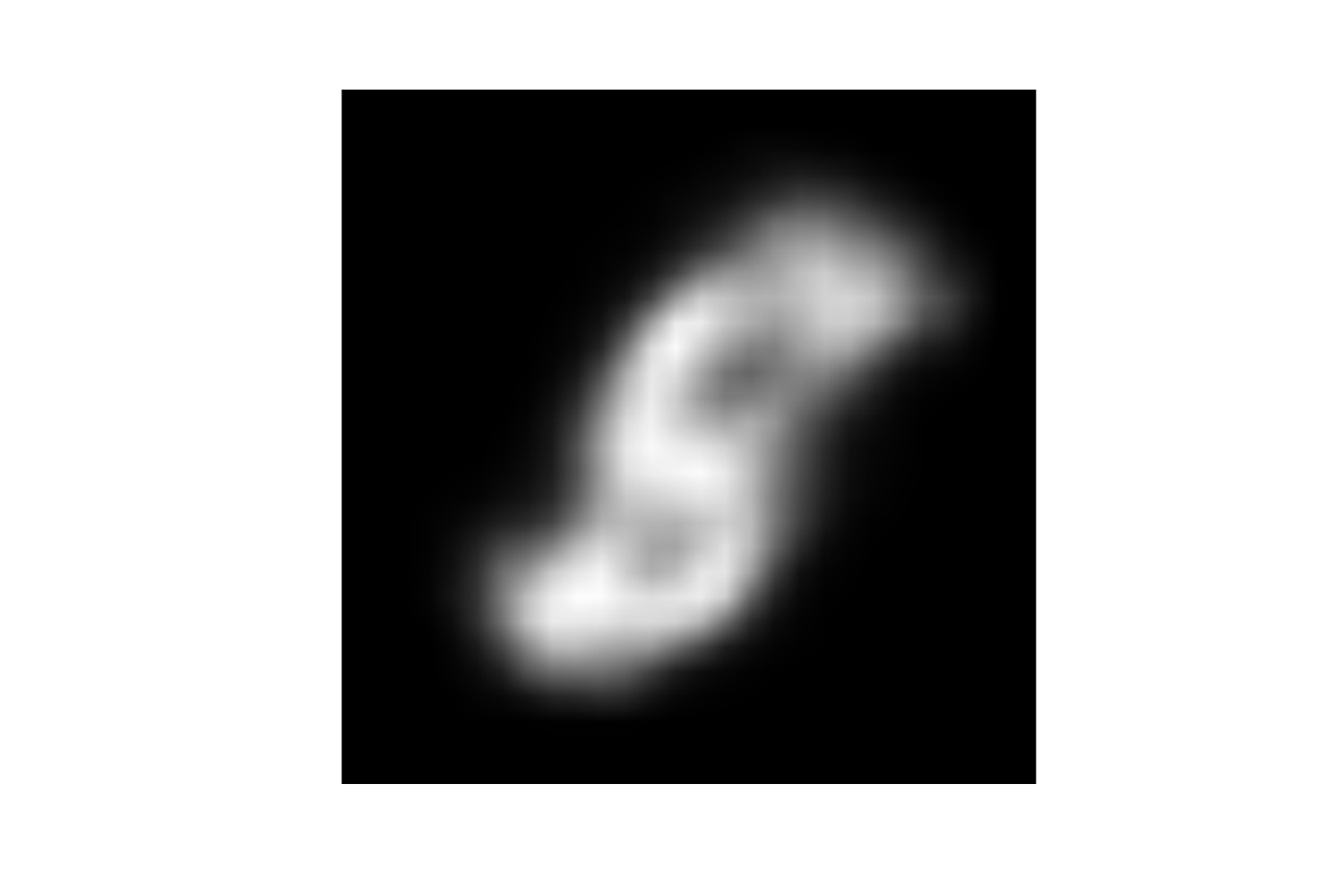}
 \hspace{-5mm}\includegraphics[scale=0.1]{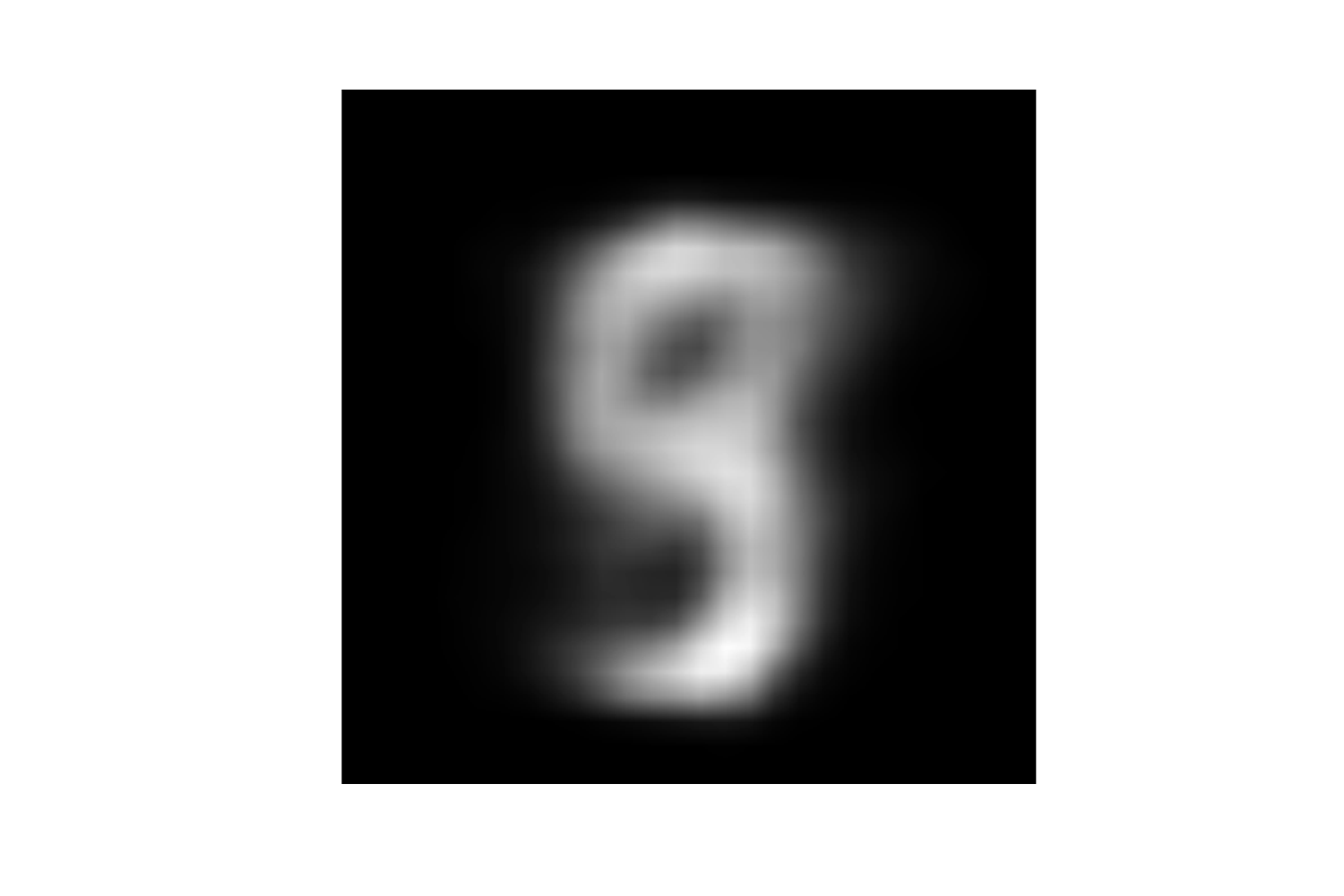}
 \hspace{-5mm}\includegraphics[scale=0.1]{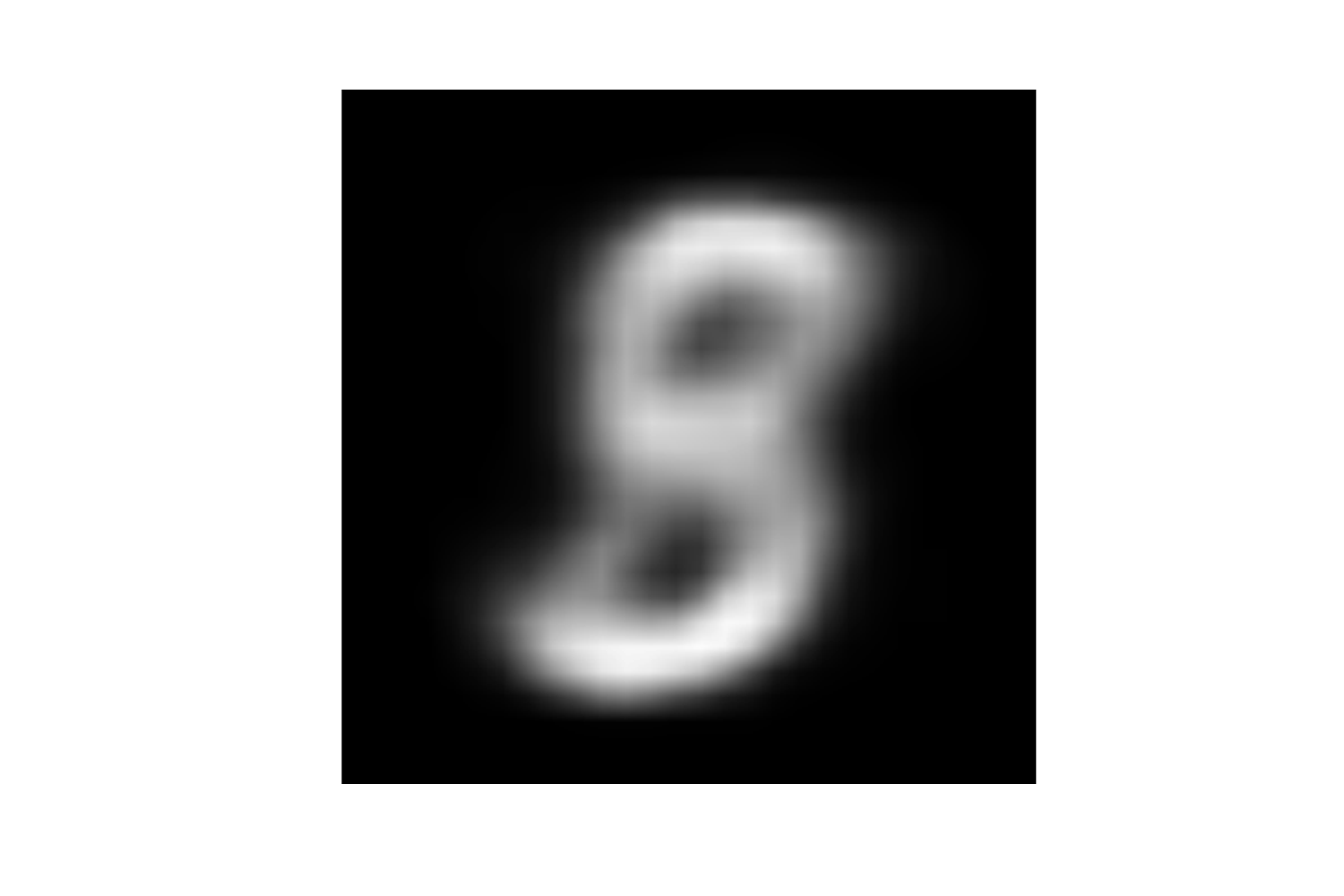}
 }
 \vspace{-1mm}
\\
\vspace{-2mm}
\subfigure[Infinite Mixture reconstruction (number of clusters $18$ using base VAE with number of hidden variables $50$)]{\label{fig:mnist_reconst_4}\includegraphics[scale=0.1]{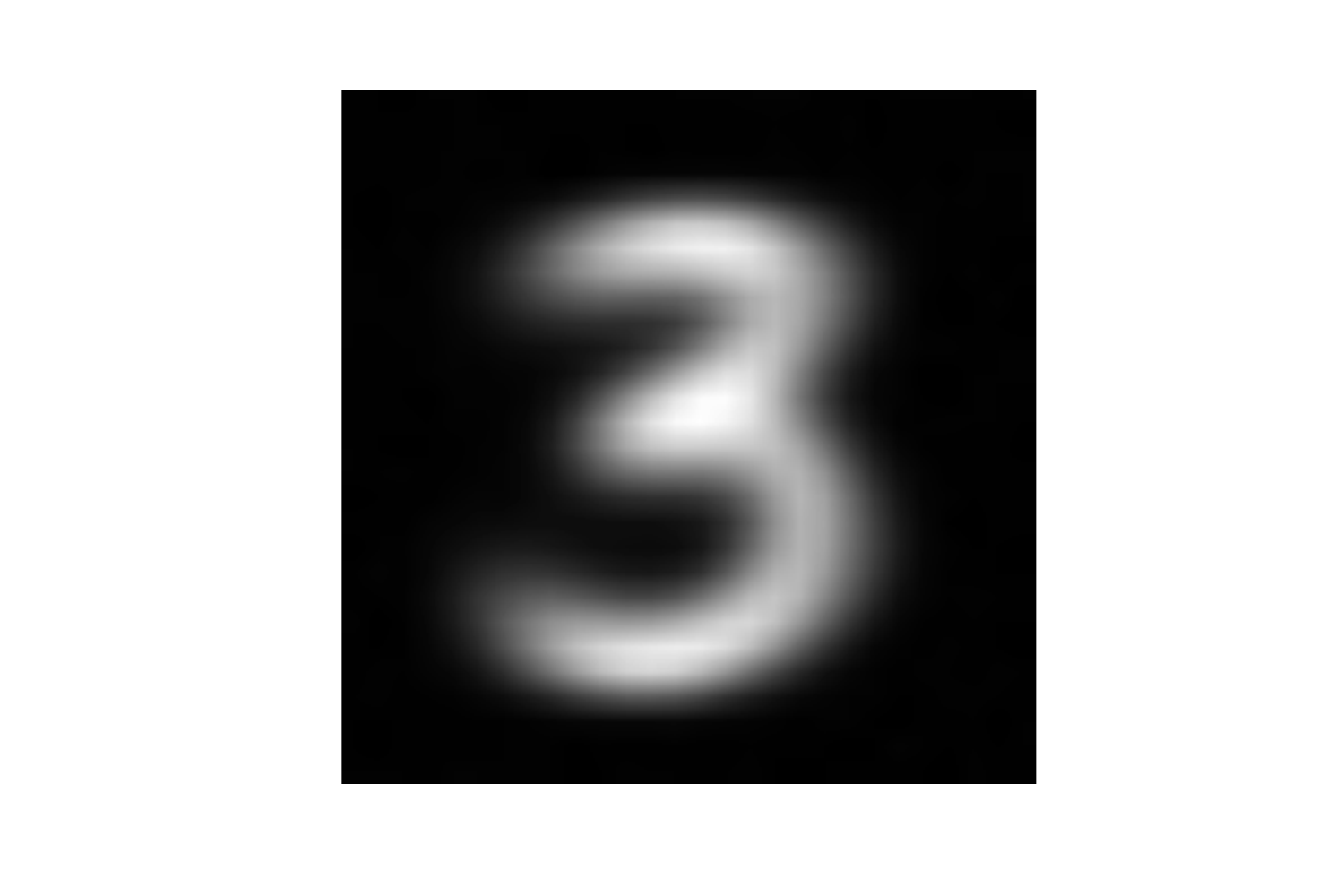}
 \hspace{-5mm}\includegraphics[scale=0.1]{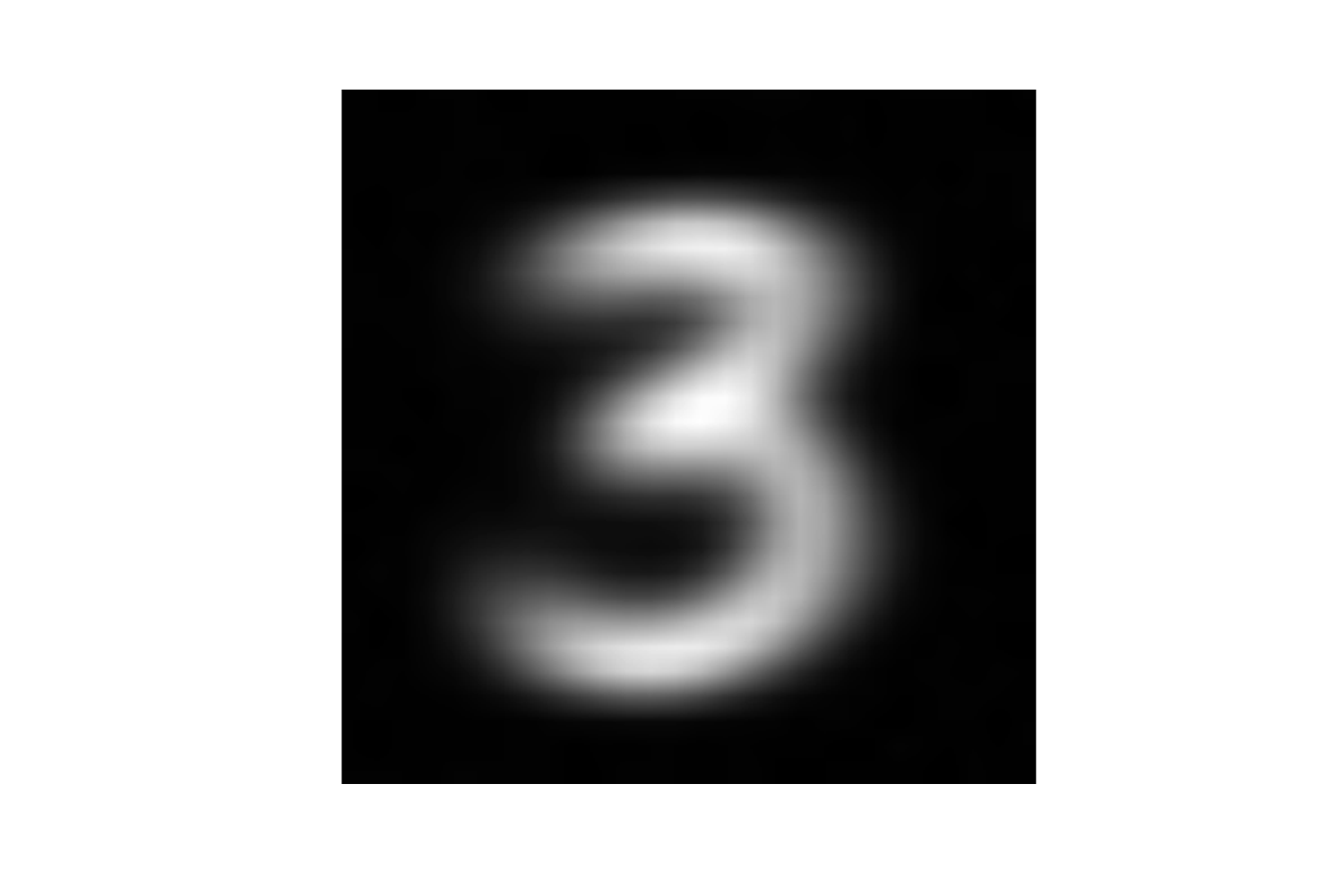}
 \hspace{-5mm}\includegraphics[scale=0.1]{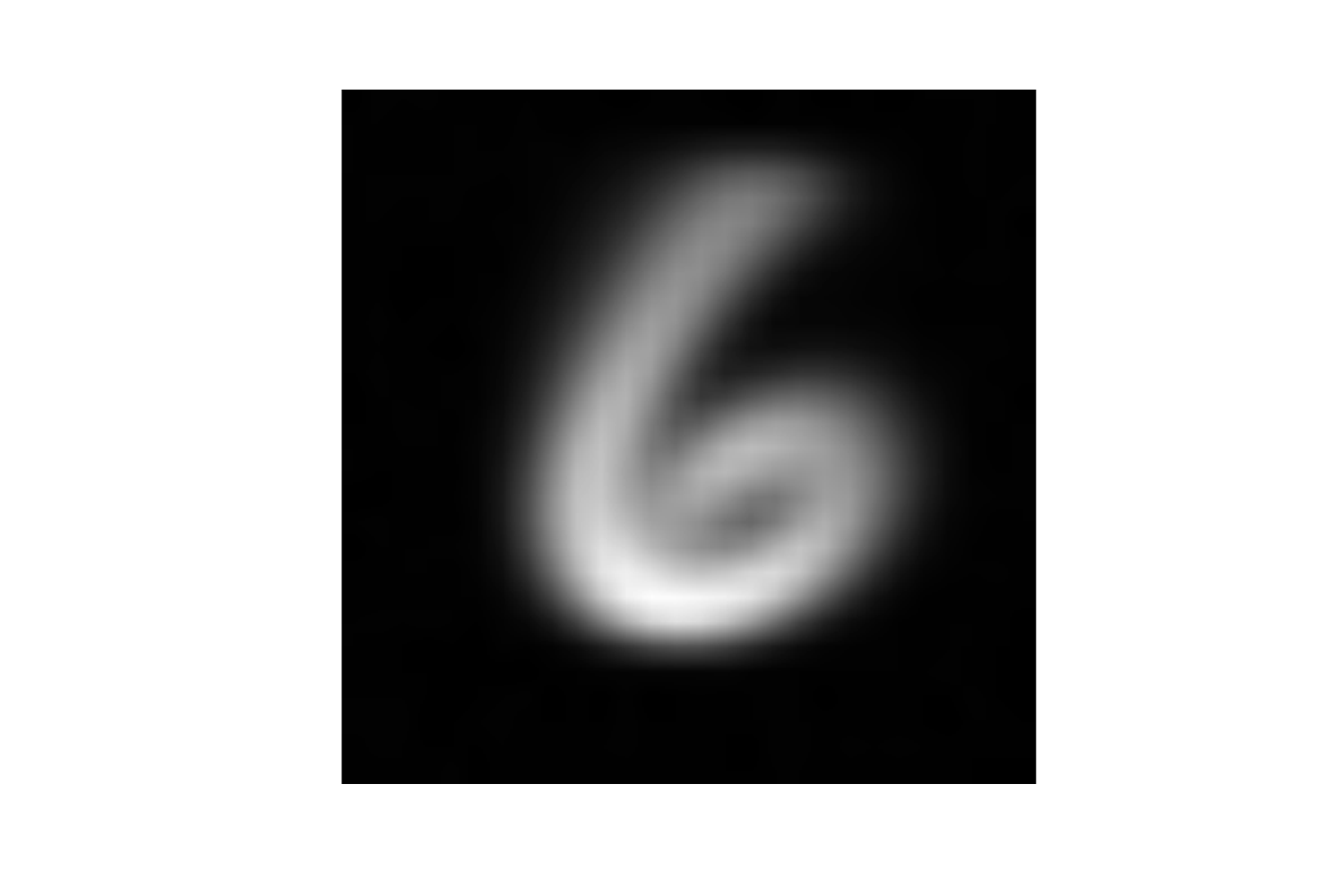}
 \hspace{-5mm}\includegraphics[scale=0.1]{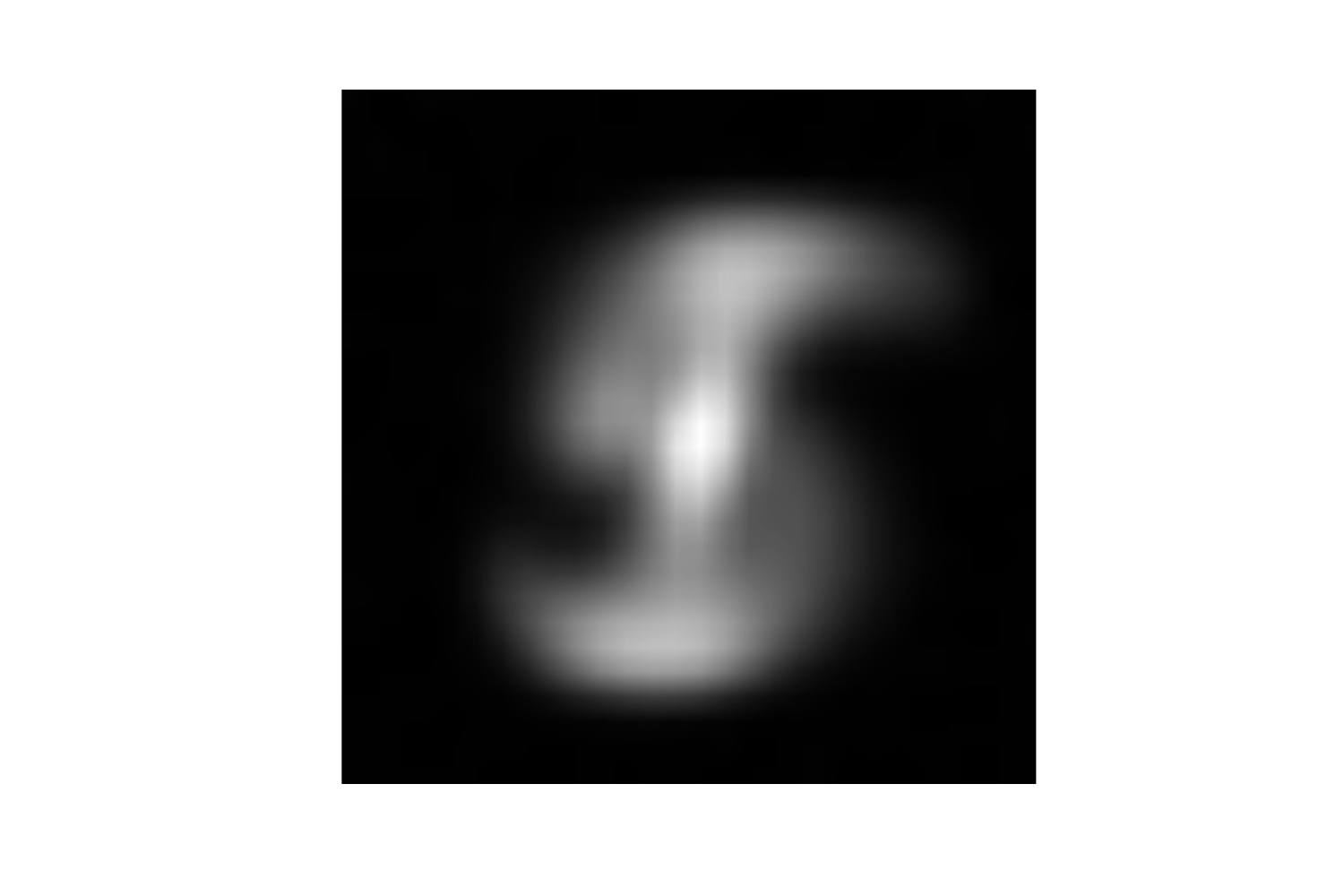}
 \hspace{-5mm}\includegraphics[scale=0.1]{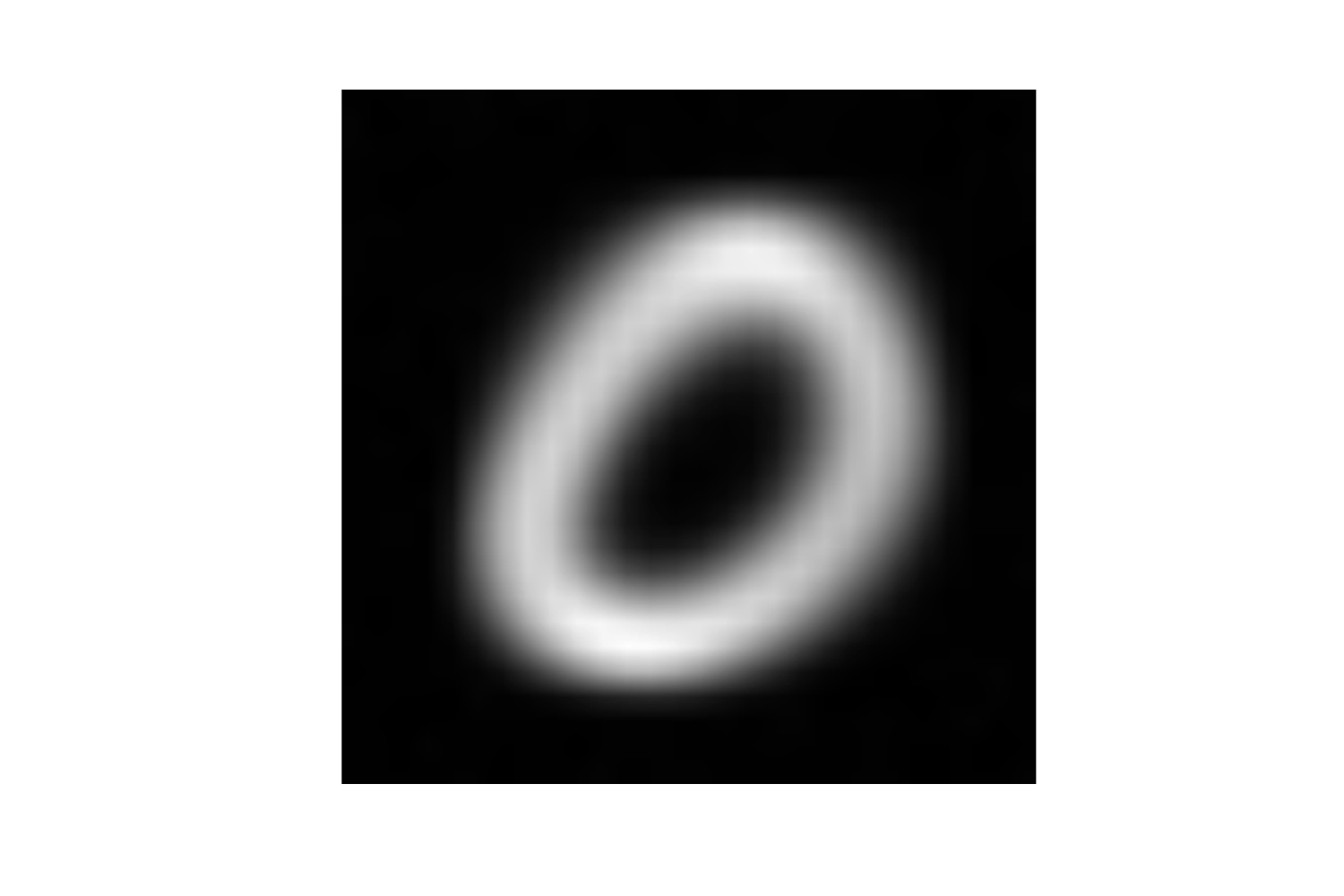}
 \hspace{-5mm}\includegraphics[scale=0.1]{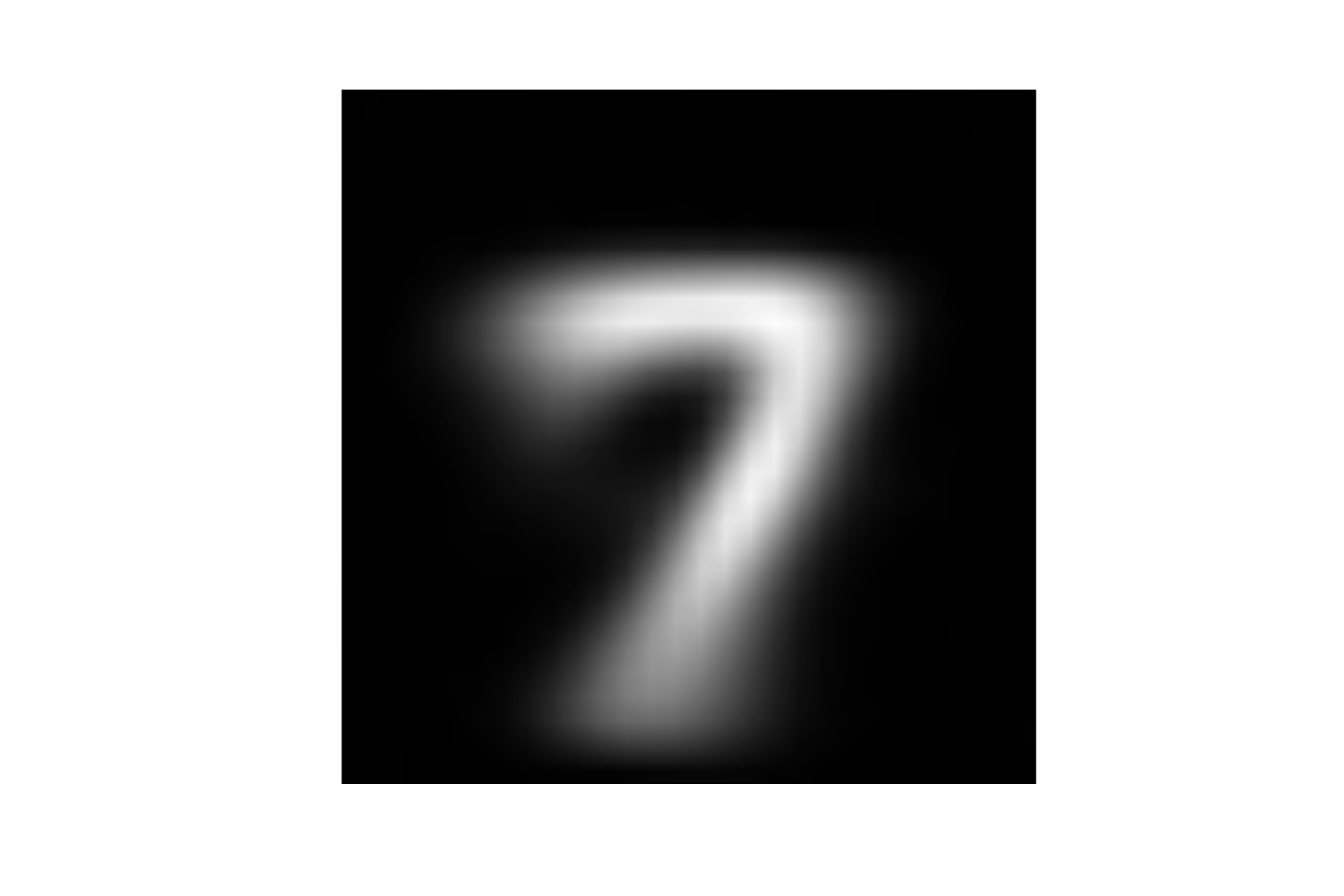}
 \hspace{-5mm}\includegraphics[scale=0.1]{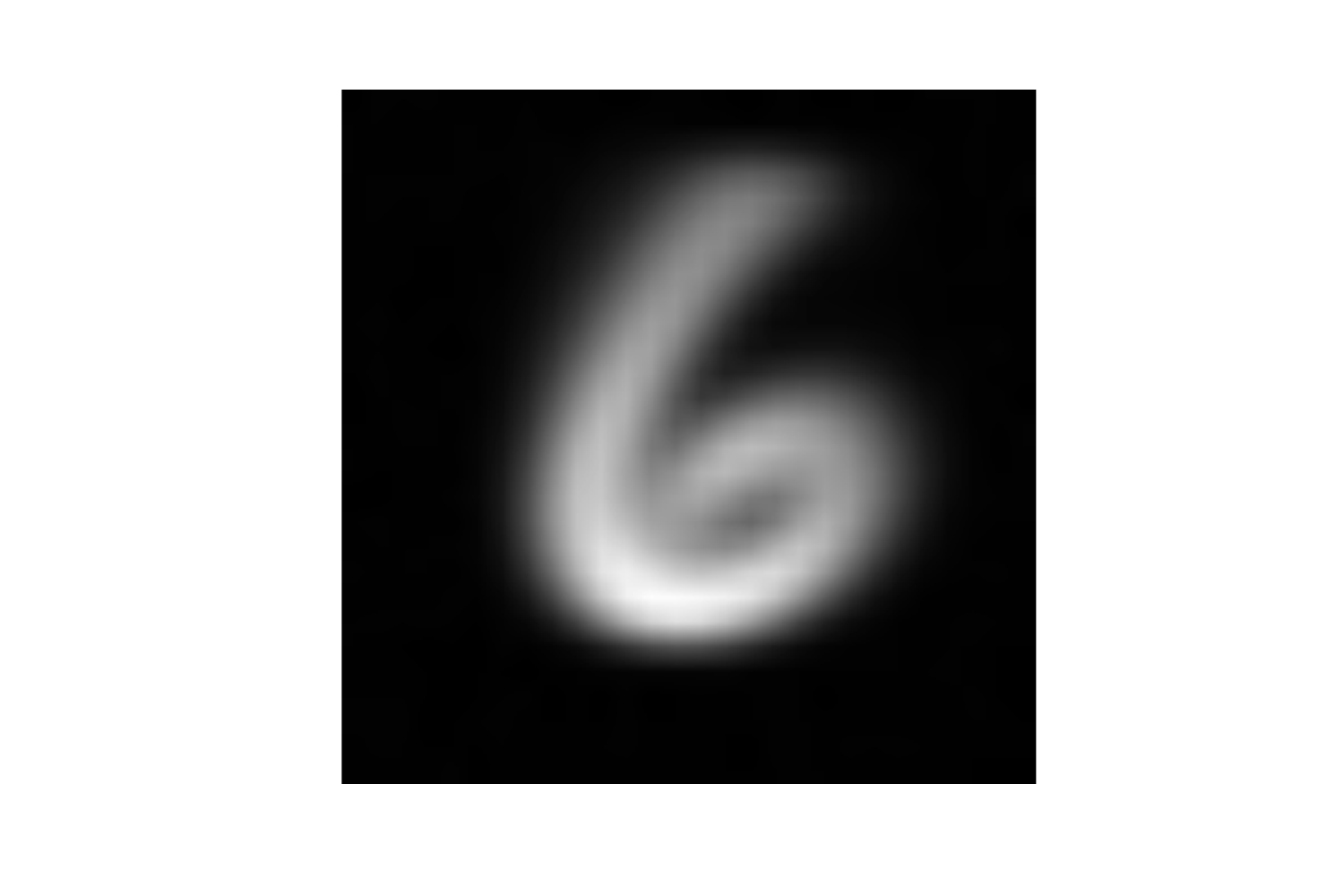}
 \hspace{-5mm}\includegraphics[scale=0.1]{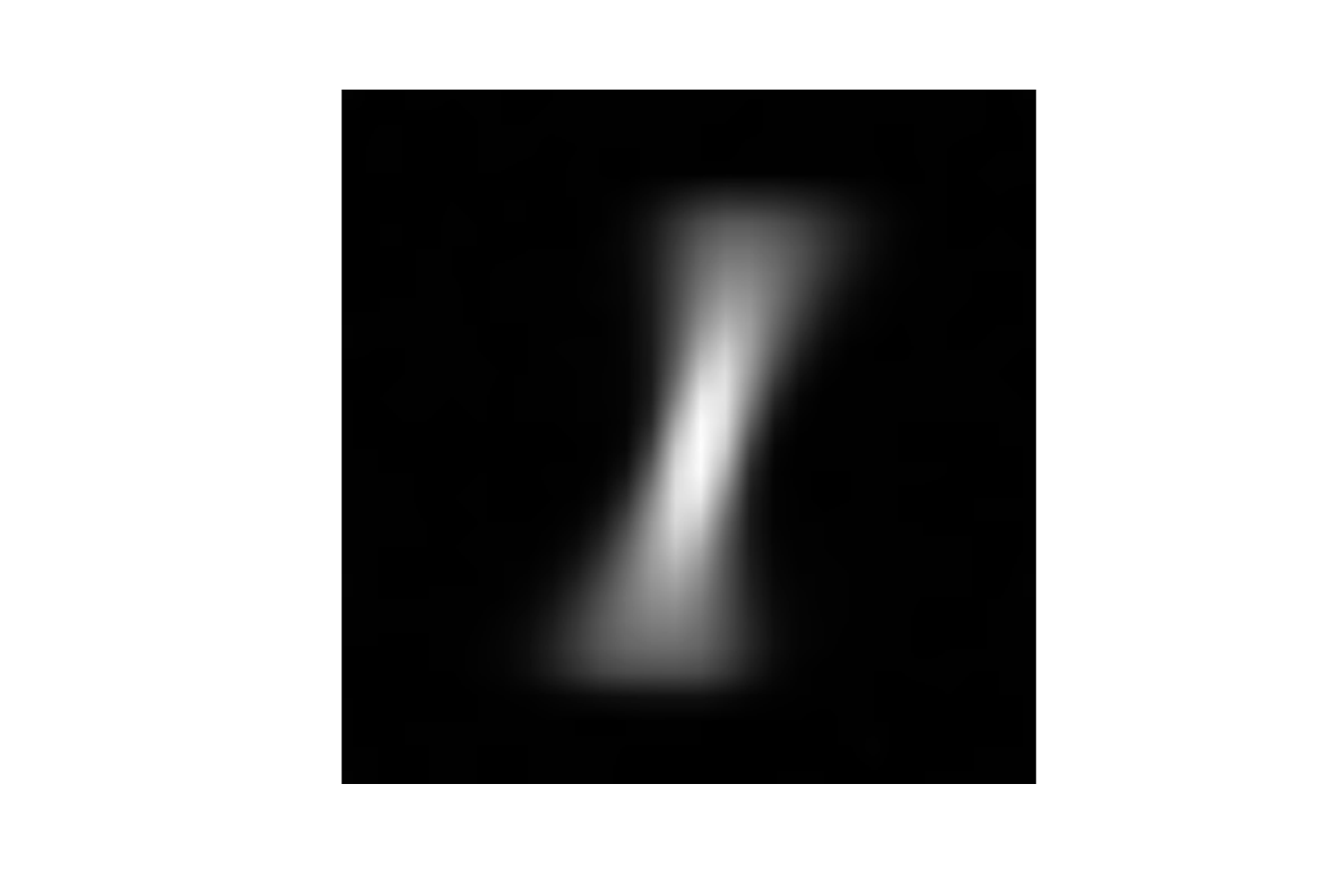}
 \hspace{-5mm}\includegraphics[scale=0.1]{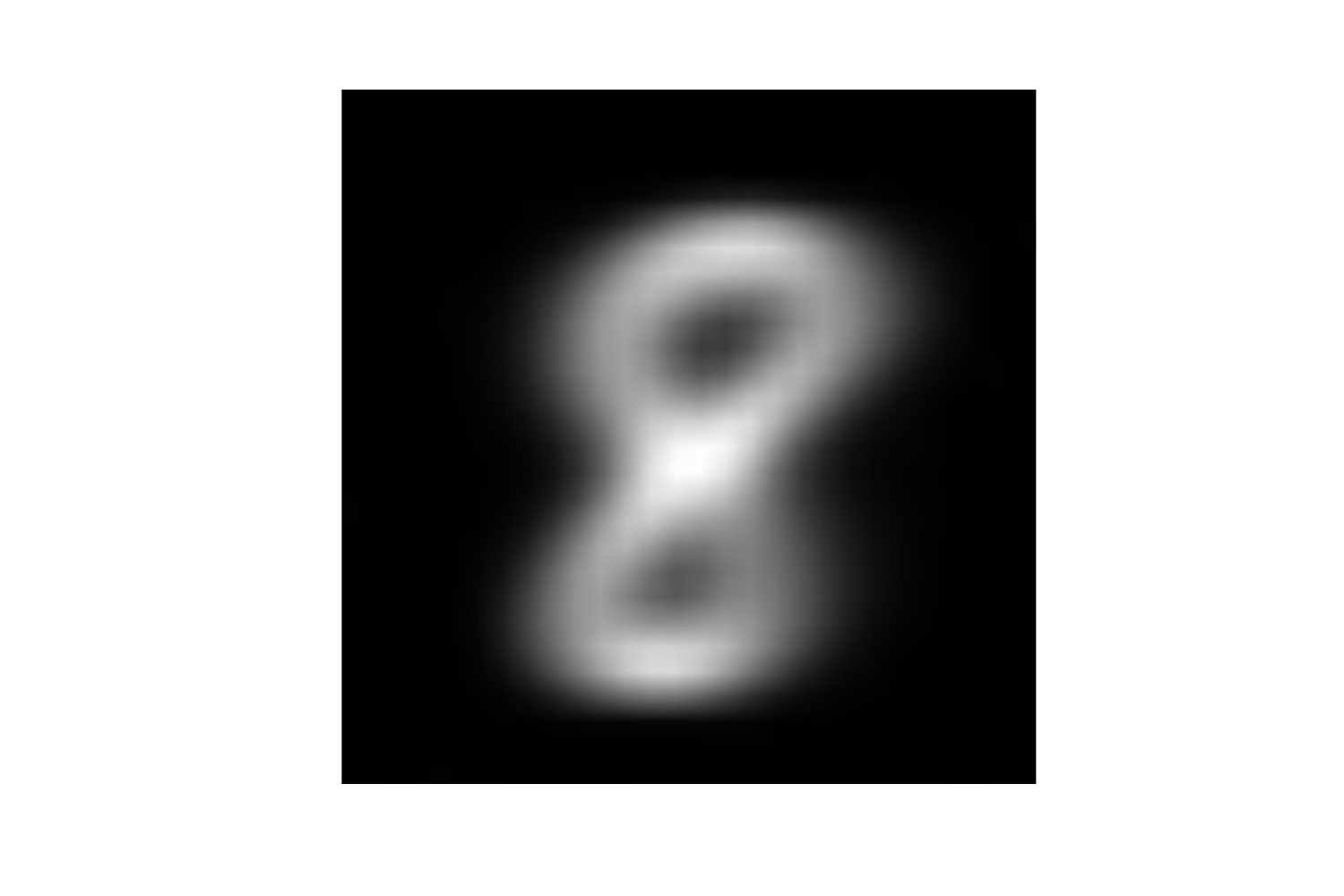}
 \hspace{-5mm}\includegraphics[scale=0.1]{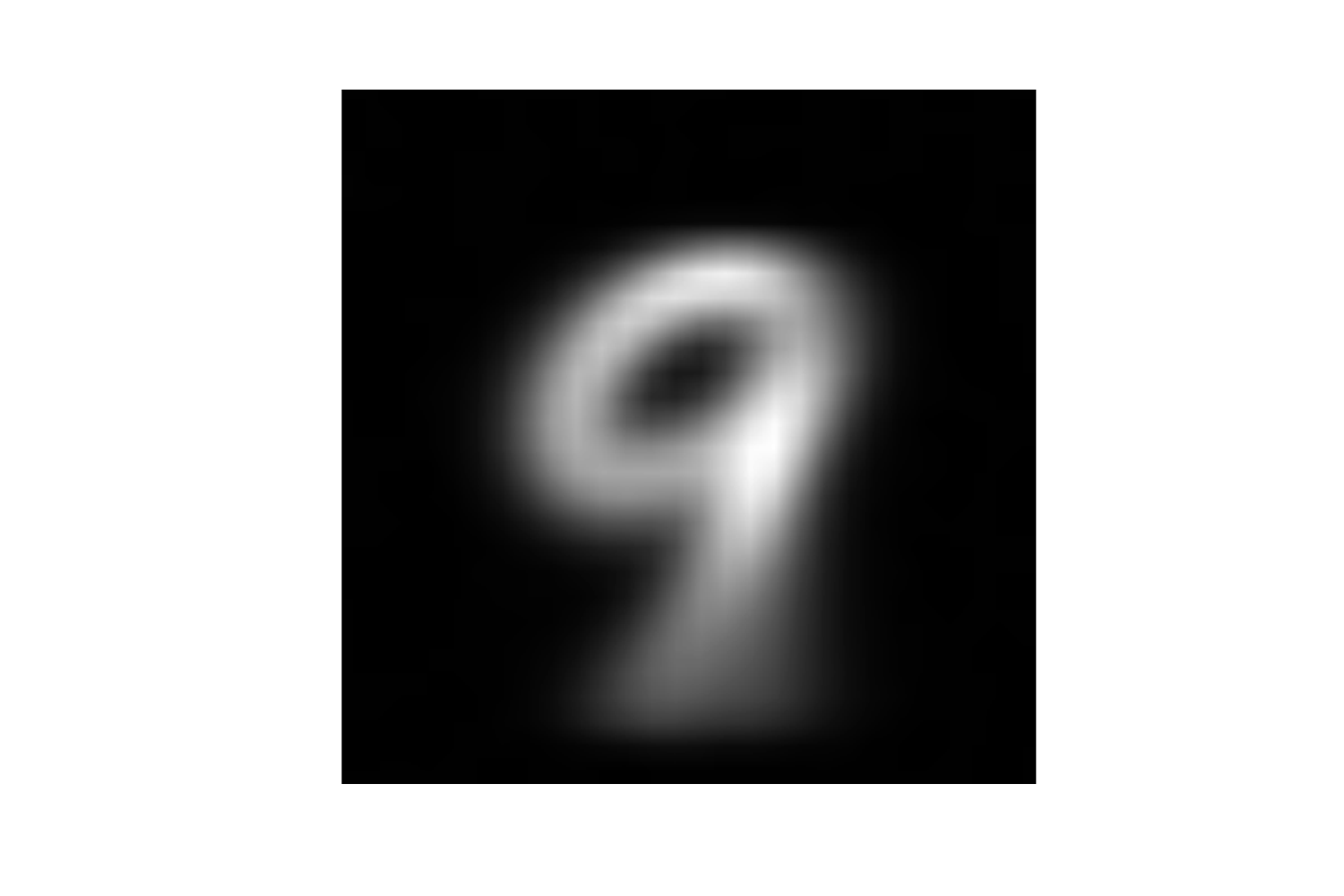}
 \hspace{-5mm}\includegraphics[scale=0.1]{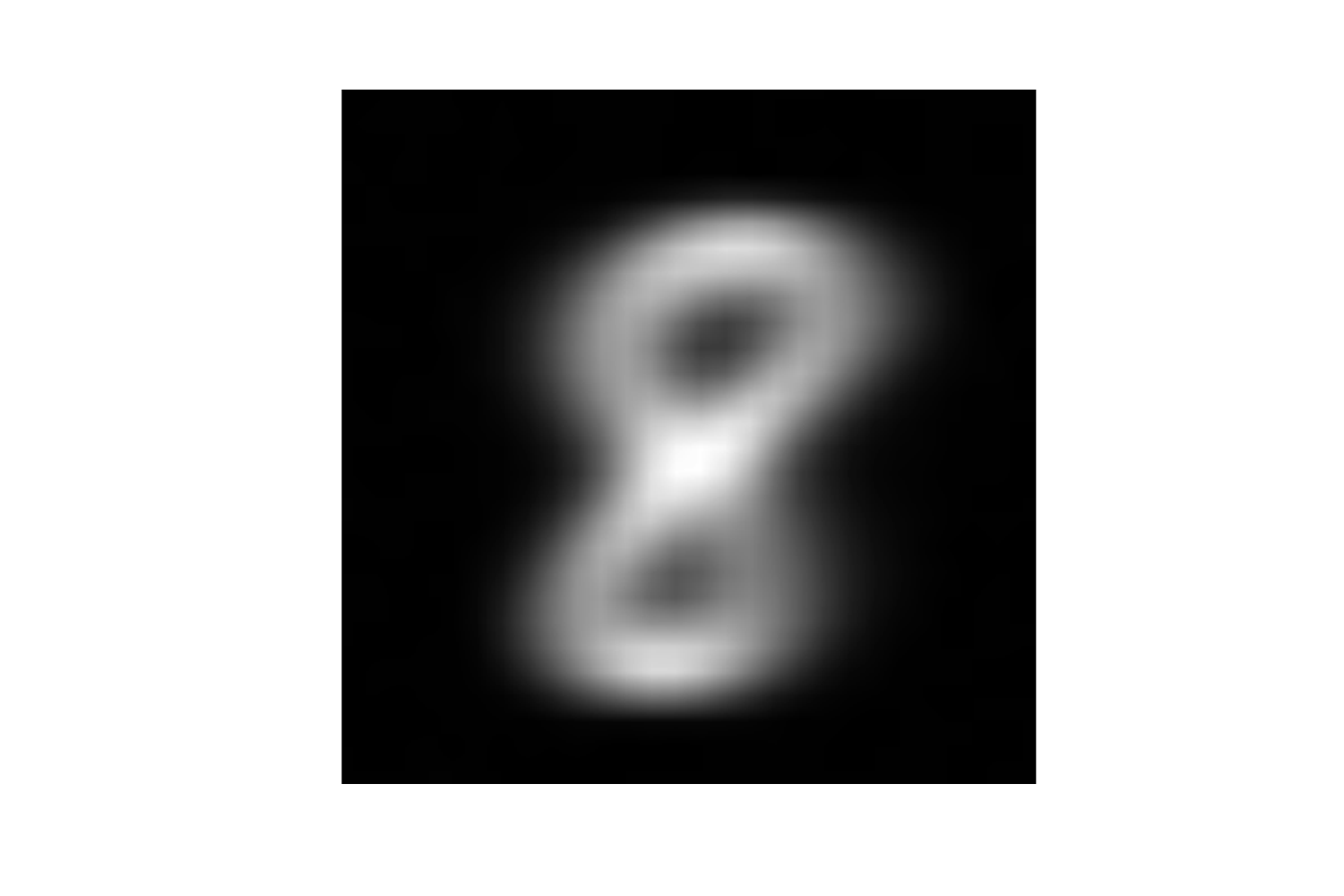}
 \hspace{-5mm}\includegraphics[scale=0.1]{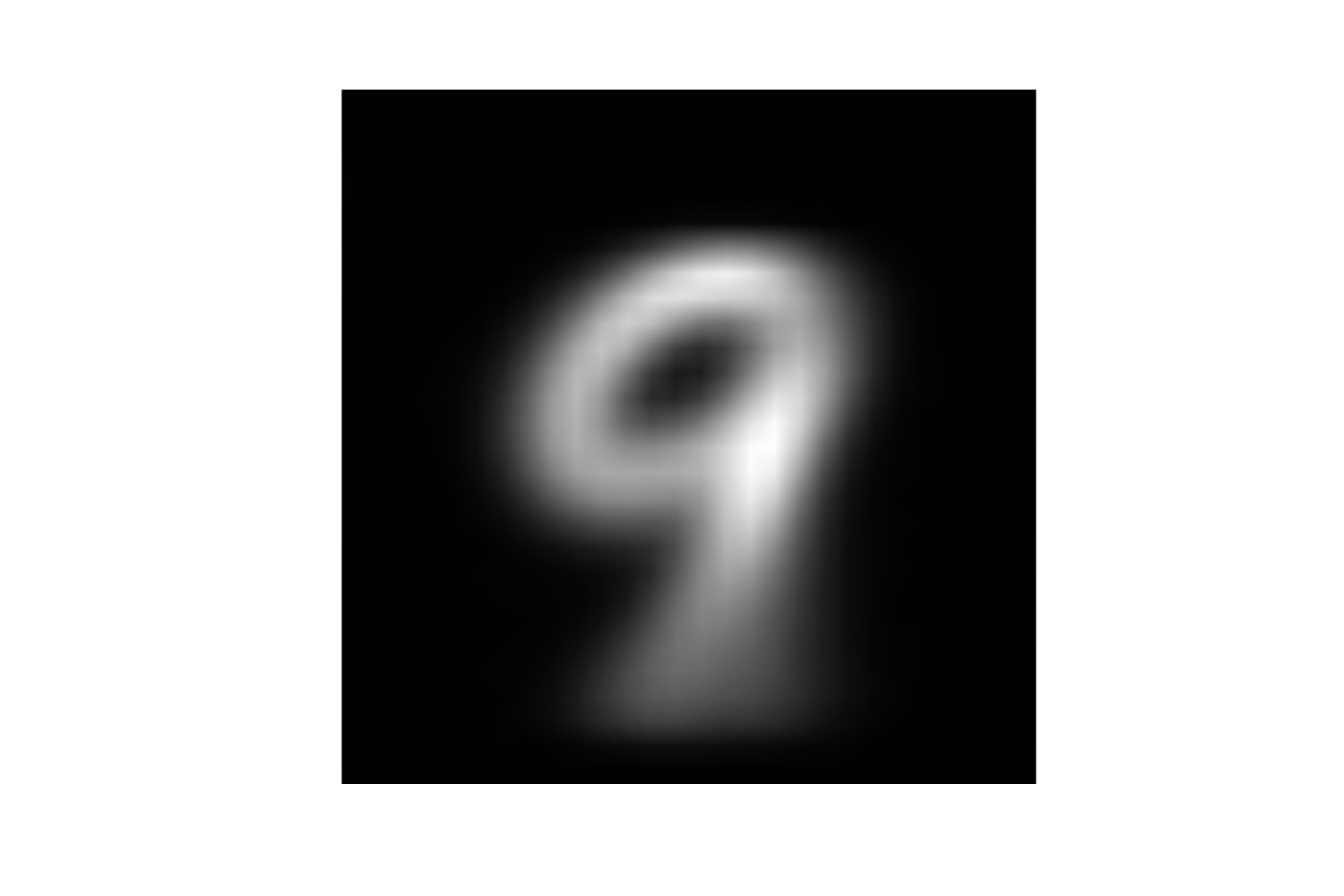}
 \hspace{-5mm}\includegraphics[scale=0.1]{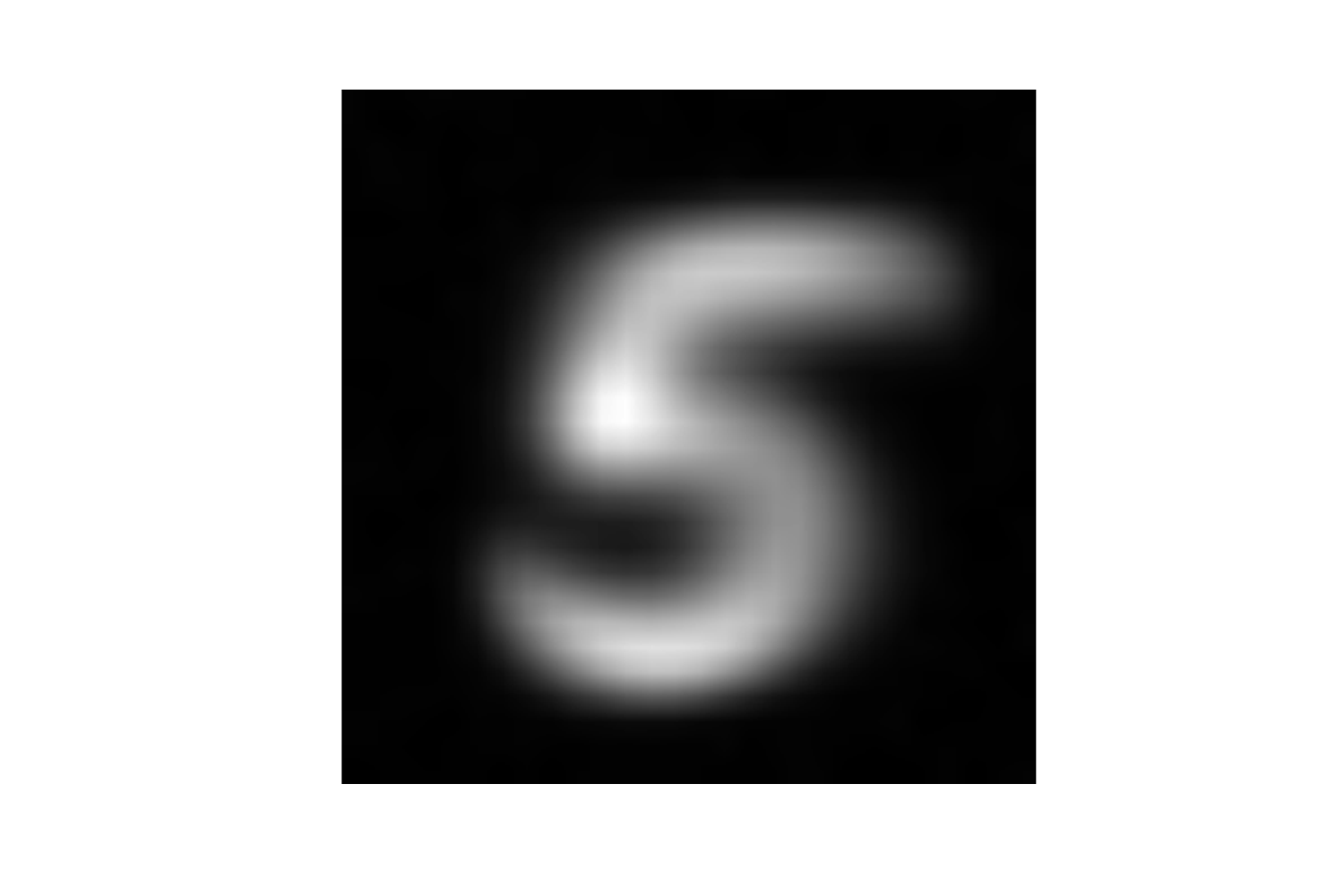}
 \hspace{-5mm}\includegraphics[scale=0.1]{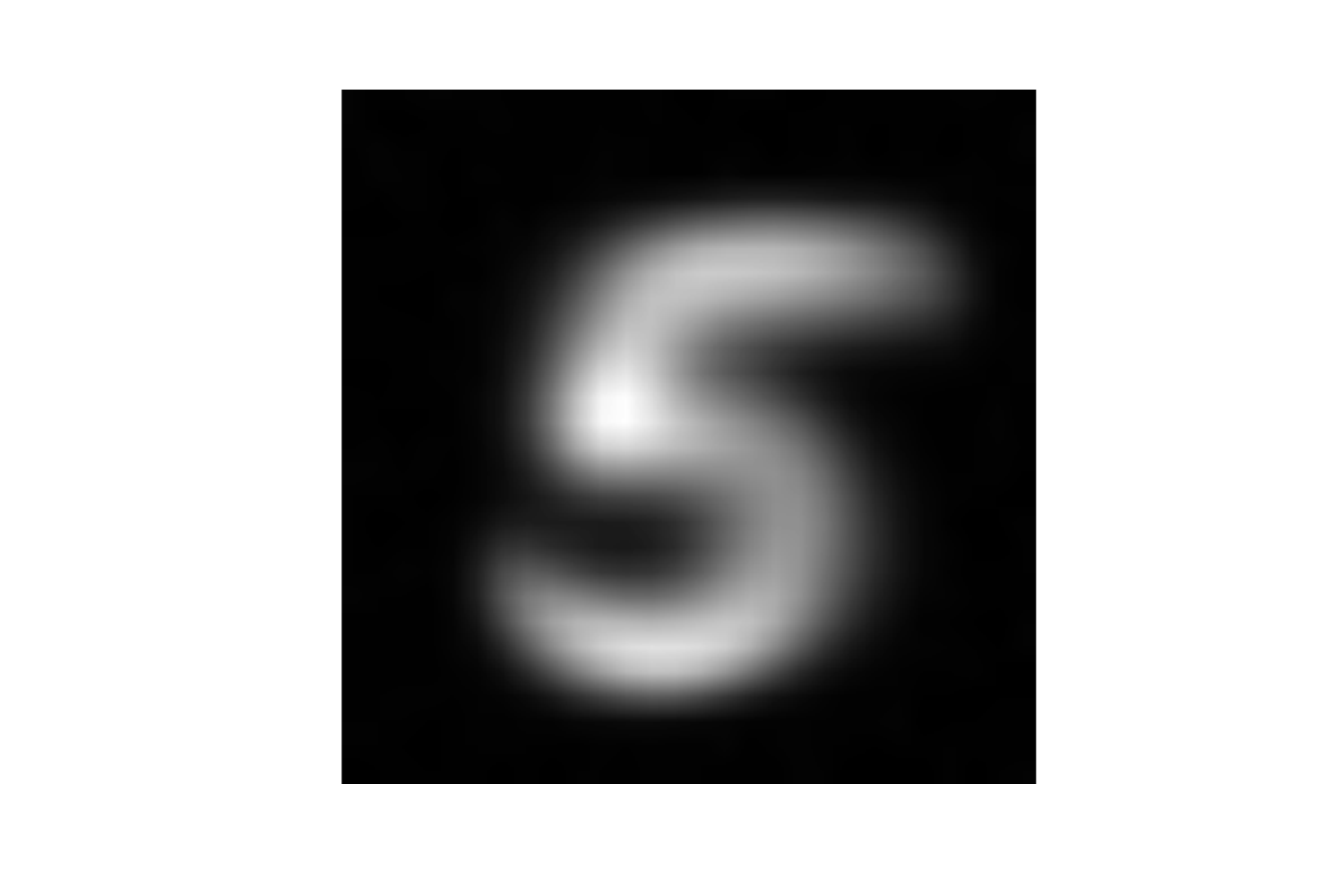}
 \hspace{-5mm}\includegraphics[scale=0.1]{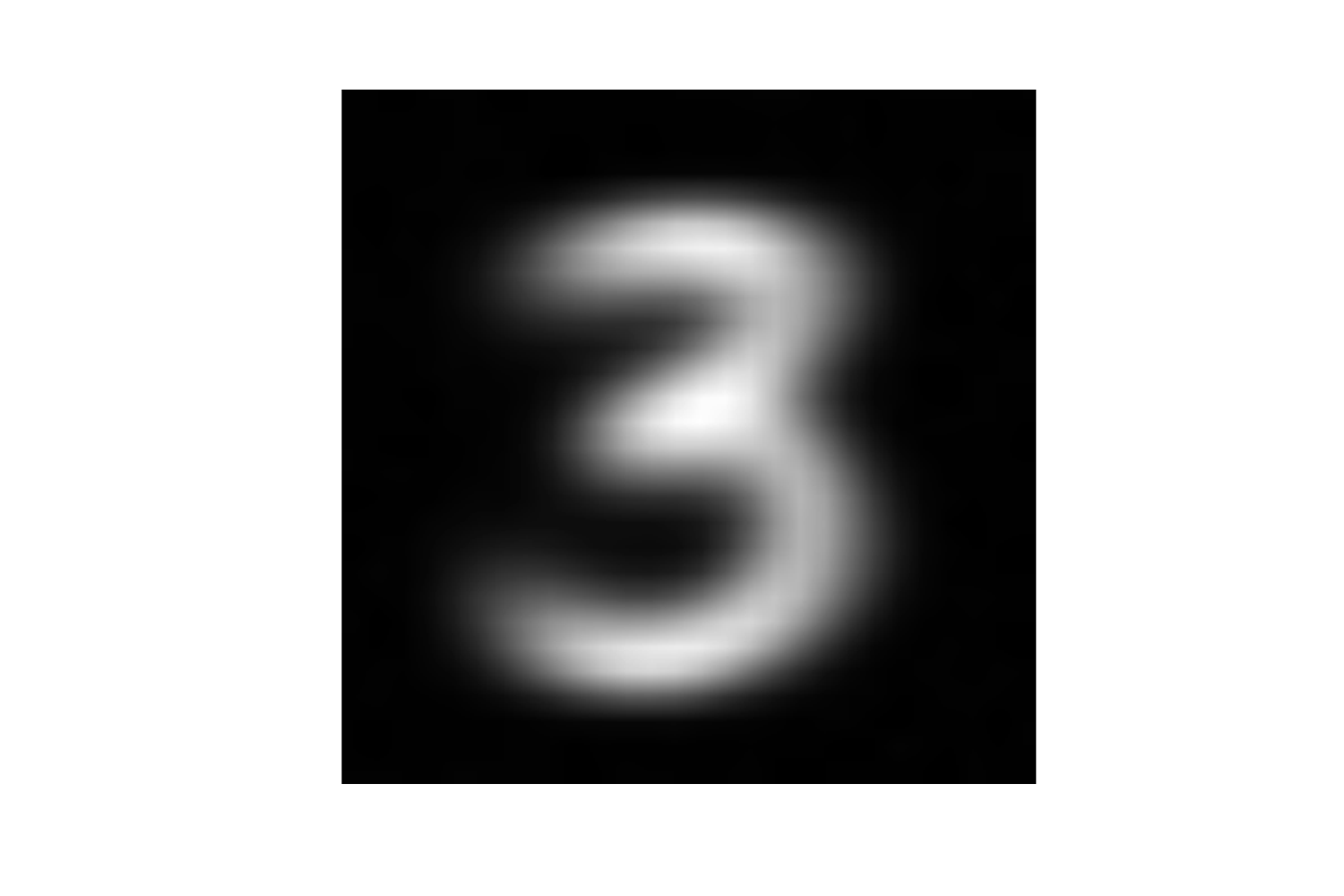}
 }
 \\
\end{tabular}
\caption{An illustration of the autoencoder's input reconstruction. First row
is the original images. Reconstructions in Figure \ref{fig:mnist_reconst_2}
and \ref{fig:mnist_reconst_3} are obtained from using a single VAE.
Images in the last row are obtained from the proposed mixture model
of $18$ VAEs each with $50$ hidden units. As seen, reconstructed
images are clearer in Figure \ref{fig:mnist_reconst_4}.}
\label{fig:mnist_recons}
\end{figure*}
MNIST dataset\footnote{http://yann.lecun.com/exdb/mnist/} contains
$60,000$ training and $10,000$ test images of size $28\times28$
of handwritten digits. Some random images from this dataset are shown
in Figure \ref{fig:mnist_reconst_1}. We use original VAE algorithm
(single VAE) with $100$ iteration and $50$ hidden variables to learn
a representation for these digits with binary distribution for the
input $p_{\btheta}(\bx|\bz)$. As shown in Figure \ref{fig:mnist_reconst_2},
these reconstructions are very unclear and at times wrong (6th column
where $7$ is wrongly reconstructed as $9$). Using this VAE as base,
we train an infinite mixture of our generative model. After $10$
iterations with $\alpha=2$, the expected reconstruction $\mathbb{E}\left[\bx\right]$
is depicted in Figure \ref{fig:mnist_reconst_4}. We use 2 samples
to compute $\mathbb{E}[\bx]$ for $c$th VAE. As observed, this reconstruction
is visually better and the mistake in the 6th column is fixed. Further,
Figure \ref{fig:mnist_reconst_3} shows using VAE with $1024$ hidden
units. It is interesting to note that even though our proposed model
has smaller number of hidden units ($900$ vs $1024$), the reconstruction
is better using our model.

In Table \ref{tbl:reconstruction} we summarize reconstruction error
(that is, $\|\bx-\mathbb{E}\left[\bx\right]\|$) for using our approach
versus the original VAE. As seen, our approach performs similarly
with the VAE when the number of hidden units are almost similar ($1000$
vs $1024$). As seen, with higher number of VAEs, we are able to reduce
the reconstruction error significantly. 

\begin{table}
\centering{}{\footnotesize{}}%
\begin{tabular}{|c|c|c|c|}
\hline 
{\footnotesize{}Method} & {\footnotesize{}$C$} & {\footnotesize{}\# hidden units} & {\footnotesize{}Error}\tabularnewline
\hline 
\hline 
\multirow{3}{*}{{\footnotesize{}Infinite Mixture}} & {\footnotesize{}2} & {\footnotesize{}100} & {\footnotesize{}$9.17$}\tabularnewline
\cline{2-4} 
 & {\footnotesize{}10} & {\footnotesize{}100} & {\footnotesize{}$5.12$}\tabularnewline
\cline{2-4} 
 & {\footnotesize{}17} & {\footnotesize{}100} & {\footnotesize{}$4.9$}\tabularnewline
\hline 
\multirow{2}{*}{{\footnotesize{}VAE}} & {\footnotesize{}1} & {\footnotesize{}100} & {\footnotesize{}$5.92$}\tabularnewline
\cline{2-4} 
 & {\footnotesize{}1} & {\footnotesize{}1024} & {\footnotesize{}$5.1$}\tabularnewline
\hline 
\end{tabular}\caption{Reconstruction error for MNIST dataset as the norm of the difference
of the input image and the expected reconstruction comparing our approach
with the original VAE. }
\label{tbl:reconstruction}
\end{table}
To test our approach in a semi-supervised setting, we use a deep Convolutional
Neural Net (CNN). Our deep CNN architecture consists of two convolutional
layers with $32$ filters of $5\times5$ and Rectified Linear Unit
(ReLU) activation and max-pooling of $2\times2$ after each one. We
added a fully connected layer with $256$ hidden units followed by
a dropout layer and then the softmax output layer. As shown in Table
\ref{tbl:mnist_semi}, our infinite mixture with $17$ base VAEs has
been able to outperform most of the state-of-the-art methods. Only
recently proposed Virtual Adversarial Network {\footnotesize{}\cite{MiyatoMaedaKoyamaEtAl2016}
}performs better than ours with small training examples.

\begin{table}
{\footnotesize{}}%
\begin{tabular}{|c|c|c|c|}
\hline 
{\footnotesize{}Method/Labels} & {\footnotesize{}100} & {\footnotesize{}1000} & {\footnotesize{}All}\tabularnewline
\hline 
\hline 
{\footnotesize{}Pseudo-label \cite{Lee2013}} & {\footnotesize{}$10.49$} & {\footnotesize{}$3.64$} & {\footnotesize{}$0.81$}\tabularnewline
\hline 
{\footnotesize{}EmbedNN \cite{WestonRatleMobahiEtAl2012}} & {\footnotesize{}$16.9$} & {\footnotesize{}$5.73$} & {\footnotesize{}$3.59$}\tabularnewline
\hline 
{\footnotesize{}DGN \cite{KingmaMohamedRezendeEtAl2014}} & {\footnotesize{}$3.33\pm0.14$} & {\footnotesize{}$2.40\pm0.02$} & {\footnotesize{}$0.96$}\tabularnewline
\hline 
{\footnotesize{}Adversarial \cite{GoodfellowPouget-AbadieMirzaEtAl2014}} &  &  & {\footnotesize{}$0.78$}\tabularnewline
\hline 
{\footnotesize{}Virtual Adversarial \cite{MiyatoMaedaKoyamaEtAl2016}} & {\footnotesize{}$2.66$} & {\footnotesize{}$1.50$} & {\footnotesize{}$0.64\pm0.03$}\tabularnewline
\hline 
{\footnotesize{}AtlasRBF \cite{PitelisRussellAgapito2014}} & {\footnotesize{}$8.10\pm0.95$} & {\footnotesize{}$3.68\pm0.12$} & {\footnotesize{}$1.31$}\tabularnewline
\hline 
{\footnotesize{}PEA \cite{BachmanAlsharifPrecup2014}} & {\footnotesize{}$5.21$} & {\footnotesize{}$2.64$} & {\footnotesize{}$2.30$}\tabularnewline
\hline 
{\footnotesize{}$\Gamma\text{-Model}$ \cite{RasmusValpolaHonkalaEtAl2015}} & {\footnotesize{}$4.34\pm2.31$} & {\footnotesize{}$1.71\pm0.07$} & {\footnotesize{}$0.79\pm0.05$}\tabularnewline
\hline 
\hline 
{\footnotesize{}Baseline CNN} & {\footnotesize{}$8.62\pm1.87$} & {\footnotesize{}$4.16\pm0.35$} & \textbf{\footnotesize{}$0.68\pm0.02$}\tabularnewline
\hline 
{\footnotesize{}Infinite Mixture} & {\footnotesize{}$3.93\pm0.5$} & \textbf{\footnotesize{}$2.29\pm0.2$} & \textbf{\footnotesize{}$0.6\pm0.02$}\tabularnewline
\hline 
\end{tabular}\caption{Test error for MNIST with 17 clusters and 100 hidden variables. Only
\cite{MiyatoMaedaKoyamaEtAl2016} reports better performance than
ours }
\label{tbl:mnist_semi}
\end{table}

\begin{figure*}
\centering{}\subfigure{\includegraphics[scale=0.18]{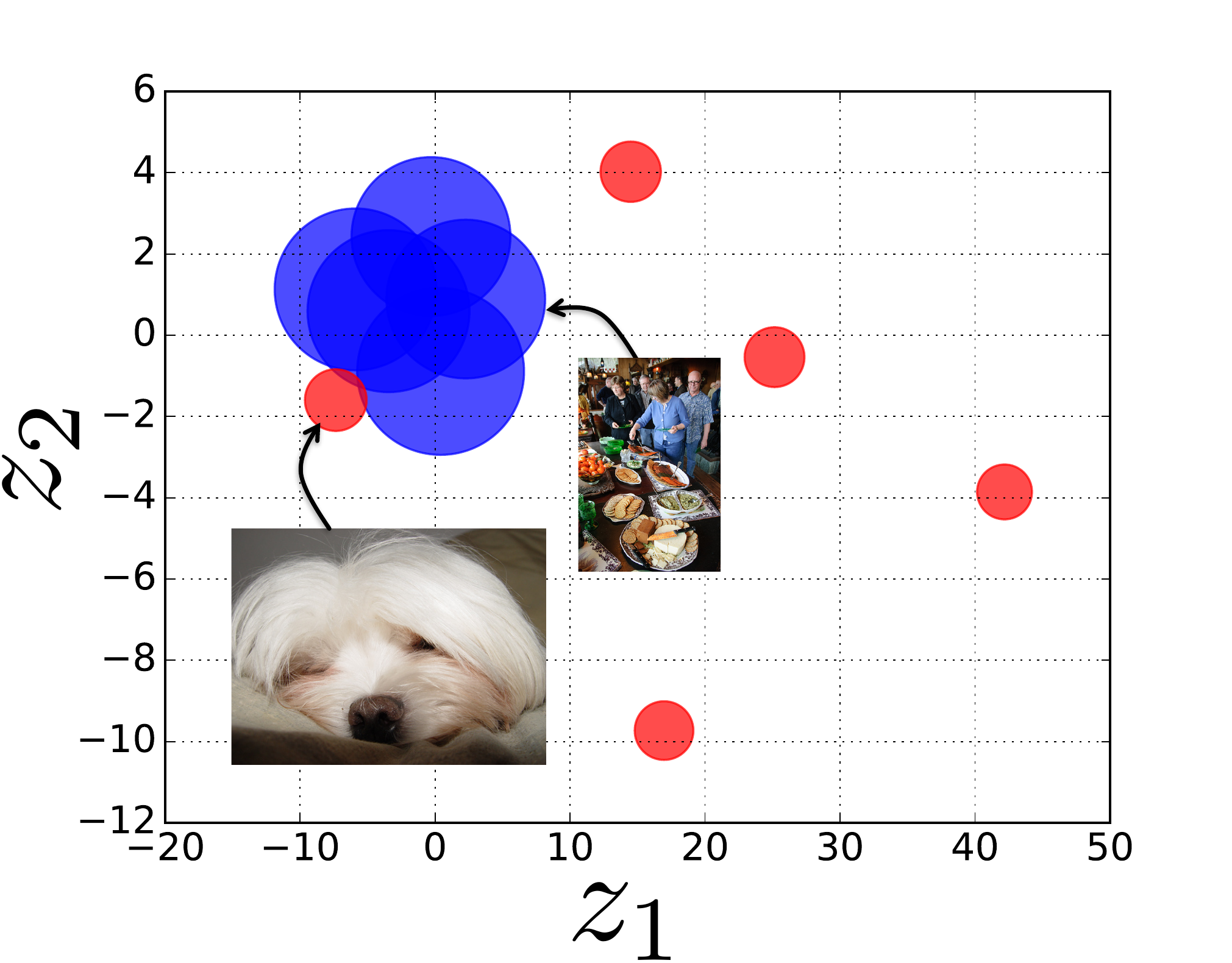}}\hspace{-3mm}\subfigure{\includegraphics[scale=0.18]{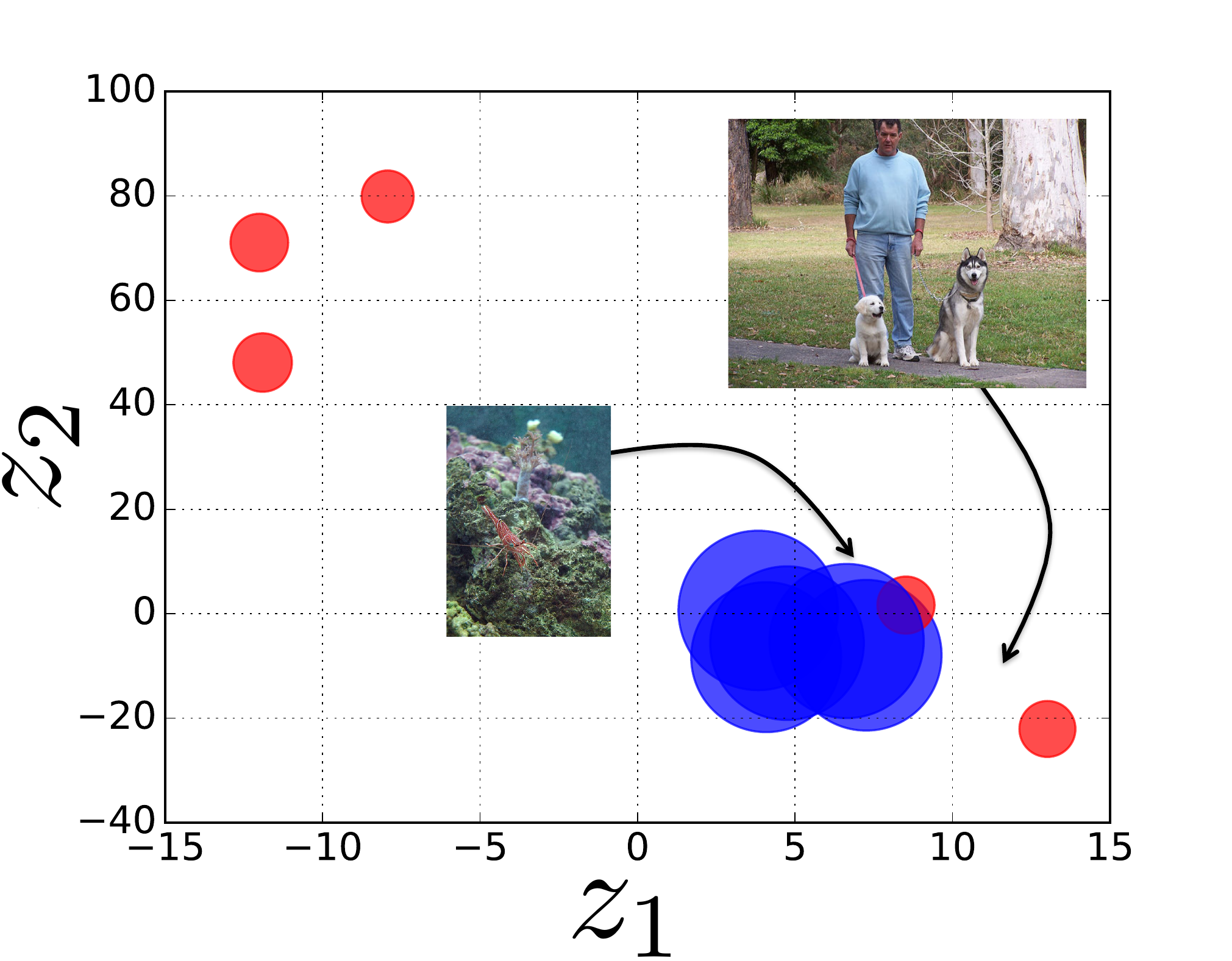}}\hspace{-3mm}\subfigure{\includegraphics[scale=0.18]{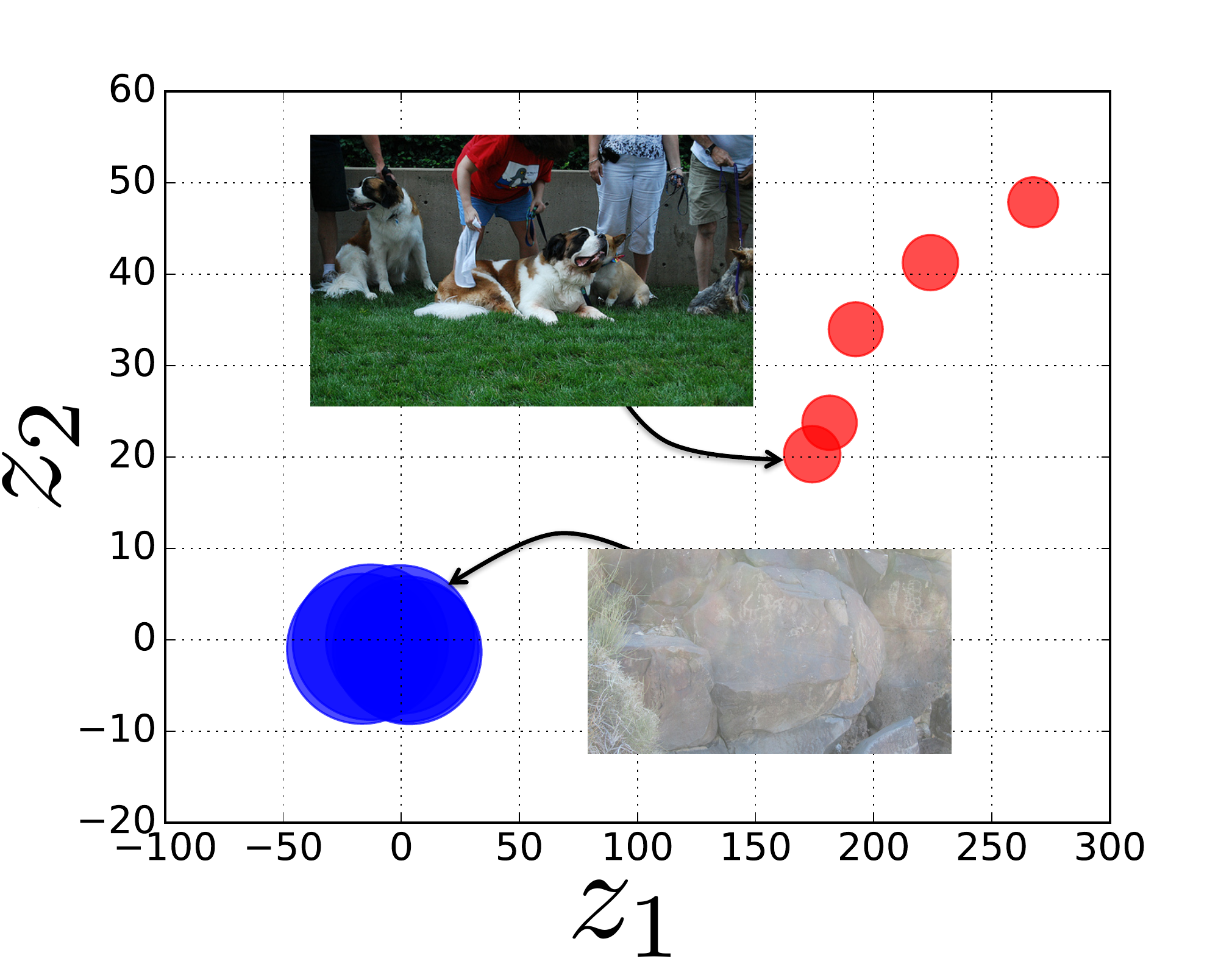}}\hspace{-3mm}\subfigure{\includegraphics[scale=0.18]{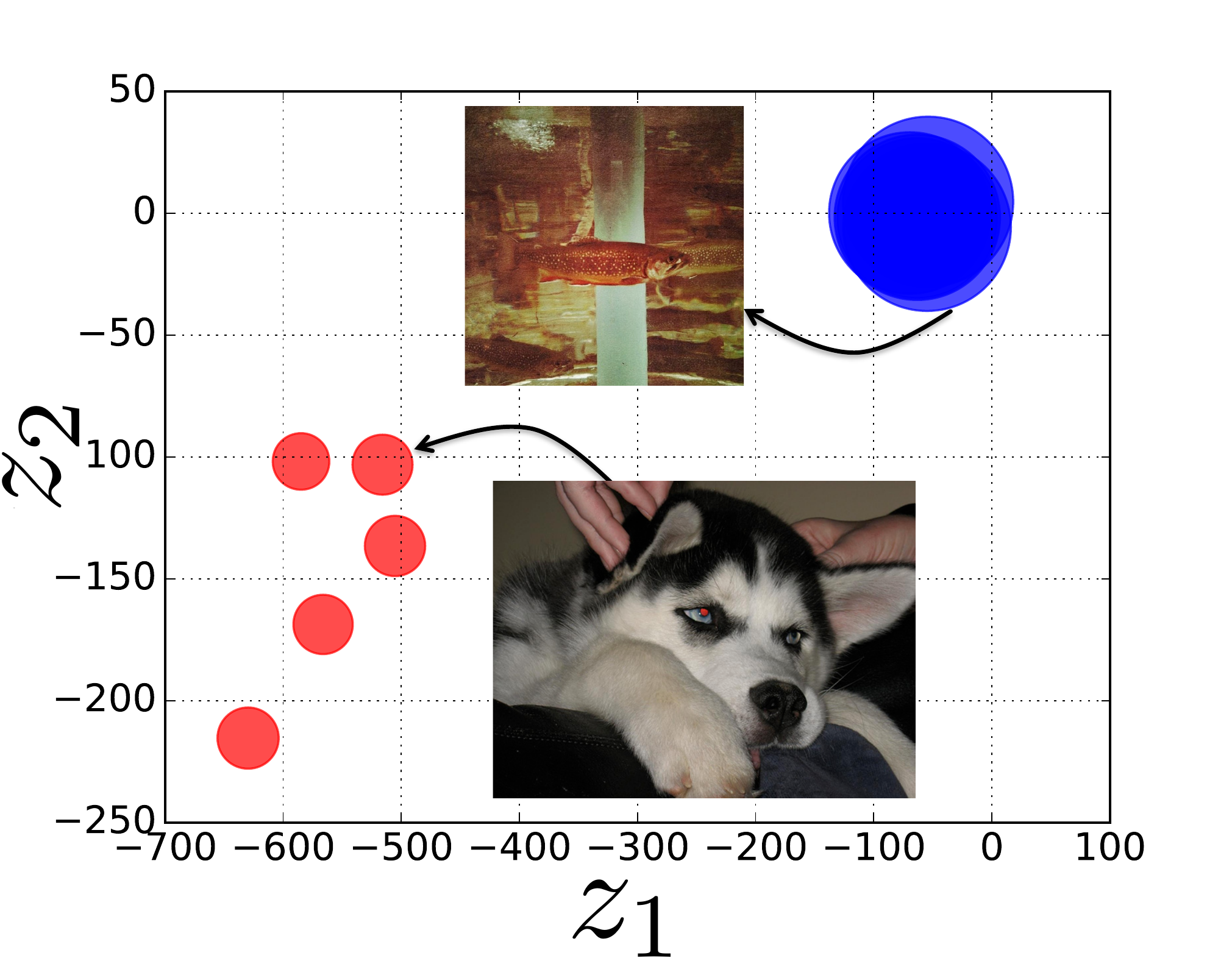}}\hspace{-3mm}\subfigure{\includegraphics[scale=0.18]{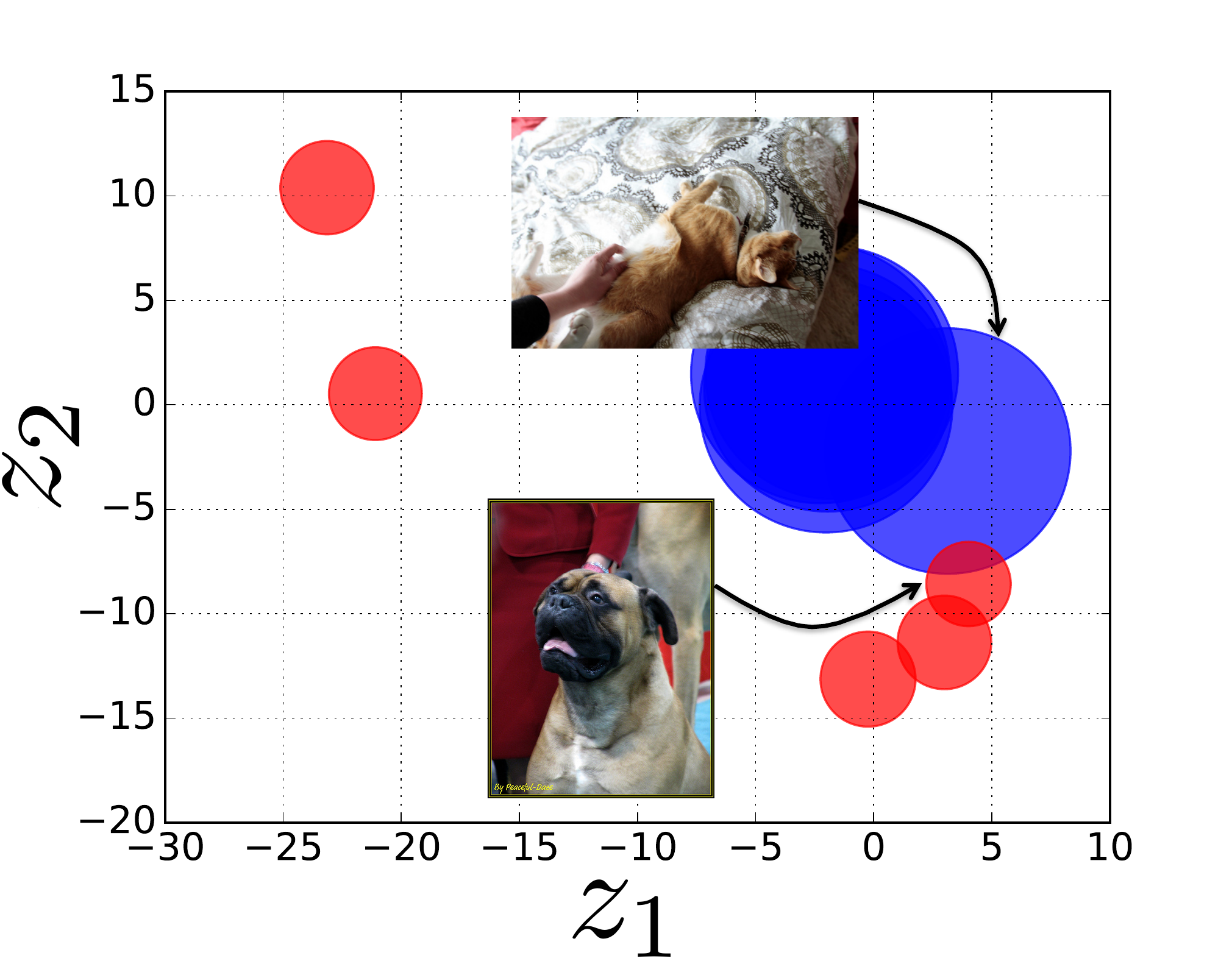}}\caption{Two dimensional latent space found from training our infinite mixture
of VAEs on Dogs dataset. We randomly selected 5 dog images and 5 images
of anything else and plotted their latent representation in each VAE
($z_{1}$ for the first dimension and $z_{2}$ for the second one).
The position of each circle represents the mean of the density for
the given image in this space and its radius is the variance ($\mu$
and $\sigma$ in Figure \ref{eq:vae_loss}, respectively). As shown,
representation of non-dogs (blue circles) are generally clustered
far away from the dogs (red circles). Moreover, dogs have smaller
variance than non-dogs, hence the VAEs are uncertain about the representation
of images that were not seen during training.}
\label{fig:dogs}
\end{figure*}

\begin{table*}
\begin{centering}
{\footnotesize{}}%
\begin{tabular}{|c|c|c|c|c|}
\hline 
{\footnotesize{}Method/Labels} & {\footnotesize{}100} & {\footnotesize{}1000} & {\footnotesize{}4000} & {\footnotesize{}All}\tabularnewline
\hline 
\hline 
{\footnotesize{}AlexNet }\cite{KrizhevskySutskeverHinton2012} & {\footnotesize{}$69.59\pm3.21$} & {\footnotesize{}$86.72\pm0.66$} & {\footnotesize{}$89.88\pm0.03$} & {\footnotesize{}$90.26\pm0.25$}\tabularnewline
\hline 
{\footnotesize{}Infinite Mixture} & {\footnotesize{}$75.81\pm1.83$} & {\footnotesize{}$89.28\pm0.19$} & {\footnotesize{}$90.68\pm0.05$} & {\footnotesize{}$91.69\pm0.17$}\tabularnewline
\hline 
\hline 
{\footnotesize{}Latent VAE+SVM} & {\footnotesize{}$49.81\pm1.87$} & {\footnotesize{}$63.28\pm0.64$} & {\footnotesize{}$74.8\pm0.2$} & {\footnotesize{}$79.6\pm0.7$}\tabularnewline
\hline 
{\footnotesize{}Latent Mixture+SVM} & {\footnotesize{}$58.1\pm2.63$} & {\footnotesize{}$72.28\pm0.2$} & {\footnotesize{}$79.8\pm0.18$} & {\footnotesize{}$83.9\pm0.24$}\tabularnewline
\hline 
\end{tabular}{\footnotesize{}}
\par\end{centering}{\footnotesize \par}
\caption{Test accuracy of AlexNet on the dogs dataset compared to our proposed
approach in the first two rows. Second two rows compare the latent
representations obtained from a single VAE compared to ours.}
\label{tbl:dogs}
\end{table*}

\subsection{Dogs Experiment}

ImageNet is a dataset containing $1,461,406$ natural images manually
labeled according to the WordNet hierarchy to 1000 classes. We
select a subset of $10$ breeds of dogs for our experiment. These
$10$ breeds are: ``Maltese dog, dalmatian, German shepherd, Siberian
husky, St Bernard, Samoyed, Border collie, bull mastiff, chow, Afghan
hound'' with $10,400$ training and $2,600$ test images. For an
illustration of the latent space and how the mixture of VAEs is able
to represent the uncertainty in the hidden variables we use this dogs
subset. We fine-tune a pre-trained AlexNet \cite{KrizhevskySutskeverHinton2012}
as the base discriminative model and share the parameters with the
generative model. In particular, we use the $4096$-dimensional output
of the $7$th fully connected layer (fc7) as the input for both softmax
experts and the VAE autoencoders. We trained the generative model
with all the unlabeled dog instance and used 1000 hidden units for
each VAE and set $\alpha=2$ and stopped with \emph{$14$ }autoencoders. 

We randomly select 5 images of dogs (from this ImageNet subset) and
$5$ images of anything else (non-dogs from Flicker with Creative
Common License) for the illustration in Figure \ref{fig:dogs}. We
plot the 2-dimensional latent representation of these images in $5$
VAEs of the learnt mixture. In each plot, the mean of the density
of the latent variable $\bz$ determines the position of the center
of the circle and the variance is shown as its radius (we use the
mean variance of the bivariate Gaussian for better illustration in
a circle). These values are calculated from each VAE network as $\mu$
and $\sigma$ in Figure \ref{eq:vae_loss}. As shown, the images of
non-dogs are generally clustered together in this latent space which
indicate they are recognized to be different. In addition, the variance
of the non-dogs are generally higher than the dogs. As such, even
when the mean of non-dogs are not discriminative enough (the dogs
and non-digs are not sufficiently well clustered apart in that VAE)
we are \emph{uncertain} about the representations that are not dogs.
This uncertainty leads to lower probability for the assignment to
the given VAE (from Equation \ref{eq:label_assign}) and subsequently
smaller weights when learning a mixture of experts model.

In Table \ref{tbl:dogs} the accuracy of AlexNet on this dogs subset
is shown and compared with our infinite mixture approach. As seen
infinite mixture performs better, particularly with smaller labeled
instances. In addition, latent representation of the infinite mixture
(computed as an expectation) when used in a SVM significantly outperforms
a single VAE. This illustrates the ability of our model in better
capturing underlying representations.

\subsection{CIFAR Dataset}

\begin{table}
{\footnotesize{}}%
\begin{tabular}{|c|c|c|c|}
\hline 
{\footnotesize{}Method/Labels} & {\footnotesize{}1000} & {\footnotesize{}4000} & {\footnotesize{}All}\tabularnewline
\hline 
\hline 
{\footnotesize{}Spike-and-slab \cite{GoodfellowCourvilleBengio2012}} &  & {\footnotesize{}31.9} & \tabularnewline
\hline 
{\footnotesize{}Maxout \cite{GoodfellowWarde-FarleyMirzaEtAl2013}} &  &  & {\footnotesize{}$9.38$}\tabularnewline
\hline 
{\footnotesize{}GDI \cite{PuYuanStevensEtAl2015}} &  &  & {\footnotesize{}$8.27$}\tabularnewline
\hline 
{\footnotesize{}Conv-Large \cite{RasmusValpolaHonkalaEtAl2015,SpringenbergDosovitskiyBroxEtAl2014}} &  & {\footnotesize{}$23.3\pm30.61$} & {\footnotesize{}$9.27$}\tabularnewline
\hline 
{\footnotesize{}$\Gamma\text{-Model}$ \cite{RasmusValpolaHonkalaEtAl2015}} &  & {\footnotesize{}$20.09\pm0.46$} & {\footnotesize{}$9.27$}\tabularnewline
\hline 
\hline 
{\footnotesize{}Residual Network \cite{HeZhangRenEtAl2015}} & {\footnotesize{}$10.08\pm1.12$} & {\footnotesize{}$8.04\pm.21$} & {\footnotesize{}$7.5\pm0.01$}\tabularnewline
\hline 
{\footnotesize{}Infinite Mixture of VAEs} & {\footnotesize{}$8.72\pm0.45$} & {\footnotesize{}$7.78\pm0.13$} & {\footnotesize{}$7.5\pm0.02$}\tabularnewline
\hline 
\end{tabular}\caption{Test error on CIFAR10 with various number of labeled training examples.
The results reported in \cite{RasmusValpolaHonkalaEtAl2015} did not
include image augmentations. Although the original approach in \cite{SpringenbergDosovitskiyBroxEtAl2014}
seems to offer up to $2\%$ error reduction with augmentation.}
\label{tbl:cifar10}
\end{table}

The CIFAR-10 dataset \cite{KrizhevskyHinton2009} is composed of 10
classes of natural $32\times32$ RGB images with $50,000$ images
for training and $10,000$ images for testing. Our experiments show
single VAE does not perform well for encoding this dataset as is also
confirmed here \cite{LarsenSoenderbyLarochelleEtAl2016}. However,
since our objective is to perform semi-supervised learning, we use
Residual network (ResNet) \cite{HeZhangRenEtAl2015} as a successful
model in image representation for discriminative learning to share
the parameters with our generative model. This model is useful for
complex problems where the unsupervised approach may not be sufficient.
In addition, autoencoders seek to preserve the distribution of the
pixel values required in reconstructing the images while this information
has a minimum impact on the final classification prediction. Therefore,
such parameter sharing in which generative model is combined with
the classifier is necessary for better prediction. 

As such we fine-tune a ResNet and use output of the $127$th layer
as the input for the VAE. We use a $2000$ hidden nodes and $\alpha=2$
to train an infinite mixture with \emph{$15$ }VAEs. For training
we augmented the training images by padding images with $4$ pixels
on each side and random cropping.

Table \ref{tbl:cifar10} reports the test error of running our approach
on this dataset. As shown, our infinite mixture of VAEs combined with
the powerful discriminative model outperforms the state-of-the-art
in this dataset. When all the training instances are used the performance
of our approach is the same as the discriminative model. This is because
with larger labeled training sizes, the instance weights provided
by the generative model are averaged and lose their impact, therefore
all the experts become similar. With smaller labeled examples on the
other hand, each softmax expert specializes in a particular aspect
of the data. 


\subsection{3D ModelNet}

\begin{figure}
\centering{}\includegraphics[scale=0.35]{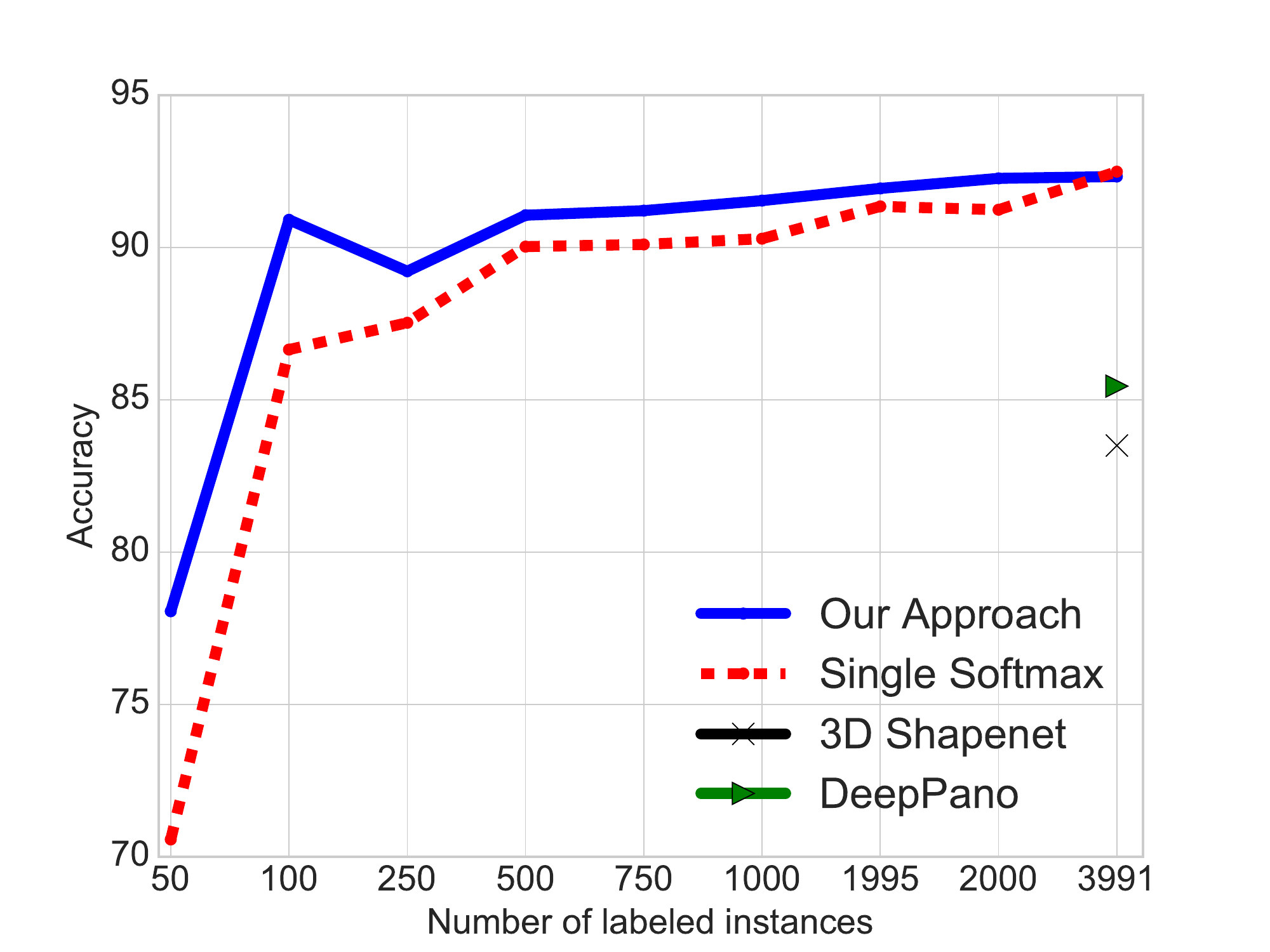}\caption{ModelNet10 compared to 3D Shapenet \cite{WuSongKhoslaEtAl2015} and
DeepPano \cite{ShiBaiZhouEtAl2015} averaged over 3 trials.}
\label{fig:modelnet}
\end{figure}

The ModelNet datasets were introduced in \cite{WuSongKhoslaEtAl2015}
to evaluate 3D shape classifiers. ModelNet has $151,128$ 3D models
classified into $40$ object categories, and ModelNet10 is a subset
based on classes in the NYUv2 dataset \cite{SilbermanHoiemKohliEtAl2012}.
The 3D models are voxelized to fit a $30\times30\times30$ grid and
augmented by $12$ rotations. For the discriminative model we use
a convolutional architecture similar to that of \cite{MaturanaScherer2015}
where we have a 3D convolutional layer with $32$ filters of size
$5$ and stride $2$, convolution of size $3$ and stride $1$, max-pooling
layer with size $2$ and a $128$-dimensional fully connected layer.
Similar to the CIFAR-10 experiment, we share the parameters of the
last fully connected layer between the infinite mixture of VAEs and
the discriminative softmax.

As shown in Figure \ref{fig:modelnet}, when using the whole dataset
our infinite mixture and the best result from \cite{MaturanaScherer2015}
match at $92\%$ accuracy. However, as we reduce the number of labeled
training examples it is clear that our approach outperforms a single
softmax classifier. 

Additionally, Table \ref{tbl:modelnet_svm} shows the accuracy comparison
of the latent representation obtained from the samples from our infinite
mixture and a single VAE as measured by the performance of SVM. As
seen, the expected latent representation in our approach is significantly
more discriminative and outperforms single VAE. This is because, we
take into account the variations in the input and adapt to the complexity
of the input. While a single VAE has to capture the dataset in its
entirety, our approach is free to choose and fit.
\begin{table}
\centering
{\footnotesize
\begin{tabular}{|c|c|c|c|}
\hline 
Method/Labels & 100 & 1000 & All\tabularnewline
\hline 
\hline 
VAE latent+SVM & 64.21 & 79.09 & 82.71\tabularnewline
\hline 
Mixture latent+SVM & 74.01 & 83.26 & 85.68\tabularnewline
\hline 
\end{tabular}}\caption{ModelNet10 accuracy of latent variable representation for training
SVM using a single VAE versus expected latent variable in our approach.}
\label{tbl:modelnet_svm}
\end{table}
Our experiments with both 2D and 3D images show the initial convolutional
layers play a crucial rule for the VAEs to be able to encode the input
into a latent space where the mixture of experts best perform. This
3D model further illustrate the decision function mostly depends on
the internal structure of the generative model rather than reconstruction
of the pixel values. When we share the parameters of the discriminative
model with the generative infinite mixture of VAEs and learn the mixture
of experts, we combine various representations of the data for better
prediction.

\section{Conclusion}

In this paper, we employed Bayesian non-parametric methods to propose an infinite
mixture of variational autoencoders that can grow to represent the
complexity of the input. Furthermore, we used these autoencoders to create a mixture of experts model for semi-supervised
learning. In both 2D images and 3D shapes, our approach provides
state of the art results in various datasets.

We further showed that such mixtures, where each component learns to represent a particular aspect of the data, are able to produce
better predictions using fewer total parameters than a single monolithic model. This applies whether the model is generative or discriminative. Moreover, in semi-supervised
learning where the ultimate objective is classification, parameter
sharing between discriminative and generative models was shown to provide better prediction accuracy.

In future works we plan to extend our approach to use variational
inference rather than sampling for better efficiency. In addition,
a new variational loss that minimizes the joint probability of the input
and output in a Bayesian paradigm may further increase the prediction accuracy
when the number of labeled examples is small.

\newpage
\bibliographystyle{ieee}
\bibliography{egbib}

\onecolumn
\appendix

\section{Mathematical Details of the Infinite Variational Autoencoder }

We have the following 
\begin{eqnarray*}
p(\bc,\btheta,\bx_{1,\ldots,n},\alpha) & = & \int\int p_{\btheta}(\bx_{1,\ldots,n}|\bc,\bz)p(\bz)p(\bc|\boldsymbol{\pi})p(\boldsymbol{\pi}|\alpha)d\boldsymbol{\pi}d\bz\\
 & = & \int p_{\btheta}(\bx_{1,\ldots,n}|\bc,\bz)p(\bz)\bigg[\int p(\bc|\boldsymbol{\pi})p(\boldsymbol{\pi}|\alpha)d\boldsymbol{\pi}\bigg]d\bz\\
 & = & \int\left(\bigg[\prod_{i}^{n}p{}_{\btheta_{\bc_{i}}}(\bx_{i}|\bz_{\bc_{i}})p(\bz_{\bc_{i}})\bigg]\bigg[\int p(\bc|\boldsymbol{\pi})p(\boldsymbol{\pi}|\alpha)d\boldsymbol{\pi}\bigg]\right)d\bz_{\bc_{i}}.
\end{eqnarray*}
To perform inference so that the distributions of the unknown parameters
are known, we use blocked \emph{Gibbs sampling} by iterating the following
two steps:
\begin{enumerate}
\item Sample for the unknown density of the observations with parameter
$\btheta$ (we drop $\alpha$because it is conditionally independent
$\bx$):
\begin{eqnarray*}
\bx_{1,\ldots,n},\btheta & \sim & p(\bx_{1,\ldots,n},\btheta|\bc)
\end{eqnarray*}
\item Sample the base VAE assignments:
\begin{eqnarray*}
\qquad\quad\bc & \sim & p(\bc|\bx_{1,\ldots,n},\btheta,\alpha)
\end{eqnarray*}
\end{enumerate}
For the first step, we use variational inference to find joint probability
of the input and its parameter $\btheta$ \emph{conditioned} on current
assignments. Using standard variational inference, we have, 
\begin{eqnarray*}
p(\bx_{1,\ldots,n},\btheta|\bc) & = & \int\prod_{i}p_{\btheta_{\bc_{i}}}(\bx_{i}|\bz_{\bc_{i}})p(\bz_{\bc_{i}})d\bz_{\bc_{i}}\\
 & = & \int\prod_{i}\frac{p_{\btheta_{\bc_{i}}}(\bx_{i}|\bz_{\bc_{i}})p_{\btheta}(\bz_{\bc_{i}})}{q_{\phi}(\bz_{\bc_{i}}|\bx_{i})}q_{\phi}(\bz_{\bc_{i}}|\bx_{i})d\bz_{\bc_{i}}\\
 & = & \int\prod_{i}p_{\btheta_{\bc_{i}}}(\bx_{i}|\bz_{\bc_{i}})\frac{p_{\btheta}(\bz_{\bc_{i}})}{q_{\phi}(\bz_{\bc_{i}}|\bx_{i})}q_{\phi}(\bz_{\bc_{i}}|\bx_{i})d\bz_{\bc_{i}}
\end{eqnarray*}
Taking the $\log$ from both sides and using Jensen's inequality,
we have the following lower bound for the joint distribution of the
observations conditioned on the latent variable assignments (VAE assignments
in the infinite mixture):
\begin{eqnarray}
\log\left(p(\bx_{1,\ldots,n},\btheta|\bc)\right) & \geq & \sum_{i}-\text{KL}\left(q_{\phi}(\bz_{\bc_{i}}|\bx_{i})\|p_{\btheta}(\bz_{\bc_{i}})\right)+\mathbb{E}_{q_{\phi}(\bz|\bx_{i})}[\log p_{\btheta_{\bc_{i}}}(\bx_{i}|\bz_{\bc_{i}})].\label{eq:vae_loss}\\
 &  & =\sum_{i}\mathbb{E}_{q_{\phi}(\bz|\bx_{i})}[\log p_{\btheta_{\bc_{i}}}(\bx_{i},\bz_{\bc_{i}})-\log q_{\phi}(\bz_{\bc_{i}}|\bx_{i})]
\end{eqnarray}
Here, $\btheta$ denotes all the parameters in the decoder network
and $\phi$ all the parameters in the encoder. Now, to compute the
expectations in both the KL-divergence and the conditional likelihood
of the second term, we use the sampling with the reparameterization
trick. Thus, Equation \ref{eq:vae_loss} is rewritten as
\begin{eqnarray*}
\mathbb{E}_{q_{\phi}(\bz|\bx_{i})}[\log p_{\btheta_{\bc_{i}}}(\bx_{i},\bz_{\bc_{i}})-\log q_{\phi}(\bz_{\bc_{i}}|\bx_{i})] & \approx & \frac{1}{L}\sum_{\ell=1}^{L}\log p_{\btheta_{\bc_{i}}}(\bx_{i},\bz_{\bc_{i}}^{\ell})-\log q_{\phi}(\bz_{\bc_{i}}^{\ell}|\bx_{i})
\end{eqnarray*}
where $\bz$ is taken from a differentiable function that performs
a random transformation of $\bx$. This differentiable transformation
function allows for using stochastic gradient descent in backpropagation
algorithm. We use $L=2$ in our experiments. 

For sampling the base VAE assignment $\bc$, we know that 
\begin{eqnarray*}
p(\bc|\bx_{1,\ldots,n},\btheta,\alpha) & = & \int\underbrace{p(\bc|\bx_{1,\ldots,n},\btheta,\boldsymbol{\pi})}_{\text{Multinomial distribution}}\underbrace{p(\boldsymbol{\pi}|\alpha)}_{\text{Dirichlet distribution}}d\boldsymbol{\pi}
\end{eqnarray*}

This integral corresponds to a Multinomial distribution with Dirichlet
prior where the number of components $C$ is a constant. We have,
\begin{eqnarray*}
p(\bc_{1}\ldots,\bc_{C}|\bx_{1,\ldots,n},\btheta,\boldsymbol{\pi}) & = & \prod_{j}^{C}\boldsymbol{\pi}_{j}^{n_{j}},\qquad\qquad n_{j}=p_{\btheta_{j}}(\bc_{i}=j|\bx_{i})\times\sum_{i=1}^{n}\mathbb{I}[\bc_{i}=j],
\end{eqnarray*}
Using the standard Dirichlet integration, we have
\begin{eqnarray*}
p(\bc_{1}\ldots,\bc_{C}|\bx_{1,\ldots,n},\btheta) & = & \int p(\bc_{1}\ldots,\bc_{C}|\bx_{1,\ldots,n},\btheta,\boldsymbol{\pi}_{1},\ldots,\boldsymbol{\pi}_{C})p(\boldsymbol{\pi}_{1},\ldots,\boldsymbol{\pi}_{C})d\boldsymbol{\pi}_{1},\ldots,\boldsymbol{\pi}_{C}\\
 & = & \frac{\Gamma(\alpha)}{\Gamma(\alpha+n)}\prod_{j=1}^{C}\frac{\Gamma(n_{j}+\alpha/C)}{\Gamma(\alpha/C)}
\end{eqnarray*}
where we can draw samples from the conditional probabilities as 
\begin{eqnarray*}
p(\bc_{i}|\bc_{i-1},\bc_{i+1}\ldots,\bc_{C},\bx_{1,\ldots,n},\btheta) & = & \frac{\eta_{j}(\bx_{i})+\alpha/C}{n-1+\alpha}.
\end{eqnarray*}
When taking the number of components to approach infinity, $C\to\infty$
it is easy to see that the results in the paper is obtained.

\section{Base Variational Autoencoder's Architecture}

The base autoencoder contains the following layers: 
\begin{enumerate}
\item Input layer: Depending on the type of the input it's dimensions are
different.
\item A fully connected layer with the number of hidden dimensions $\mathbf{h}$.
This number of hidden dimensions is what is changed during training
and infinite mixture uses infinite hidden dimensions. We use batch
normalization in the input of this layer that according to our experiments
helps with the convergence and better performance. The output of the
batch normalization is used in $\tanh$ nonlinearity units. The original
VAE paper did not use batch normalization.
\item The output of the last hidden layer is used in another fully connected
layer with linear units to estimate the mean of the density layer
$\mu$. This is the mean of the Gaussian density of the latent variable
$\bz$. Since this density is multivariate, we found $10$ percent
of the hidden dimensions to performing the best. For the Dogs experiment
in the paper, we used $2$-dimensional latent space.
\item Similar to the mean layer $\mu$, we have another layer with the same
dimensions for estimating the diagonal entries of the density of the
latent space $\sigma$.
\item For the decoder, we need to sample from the density of the latent
variable $\bz$ to compute the reconstruction. We use two samples
though-out our experiments to estimate the expected reconstruction
following below steps for the decoder:
\begin{enumerate}
\item Sample from the latent multivariate Gaussian distribution with mean
$\mu$ and variance $\sigma$.
\item The sample is used in another fully connected layer with hidden dimensions
$\mathbf{h}$. We use batch normalization in the input of this layer
too. Batch normalization helps with searching for a latent space in
a lower dimensions. The output of the batch normalization is used
in $\tanh$ nonlinearity units. 
\item The output of the batch normalized latent space is used in another
fully connected layer with sigmoid nonlinearity unit to reconstruct
the input.
\end{enumerate}
\end{enumerate}
It should be noted that this non-symmetric autoencoder corresponds
to the binary VAE described in the original paper (with minor changes
that helped with its convergence stability and performance). We found
this architecture to perform better than its alternative symmetric
one for the semi-supervised learning application of ours for CIFAR-10
and MNIST. In evaluation of the model for computing the loss, we use
the cross-entropy measure for $p(\bx|\bz)$ since the variables are
considered binary.

For 3D ModelNet and Dogs dataset, we use a symmetric variant that
showed to be more effective. In the symmetric version, we changed
all the $\tanh$ units to softplus ( $\log(1+\exp(x))$). The final
step 5c is changed to the following: We use the hidden layer $\mathbf{h}$
to feed into two fully connected layers for the mean and variance
of the decoder $\mu_{\text{dec}}$, $\sigma_{\text{dec}}$. We then
sample from this decoding density for the reconstruction of the input.
For computing the loss, we just use the log-likelihood of the reconstruction.

All the code is implemented in Python using Lasagne\footnote{https://github.com/Lasagne/Lasagne}
and Theano\footnote{http://deeplearning.net/software/theano/}. We
will release the code for the submission amongst with the dogs dataset.

\end{document}